%% file: main.tex
\documentclass[journal]{IEEEtran}

\usepackage{graphicx}
\usepackage[caption=false,font=footnotesize]{subfig}
\usepackage[colorlinks=false]{hyperref}

\usepackage{textcomp}

\usepackage{amsmath}
\usepackage{amssymb}

\usepackage{algorithm}
\usepackage{algpseudocode}

\usepackage{tikz}
\usepackage{pgfplots}
\usepackage{xcolor}
\usepackage{fancyhdr}
\usetikzlibrary{arrows,positioning}
\pgfplotsset{compat=1.3}

\newcommand{\figref}[1]{Fig.~\ref{#1}}
\newcommand{\tabref}[1]{\autoref{#1}}
\renewcommand{\algref}[1]{\autoref{#1}}
\newcommand{\sectref}[1]{\autoref{#1}}

\newcommand{\algline}[1]{$\ell$.~{\footnotesize #1}}
\newcommand{\alglines}[1]{$\ell\ell$.~{\footnotesize #1}}

\newcommand{\vect}[1]{\mathbf{#1}}
\newcommand{\matx}[1]{\mathbf{#1}}
\newcommand{\point}[1]{\vect{#1}}
\newcommand{\pp}{\point{x}}
\newcommand{\pq}{\point{y}}

\newcommand{\myspace}[1]{\mathbb{#1}}
\newcommand{\I}{\mathcal{I}}
\newcommand{\R}{\myspace{R}}
\newcommand{\N}{\myspace{N}}

\newcommand{\f}[1]{\mathrm{#1}}
\newcommand{\set}[1]{\mathbf{#1}}

\newcommand{\T}{\mathsf{T}}

\newcommand{\obj}[1]{\mathcal{#1}}

\newcommand{\fr}[1]{\mathcal{#1}}
\newcommand{\tf}[3]{{}_{#2}^{#3}{#1}}
\newcommand{\TF}[3]{{}_\fr{#2}^\fr{#3}{#1}}

\newcommand{\map}[1]{{#1}}

\newcommand{\U}{\map{U}}
\newcommand{\G}{\map{G}}
\newcommand{\E}{\map{E}}

\newcommand{\SU}{\myspace{S}_\U}
\newcommand{\SG}{\myspace{S}_\G}

\let\oldmaketitle\maketitle
 
\renewcommand{\maketitle}{%
  \oldmaketitle
  \thispagestyle{fancy}
}

\begin{document}
\title{Robust Intrinsic and Extrinsic Calibration of RGB-D Cameras}
%
%
%

\author{Filippo~Basso,
        Emanuele~Menegatti,
        and~Alberto~Pretto%
\thanks{Basso is with IT+Robotics (\url{www.it-robotics.it}), e-mail: \href{mailto:filippo.basso@it-robotics.it}{filippo.basso@it-robotics.it}. Menegatti is with the Department of Information Engineering, University of Padua, Italy, e-mail: \href{mailto:emg@dei.unipd.it}{emg@dei.unipd.it}. Pretto is with the Department of Computer, Control, and Management Engineering ``Antonio Ruberti'', Sapienza University of Rome, Italy, e-mail: \href{mailto:pretto@diag.uniroma1.it}{pretto@diag.uniroma1.it}}.}

%


\maketitle

\begin{abstract}

Color-depth cameras (RGB-D cameras) have become the primary sensors in most robotics systems, from service robotics to industrial robotics applications. Typical consumer-grade RGB-D cameras are provided with a coarse intrinsic and extrinsic calibration that 
generally does not meet the accuracy requirements needed by many robotics applications (e.g., highly accurate 3D environment reconstruction and mapping, high precision object recognition and localization, \dots).\\
In this paper, we propose a human-friendly, reliable and accurate calibration framework that enables to easily estimate both the intrinsic and extrinsic parameters of a general color-depth sensor couple. Our approach is based on a novel two components error model.
This model unifies the error sources of RGB-D pairs based on different technologies, such as structured-light 3D cameras and time-of-flight cameras. 
\\
Our method provides some important advantages compared to other state-of-the-art systems: it is general (i.e., well suited for different types of sensors), based on an easy and stable calibration protocol, provides a greater calibration accuracy, and has been implemented within the ROS robotics framework.\\
We report detailed experimental validations and performance comparisons to support our statements.

\end{abstract}


%
\IEEEpeerreviewmaketitle

\input{introduction}
\input{related_work}
\input{depth_error}
\input{approach}
\input{undistortion_map}
\input{global_map}
\input{experiments}

\input{conclusions}

\ifCLASSOPTIONcaptionsoff
  \newpage
\fi



\bibliographystyle{IEEEtran}
\bibliography{IEEEabrv,bibl,common,iaslab}
%
%
%

%

\begin{IEEEbiography}[{\includegraphics[width=1in,height=1.25in,clip,keepaspectratio]{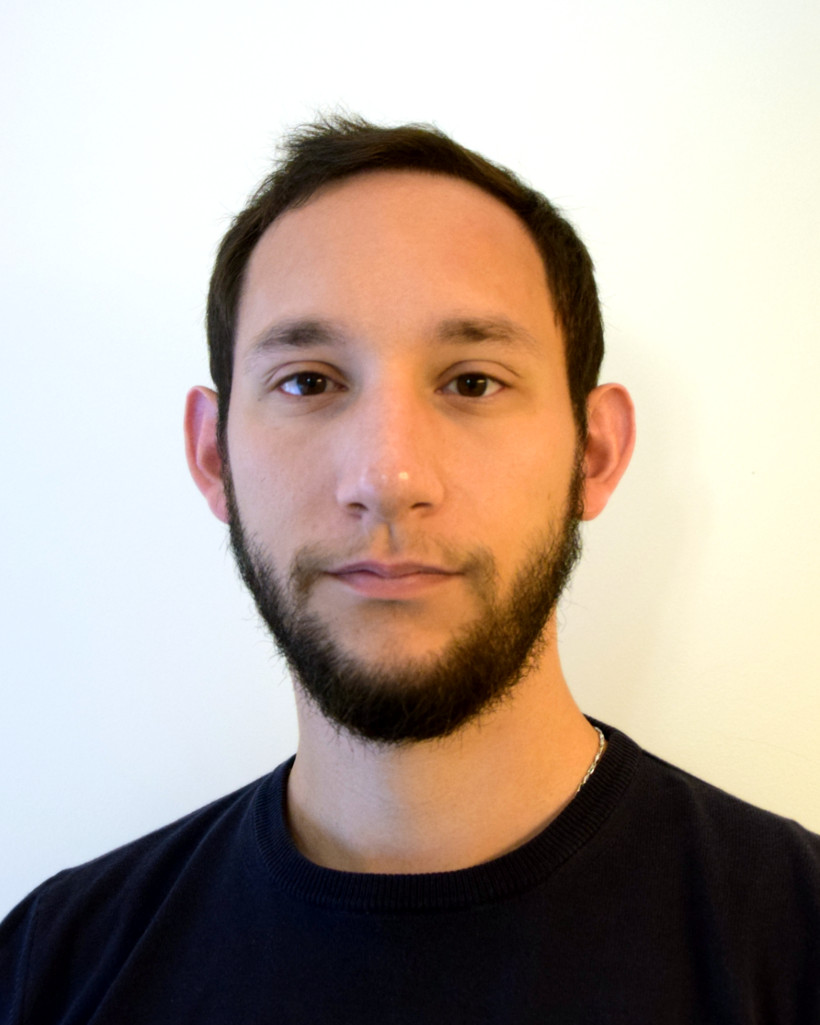}}]{Filippo Basso}
was born in Italy in 1987. He received the B.Sc., M.Sc., and Ph.D. degrees in Computer Engineering from University of Padua, Padua, Italy, in 2009, 2011, and 2015, respectively.
He joined IT+Robotics, Padua, Italy, in 2015, a company active in the
field of industrial vision, robotics, and workcell simulation applications, where he is currently a Senior Developer and R\&{}D Manager.
His main areas of research interest are 2D/3D computer vision, sensor fusion, and robotics.
\end{IEEEbiography}

\begin{IEEEbiography}[{\includegraphics[width=1in,height=1.25in,clip,keepaspectratio]{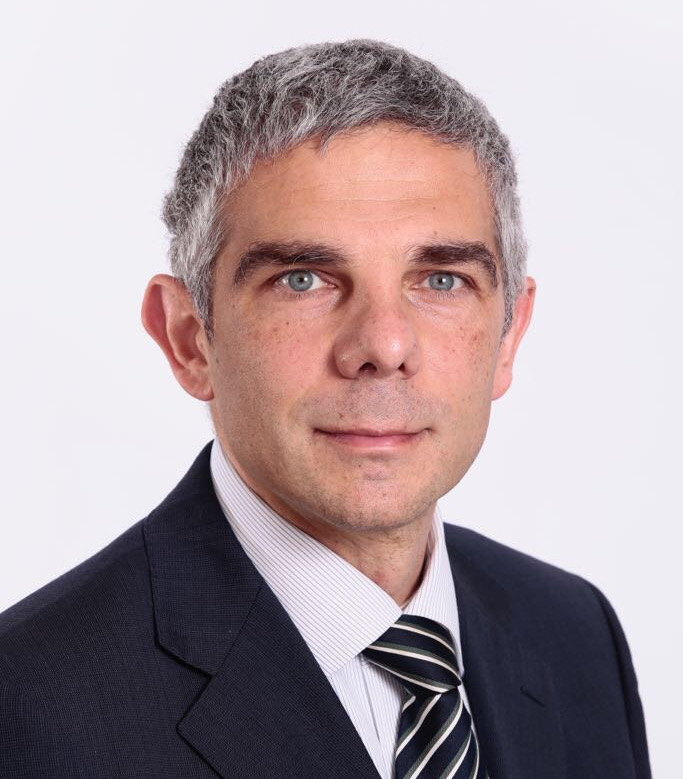}}]{Emanuele Menegatti}
Menegatti is Full Professor of the School of Engineering at Dept. of Information Engineering of University of Padua since 2017. He received his Ph.D. in Computer Science in 2003, in 2005 he became Assistant Professor and Associate Professor in 2010. Menegatti was guest editor of five special issues of the journal Robotics and Autonomous Systems Elsevier. Menegatti's main research interests are in the field of Robot Perception and 3D distributed perception systems. In particular, he is working on RGB-D people tracking for camera network, neurorobotics and service robotics. He is teaching master courses on “Autonomous Robotics”, “Three-dimensional data processing” and bachelor course in “Computer Architecture” and a course for school teachers on “Educational Robotics”.
He was coordinator of the FP7 FoF-EU project “Thermobot” and local principal investigator for the European Projects “3DComplete” and “FibreMap”  and “Focus” in FP7; “eCraft2Learn” and “Spirit” in H2020. 
He was general chair of the 13th International Conference "Intelligent Autonomous System" IAS-13 and was program chair of IAS-14 and IAS-15. He is author of more than 50 publications in international journals and more than 120 publications in international conferences.
In 2005, Menegatti founded IT+Robotics, a Spin-off company of the Univ. of Padua, active in the field of industrial robot vision, machine vision for quality inspection, automatic off-line robot programming. In 2014, he founded EXiMotion a startup company active in the field of educational robotics and service robotics.
\end{IEEEbiography}

\begin{IEEEbiography}[{\includegraphics[width=1in,height=1.25in,clip,keepaspectratio]{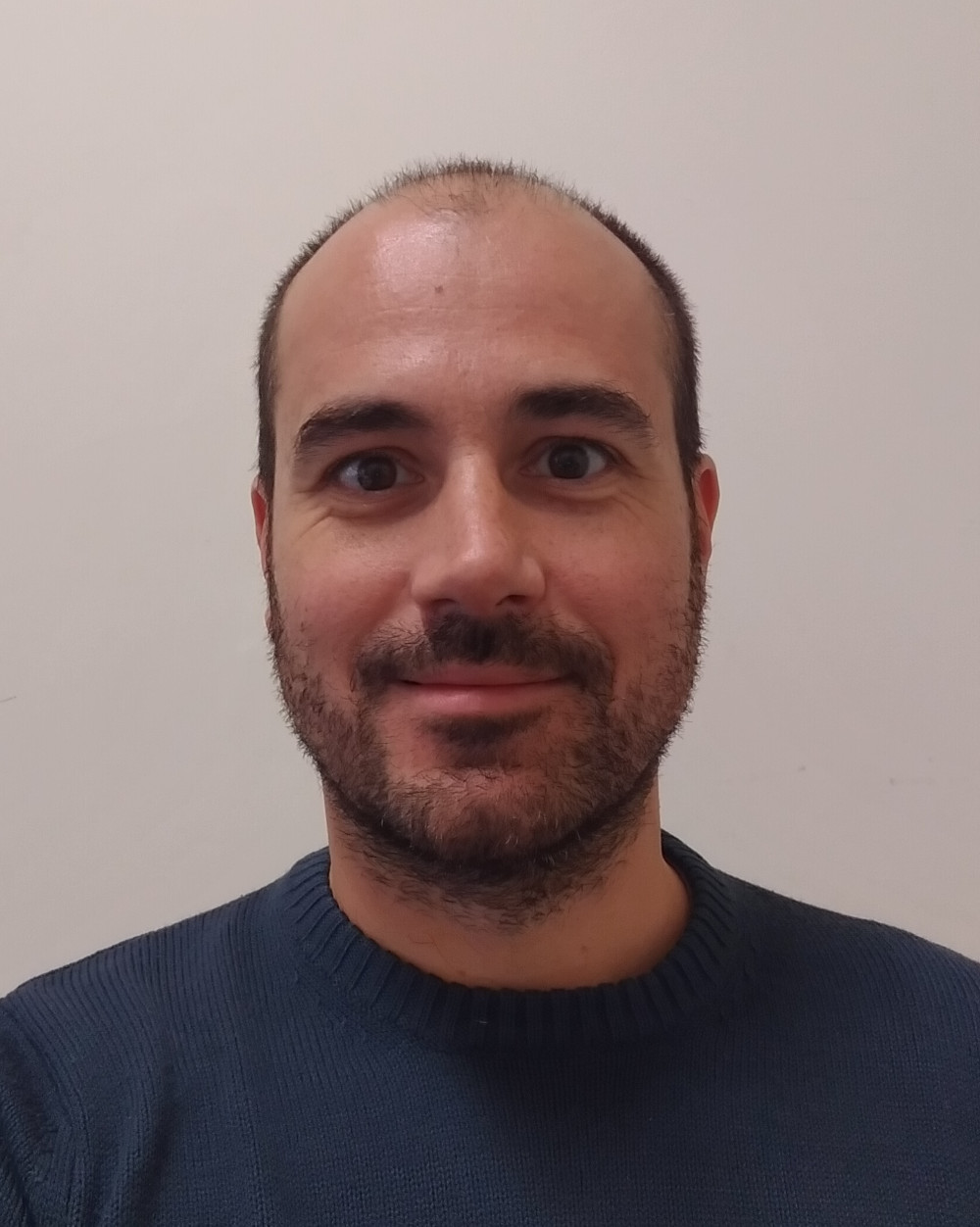}}]{Alberto Pretto}
Alberto Pretto is Assistant Professor at Sapienza University of Rome since October 2013. 
He received his Ph.D. degree in October 2009 from the University of Padua, where he worked as a postdoctoral researcher at the Intelligent Autonomous Systems Lab (Department of Information Engineering). Between 2011 and 2012 he spent a 9 months visiting research fellowships at the UCLA VisionLab, Los Angeles (USA). In 2004 and 2005, he has been working as software engineer at Padova Ricerca Scpa. In 2005, he was one of the funders of IT+Robotics Srl, a spin-off company of the University of Padua working on robotics and machine vision. 
Alberto Pretto's main research interests include robotics and computer vision.
\end{IEEEbiography}








\end{document}

%% file: introduction.tex
\section{Introduction}

\IEEEPARstart{T}{he} availability of affordable depth sensors in conjunction with common RGB cameras, often embedded in the same device (called RGB-D cameras), has provided mobile robots with a complete and instantaneous representation of both the appearance and the 3D structure of the surrounding environment.
Many robotic tasks highly benefit from using such sensors, e.g., SLAM and navigation \cite{endres2013tro, LabbeMichaud2014}, tracking \cite{Munaro2014, choi2013}, object recognition and localization \cite{Tang2012} and many others. While color information is typically provided by RGB cameras, there are many technologies able to provide depth information, e.g. time-of-flight (ToF) cameras, laser range scanners and sensors based on structured-light (SL).
Even if there are many devices able to provide both color and depth data, as far as we know, there are no integrated, CMOS-like, imaging sensors able to provide both color and depth information yet. Most of the RGB-D sensors currently used in robotics applications (among others, the Microsoft Kinect 1 and Kinect 2, the Asus Xtion, and the Intel RealSense) are composed by an RGB camera and a depth camera rigidly coupled in the same frame.
In order to obtain a reliable and accurate scene representation, not only the intrinsic parameters of each camera should be precisely calibrated, but also the extrinsic parameters relating the two sensors should be precisely known. 
RGB-D devices are often factory calibrated, with the calibration parameter set stored inside a non-volatile memory. Unfortunately, the quality of such calibration is only adequate for gaming purposes. For instance, with a default setup, the acquired point clouds can suffer from a non accurate association between depth and RGB data, due to a non perfect alignment between the camera and the depth sensor. Moreover, depth images can suffer from an irregular geometric distortion and a systematic bias in the measurements.
A proper calibration method for robust robotics applications is needed.

In this paper, we propose a novel, two-steps calibration method that employs a simple data collection procedure that only needs a minimally structured environment and that does not require any parameters tuning or a great interaction with the calibration software. 
The proposed method automatically infers the intrinsic calibration of the depth sensor by means of two general correction maps and, as a ``side effect'', the rigid body transformation that relates the two cameras, i.e., the camera pair extrinsic calibration. We assume that the RGB camera has been previously calibrated using a standard method, e.g., \cite{zhang2000PAMI}, while for the depth sensor calibration, we employ a two components error model that includes a pixel-based distortion error along with a systematic error. Even if the principal target of the proposed method are structured-light RGB-D sensors, this two components error model is designed to be ``technology-agnostic'', thus well generalizes also with sensors based on different technologies such as ToF cameras, as shown in error analysis and experiments.\\
The calibrated measurements are obtained employing two maps.
We propose to represent the first map, used to correct the distortion error, by means of a set of functions of the depth measurements, iteratively fitted to the acquired data during a first calibration stage.
The second map is obtained in a second stage of the calibration procedure, when we include the systematic error and the transformation between the sensors. At this point we exploit the plane-to-plane constraints between color and depth data to align the two sensors and to infer the systematic error correction functions inside a non-linear optimization framework.

The main contributions of this paper are the following:
\begin{itemize}
 \item A general and experimentally supported measurements error model, that well describes the error of different depth sensor types in an unified way.
 \item A spatial/parametric undistortion map that models in a compact and efficient way the distortion effect of structured-light depth sensors.
 \item A novel optimization framework that aims to estimate the camera-depth sensor rigid displacement along with the parametric model which describes the systematic error on the depth measurements.
 \item An open source implementation of the proposed method, integrated inside the ROS (Robot Operating System) framework \cite{ros}. The code along with a tutorial of the calibration process is available on the following website:
 \begin{quote}
  \url{http://iaslab-unipd.github.io/rgbd_calibration}\footnote{The copy and paste function may not work properly with this url due to the underscore symbol.}.
\end{quote}
\end{itemize}
An exhaustive set of tests, proving the soundness of the proposed method, is reported.
We also compare our method with other state-of-the-art systems, using the original implementations provided by their authors: our method appears to be more stable and able to provide the most accurate results.
Finally we report some experiments of a RGB-D visual odometry system applied to a mobile robot, where we show that the accuracy in the ego-motion estimation highly benefits from using RGB-D data calibrated with our system.

This paper is structured as follows. Related work is reviewed in \sectref{sec:depth_related_work}. In \sectref{sec:depth_error_model} the error on the depth values provided by the sensor is analyzed and discussed.
\sectref{sec:calibration_algorithm_approach} gives a quick overview of the calibration procedure.
The first calibration step is detailed in \sectref{sec:undistortion_map_estimation}, while \sectref{sec:global_map_estimation} describes the second calibration step.
The results of the calibration procedure and performance comparisons are reported in \sectref{sec:depth_experiments}.
Finally, some conclusions are reported in \sectref{sec:depth_conclusions}.

\subsection{Basic Notations}

We use non-bold characters $x$ to represent scalars, bold lower case letters $\vect{x}$ to represent vectors, with no distinction
between cartesian coordinates and homogeneous coordinates. The coordinates of a point $\pp$ with respect to the coordinate frame $\obj{A}$ are denoted by $\TF{\pp}{}{A}$; $\TF{\matx{T}}{A}{B}$ denotes the homogeneous transformation matrix\footnote{Here we implicitly assume that the points are expressed using homogeneous coordinates.} from the reference frame $\obj{A}$ to the frame $\obj{B}$ , such that $\TF{\pp}{}{B}  = \TF{\matx{T}}{A}{B} \TF{\pp}{}{A}$.\\
An RGB camera is denoted by $\obj{C}$ and it provides an RGB image $\I_\obj{C}$; a depth sensor is denoted by $\obj{D}$ and it provides a depth image $\I_\obj{D}$. From an RGB image of a scene that contains a checkerboard it is possible to extract the checkerboard corners $\TF{\map{B}}{}{\I_C}$, where the superscript $\I_\obj{C}$ explicits the fact that the corners are expressed in 2D pixel coordinates, i.e., $\TF{\map{B}}{}{\I_C} = \{(u,v)_{1}, \dots, (u,v)_{n}\}$. From a depth image $\I_\obj{D}$ it is possible to generate a point cloud $\TF{\map{C}}{}{D}$, where the superscript $\obj{D}$ explicits the fact that the point cloud is expressed with respect to the coordinate frame $\obj{D}$, i.e., $\TF{\map{C}}{}{D} = \{\pp_{1}, \dots, \pp_{m}\}$.

%% file: related_work.tex
\section{Related Work}
\label{sec:depth_related_work}

The availability in the market of affordable depth sensors (structured-light 3D sensors and time-of-flight cameras) has greatly increased the interest in depth sensor systems in both the robotics and computer vision communities. \\
Several ToF cameras have been proposed over the years, and at the same time many researches began to analyze the sources of errors of such sensors \cite{FoixSensors2011, MarioOE2009} and to propose suitable calibration methods to solve them \cite{Horaud2016}. A number of ToF-specific calibration methods \cite{LINDNER20101318, KuznetsovaECCVWS2015, BMVC2015_102} address the systematic nonlinear depth distortion that affects such sensor (often referred to as \textit{wiggling} error), an error  that depends only on the measured depth for each pixel \cite{FoixSensors2011}. Lindner \emph{et al.} \cite{LINDNER20101318} proposed a method that combines intrinsic parameter estimation with wiggling and intensity adjustment, where the wiggling error has been faced by means a depth correction B-spline function. 
Other methods use regression to compensate such depth error: Kuznetsova and Rosenhahn \cite{KuznetsovaECCVWS2015} exploit a non-parametric Gaussian kernel regression, while Ferstl \emph{et al.} \cite{BMVC2015_102} use Random Regression Forest: both these methods require to couple the depth sensor with an RGB camera.\\
Kim \emph{et al.} \cite{kim2008design} presented an analysis of the measurement errors of a Mesa SwissRanger time-of-flight camera, highlighting the presence of two components: a random noise and a systematic depth error, consistent over time. Using this depth error model, they proposed a three-stages calibration procedure that aims to estimate both the intrinsic and extrinsic parameters of an array of sensors composed by both ToF and video cameras.
Jung \emph{et al.} \cite{JungJPHKK11} proposed to calibrate extrinsics and intrinsics of a color camera and a ToF camera pair by using a pattern with 4cm-diameter holes that can be simultaneously localized by both sensors. Unfortunately, the error models used in both methods are conceived for a SwissRanger-like sensor, and can not be easily adapted to depth sensors based on different technologies.

Early methods addressing the calibration of RGB-D pairs were intended to estimate only the extrinsic parameters that relate the two sensors. Mei and Rives \cite{mei06b} addressed the problem of finding the relative pose between a 2D laser range finder and a catadioptric camera by using a planar calibration pattern perceived from both sensors. Scaramuzza \textit{et al.} \cite{scaramuzza2007} proposed to map 3D range information collected with a 3D tilting laser range finder into a 2D map (called Bearing Angle image) that highlighted the salient points of a scene thus the user could manually associate points between the two sensors in an easy way. The final extrinsic calibration was then obtained using a Perspective-n-Point (PnP) algorithm followed by a non-linear refinement step.\\

Depth sensors based on structured light technology have gained a great popularity thanks to the first version of Microsoft Kinect, an affordable and effective RGB-D pair based on active stereo matching that exploits a near-infrared random pattern projector. Early works on the Kinect sensor calibration \cite{burrus_online2011, konolige_online2011, smisek20113d, khoshelham2012accuracy} were based on standard RGB camera calibration techniques.
They used a checkerboard pattern to calibrate both the RGB and the IR cameras, often blocking the IR projector and illuminating the target with a halogen lamp in order to better highlight the checkerboard corners in the IR camera.
Smisek \emph{et al.} \cite{smisek20113d} showed that Kinect depth sensors are affected by a sort of radially symmetric distortion.
To correct such distortion, they estimated a $z$-correction image built as the pixel-wise mean of the residuals of the plane fitted to the depth data.
The $z$-correction image was subtracted from the $z$ coordinate of the cloud points to obtain the actual depth values.
Unfortunately it is currently well known that in the general case this distortion depends on the depth, i.e. becomes much stronger for increasing depths (e.g., see \sectref{sec:depth_error_model}).
Also Zhang and Zhang \cite{zhang2011calibration} and Mikhelson \emph{et al.} \cite{mikhelson2014automatic} exploited a checkerboard pattern but, differently from the previous approaches, they didn't search for checkerboard corners in the IR image.
Zhang \emph{et al.} \cite{zhang2011calibration} used the fact that points on the checkerboard should be co-planar, and the plane equation can be estimated with the RGB camera.
They also proposed to model the depth error of the Kinect sensor by treating the depth value $z$ as a linear function of the real one $z^*$, that is $z = \mu z^* + \eta$.
A limitation of this method comes from the fact that it requires to have a good initialization of the unknown parameters.
Mikhelson \emph{et al.} \cite{mikhelson2014automatic} extracted the structural corners from the point cloud derived from the depth images, in order to locate the checkerboard position also in the depth image plane.
This approach assumes that the depth image is not affected by any distortion: as mentioned before, this assumption does not always apply for Kinect-like sensors.
Moreover, from our experience, extracting structural corners from a point cloud is an operation that sometimes provides poor results.

Herrera \emph{et al.} \cite{herrera2011accurate, herrera2012joint} described a calibration method that exploits a checkerboard pattern attached to a big plane to calibrate two color cameras and a Kinect-like depth sensor.
This method works directly on the raw data provided by the depth sensor (instead of on the metric data) and, alongside the camera-depth sensor relative displacement, it estimates a disparity distortion correction map that depends on the observed disparity.
They estimated a coefficient for each pixel $\map{D}(u, v) \in \R$ and two global coefficients $\alpha_0, \alpha_1 \in \R$ such that the actual depth value $z^*$ can be computed as
\begin{equation*}
z^* = z + \map{D}(u, v) \cdot \exp(\alpha_0 - \alpha_1 \cdot z) \enspace.
\end{equation*}
Moreover, they used the 4 corners of the checkerboard plane as the initial guess of the relative displacement between the cameras and the depth sensor.
For short distances their approach seems to obtain good results, as reported also in \cite{staranowicz2015pratical, xiang2015review}.

An improvement over the work of Herrera \emph{et al.} is the one presented by Raposo \emph{et al.} \cite{raposo2013fast}.
They proposed several modifications to the estimation pipeline that allow their algorithm to achieve a calibration accuracy similar to \cite{herrera2012joint} while using less than $1/6$ of the input frames and running in $1/30$ of the time.

Canessa \emph{et al.} \cite{canessa2014calibrated}, instead, proposed to model the depth error by means of a second degree polynomial for each pixel.
In their work, authors first estimated the pose of a ``virtual'' depth sensor with respect to the RGB camera using an incandescent lamp bulb to light the checkerboard and make the depth map saturate in correspondence of the white squares. Then, they positioned a Kinect in front of a plane with a checkerboard attached and acquired a set of images from $0.6$ m to $2$ m.
Finally, they fitted a second degree polynomial to the sample set of every pixel.
Actually, the need of an incandescent lamp makes this system quite cumbersome.

Teichman \emph{et al.} \cite{teichman2013unsupervised} proposed a completely different calibration approach for Kinect-like devices: the undistortion map is estimated by means of a SLAM framework, in an unsupervised way.
Their algorithm estimates 6 depth multipliers, at $0, 2, \dots, 10$ meters, and it corrects the depth measurements using a linear interpolation: to the best of our knowledge, this is one of the first approaches that proved to be able to correct depth data at more than 5 meters.
The main drawback of their approach is the time it needs to reach a solution: the optimization procedure takes several hours to converge.
Moreover, according to \cite{xiang2015review}, it seems not to perform as well as \cite{herrera2012joint} for short distances.
Indeed, in \cite{teichman2013unsupervised} close-range measurements (i.e., less than 2 meters) are considered reliable and directly used in the SLAM pipeline to infer the geometry of the scene.

Fiedler et al. \cite{fiedler2013impact} investigated the influence of thermal and environmental conditions on the Kinect's data quality.
The experiments turned out that variations of the temperature and air draft have a notable influence on Kinect images and range measurements.
They derived temperature-related rules to reduce errors in both the calibration and measurement process.

Recently, Di Cicco \emph{et al}. \cite{DiCicco2015} proposed a non-parametric, unsupervised intrinsics calibration method for depth sensors.
The best plane fitted to the data is used as a reference, i.e., the average depth is considered reliable.
A complete, discretized undistortion map for the depth data is estimated using a machine learning approach.
In this approach the extrinsics calibration of a general RGB-D pair is not considered.

Staranowicz \emph{et al.} \cite{staranowicz2015pratical} proposed an RGB-D pair calibration algorithm that uses a spherical object moved in front of the camera as a calibration pattern.
A limitation of this method is that they focus on the estimation of the rigid displacement between the two cameras only, i.e., the depth error is not corrected.

In our previous work \cite{bassoICRA2014}, we proposed a spatial/parametric undistortion map that models the distortion effect of Kinect-like depth sensors in a compact way.
Our approach uses a calibration pattern similar to the one used in \cite{herrera2011accurate}: besides the undistortion map, we estimate the camera-depth sensor alignment along with a parametric model that well describes the systematic error of the depth measurements.
Results were very promising, but this method could get stuck in a local minimum if the depth camera intrinsic parameters (focal lengths and central point) were poorly estimated. In this work we improve our previous method by addressing this limitation with a new, more general error model and a new  calibration pipeline that refines also the depth camera intrinsic parameters. The calibration protocol employed in our method is inspired by \cite{herrera2011accurate} and \cite{teichman2013unsupervised}, while our error model has been designed taking inspiration from both the error models presented in \cite{kim2008design} and \cite{canessa2014calibrated}. 

%% file: depth_error.tex
\section{Depth Error Analysis} \label{sec:depth_error_model}

In this section we introduce our depth error correction model, derived from an experimental analysis performed using two popular structured-light based depth sensors (the Microsoft Kinect and the Asus Xtion Pro Live RGB-D cameras) and a time-of-flight sensor (the Microsoft Kinect 2 RGB-D camera). 

\begin{figure}
\centering
\scriptsize
\subfloat[\textsc{kinect1a} -- top]{\includegraphics[width=0.44\linewidth]{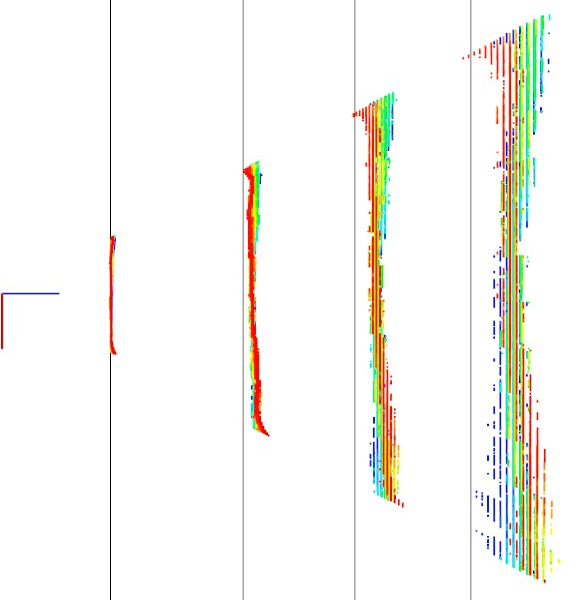}}
\subfloat[\textsc{kinect1b} -- top]{\includegraphics[width=0.44\linewidth]{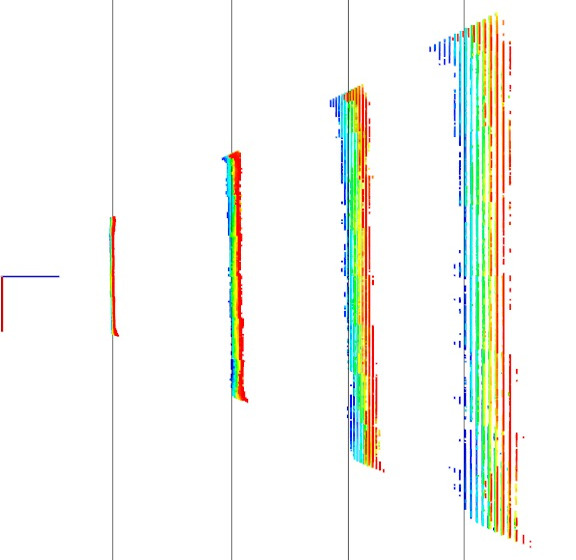}}\\
\subfloat[\textsc{asus} -- top]{\includegraphics[width=0.44\linewidth]{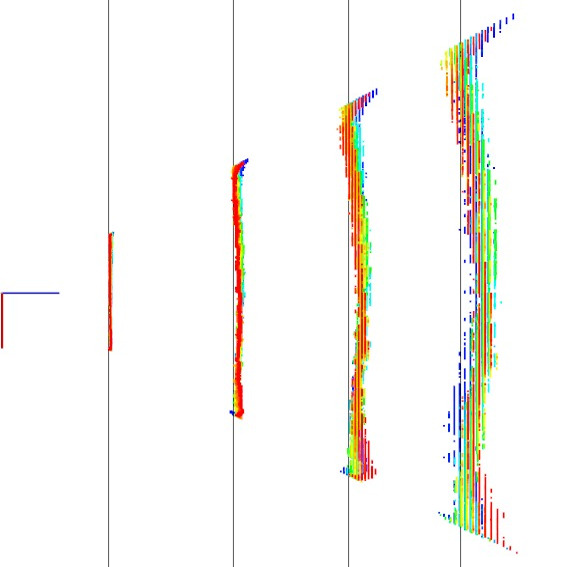}}
\subfloat[\textsc{kinect1b} -- side]{\includegraphics[width=0.44\linewidth]{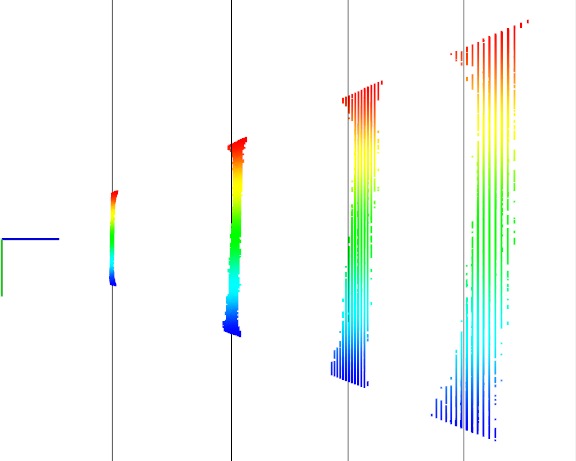}}
\caption[Top and side views of the point clouds generated by some different structured-light depth
         sensors.]
        {Top and side views of the point clouds generated by some different structured-light depth
         sensors: two Kinects (\textsc{kinect1a}, \textsc{kinect1b}) and an Asus Xtion Pro Live
         (\textsc{asus}). The gray lines represent the ground truth measured with the laser distance meters, points with different y-coordinates are drawn with different colors.}
\label{fig:depth_error1}
\end{figure}

\begin{figure}
\centering
\scriptsize
\subfloat[\textsc{kinect2} -- top]{\includegraphics[width=0.44\linewidth]{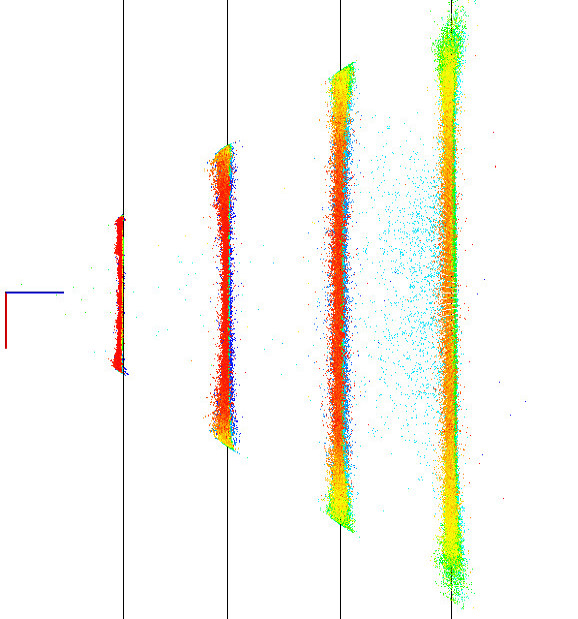}}
\subfloat[\textsc{kinect2} -- side]{\includegraphics[width=0.44\linewidth]{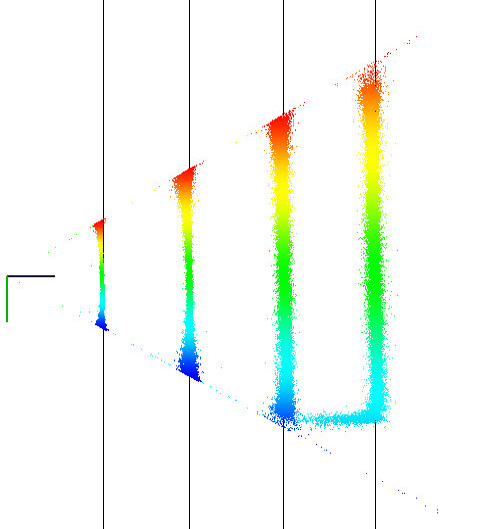}}\\

\caption{Top and side views of the point clouds generated by a Kinect 2 ToF camera (\textsc{kinect2}). The gray lines represent the ground truth measured with the laser distance meters, points with different y-coordinates are drawn with different colors. Note that the cloud on the right frames a part of the floor too.}
\label{fig:depth_error2}
\end{figure}

%


To analyze the systematic and random errors in the depth measurements, we positioned each sensor parallel to a flat wall at increasing distances. 
For each position, we collected the sensors readings (i.e., the depth images and the generated point clouds) while measuring the real distances (i.e., the ground truth) using two high precision laser distance meters (e.g., \figref{fig:setup_a}). Some qualitative results of such analysis are reported in \figref{fig:depth_error1} and \figref{fig:depth_error2}.
From our analysis, we noticed that:
\begin{enumerate}
 \item In the case of the SL sensors, the surfaces defined by the point clouds are not properly planar as they should be, and this ``local distortion'' effect becomes more accentuated for increasing distances (this is the \emph{myopic} property defined in  \cite{teichman2013unsupervised}). Moreover, each sensor has a different ``distortion pattern''. The ToF sensor is also affected by similar local distortion effect, but to a lesser extent (typically less than 1 cm).
 \item The average depth of each point cloud is not generally correct and, in the case of SL sensors, sometimes even the average orientation is wrong.
 \item The quantization error for the SL sensors becomes not negligible for increasing distances. ToF sensors are affected by a negligible quantization error\footnote{Continuous-wave based ToF sensors like the Kinect 2 measure the depth by means of a phase differences between an emitted sinusoidal light wave signal and the backscattered signals received. This phase is evaluated in closed-form using four equally spaced samples \cite{Horaud2016}: the quantization error, that is mainly due to the precision used in storing the samples, is usually neglected.}
\end{enumerate}

The effect of 1) is to produce a local alteration of an object shape, while 2) is a systematic bias in the measurements.

In this work we aim to remove both 1) and 2), while it is usually not possible to remove the quantization error 3) of SL sensors\footnote{In structured-light based depth sensors, the quantization error originates from the discrete nature of the disparity map used to extract the depth: this error is commonly mitigated by means of sub-pixel stereo matching algorithms, unfortunately these algorithms require to access low-level data that is usually not accessible from the user side.}.

In the following, we refer to these error components as \emph{1) distortion error} and \emph{2) global error}, respectively.
These errors in the case of SL sensors arise from a combination of two main sources: the radial and tangential lens distortion of the camera used for stereo triangulation \cite{khoshelham2012accuracy}, and the misalignment between the pattern projector and the camera.
In the case of ToF cameras, there are different sources of errors that contribute to the \emph{distortion error}, among others the built-in pixel-related errors and the amplitude-related errors, while the \emph{global error} arise from the so called \textit{wiggling} error that appears due to irregularities in the internal modulation process \cite{FoixSensors2011}. Our method does not directly address other ToF related sources of errors, such as multipath interference and reflectivity related deviations..\\

\begin{figure}
\centering
\scriptsize
\subfloat[Distortion]{
\begin{tikzpicture}
\begin{axis}[width=0.58\linewidth,
	xlabel={Ground truth depth [m]}, ylabel={RMS error [cm]},
	xmin=0.95, xmax=4.4,
	legend pos=north west,
	legend cell align=left
	]
	\addplot+[no marks] table[x index=0,y index=2] {data/distortion_error_47A.txt};
	\addplot+[no marks] table[x index=0,y index=2] {data/distortion_error_51A.txt};
	\addplot+[no marks] table[x index=0,y index=2] {data/distortion_error_asus.txt};
	\addplot+[no marks,green!60!black] table[x index=0,y index=2] {data/distortion_error_K2.txt};
	\addlegendentry{\textsc{kinect1a}};
	\addlegendentry{\textsc{kinect1b}};
	\addlegendentry{\textsc{asus}};
	\addlegendentry{\textsc{kinect2}};
\end{axis}
\label{fig:distortion_error}
\end{tikzpicture}}
\subfloat[Global error]{
\begin{tikzpicture}
\begin{axis}[width=0.58\linewidth,
	xlabel={Ground truth depth [m]}, ylabel={Distance error [cm]},
	xmin=0.95, xmax=4.4,
	legend pos=north west,
	legend cell align=left
	]
	\addplot+[no marks] table[x index=0,y index=2] {data/average_distance_error_47A.txt};
	\addplot+[no marks] table[x index=0,y index=2] {data/average_distance_error_51A.txt};
	\addplot+[no marks] table[x index=0,y index=2] {data/average_distance_error_asus.txt};
	\addplot+[no marks,green!60!black] table[x index=0,y index=2] {data/average_distance_error_K2.txt};
	\addlegendentry{\textsc{kinect1a}};
	\addlegendentry{\textsc{kinect1b}};
	\addlegendentry{\textsc{asus}};
	\addlegendentry{\textsc{kinect2}};
\end{axis}
\end{tikzpicture}
\label{fig:average_distance_error}}
\caption[Error on the depth estimation for four different depth sensors. (a) RMS error caused by the distortion; (b) Error on the average distance point estimation.]
        {Error on the depth estimation for four different depth sensors. (a) RMS error caused by the distortion.
         The error is computed by fitting a plane to the point cloud acquired in front of a flat
         wall and computing the point-plane distance for all its points. (a) Error on the average distance point estimation.
         The error is computed, for each cloud, by averaging the depth of all its points and
         comparing it to the ground truth computed with the two laser distance meters. }

\end{figure}
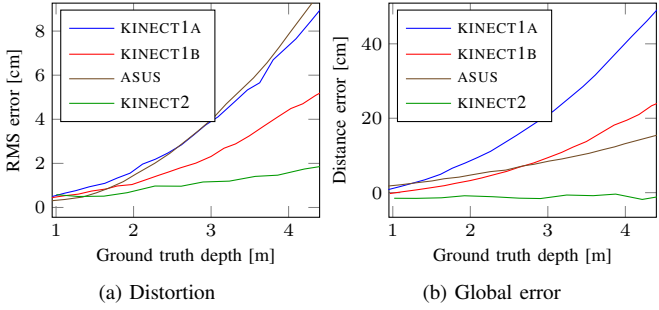

In order to analyze the distortion error trend, we compared the measured point clouds with the planes that best fit to them (some results are reported in \figref{fig:distortion_error}, where we used three SL sensors and one ToF sensor). In particular, for each incoming point cloud, we computed the Root Mean Square (RMS) error on the (signed) distance between the plane and the points. It can be noticed that for \textit{all} the tested sensors, such error is super-linear with respect to the measured depth values, despite the sources of error are different. It is also important to note the different error trends between two sensors of the same type (\textsc{kinect1a} and \textsc{kinect1b}): this is a further evidence that an effective calibration is an essential requirement for such sensors. In \figref{fig:average_distance_error} we reported the global error of the tested sensors for increasing distances, i.e. the difference between the average depth of the acquired clouds and the ground truth. It can be noticed that in the case of SL sensors, such error is super-linear with respect to the measured depth values, while in the case of a ToF sensor\footnote{The global error trend for the Kinect 2 sensor is plotted also in \figref{fig:G_final_results}, where a more suitable scale has been used.}, this error has a wiggling trend, confirming the results of other comprehensive ToF error analysis \cite{FoixSensors2011,7360902}

%

\subsection{Error Correction Model}

To model the effects of the errors introduced by depth sensors, as in \cite{herrera2012joint, canessa2014calibrated,
teichman2013unsupervised, DiCicco2015}, we propose to estimate
a depth correction function in a per-pixel basis.
That is, given a depth sensor $\obj{D}$ that provides a depth image $\I_\obj{D}$ of size $H_\obj{D} \times W_\obj{D}$, a pixel
$(u, v)^\T$ and the corresponding depth value $z = \I_\obj{D}(u, v)$, the real depth $z^*$ is computed as:
\begin{equation}
 z^* = \f{f}_{u, v}(z).
 \label{eq:f_u_v}
\end{equation}
$z$ represents both the depth measured by the sensor and the $z$-coordinate of the corresponding 3D point perceived by the depth camera, in a reference system with the $z$ axis corresponding to the optical axis of the camera.\\ Starting from the considerations made above, we express each
$\f{f}_{u, v}(\cdot)$ in \eqref{eq:f_u_v} as a composition of two functions:
$\f{u}_{u, v}(\cdot)$ that takes into account the local distortion 1), and $\f{g}_{u, v}(\cdot)$
that makes a global correction 2) of the depth values.
That is, the real depth $z^*$ is estimated as:
\begin{equation}
 z^* = \f{f}_{u, v}(z) = (\f{g} \circ \f{u})_{u, v}(z),
 \label{eq:f_comp_u_v}
\end{equation}
or, alternatively, given the 3D point $\pp = (x, y, z)^\T$ associated with the pixel $(u, v)^\T$, the real 3D point $\pp^*$ is estimated as:
\begin{equation*}
 \pp^* = \dot{\f{f}}_{u, v}(\pp) = (\dot{\f{g}} \circ \dot{\f{u}})_{u, v}(\pp),
\end{equation*}
where
\begin{equation*}
 \dot{\f{f}}_{u, v}(\pp) = \pp \cdot \frac{\f{f}_{u, v}(z)}{z}.
\end{equation*}
%


We define $\U$ as the map that associates a pixel $(u, v)^\T$ to an undistortion 
function $\f{u} : \R \to \R$, that is, $\U(u, v) = \f{u}_{u, v}(\cdot)$.
In the same way, we define $\G(u, v) = \f{g}_{u, v}(\cdot)$ as the map that associates a pixel $(u, v)^\T$ to a function of the depth that correct the global error.


%% file: approach.tex
\section{Calibration Approach} \label{sec:calibration_algorithm_approach}

As confirmed by the experimental evidence (see \sectref{sec:depth_error_model}), the error on the depth
measurements is a smooth function.
Thus we can assume that given two close 3D points $\pp$ and $\pq$ along the
same direction, i.e $\pq = (1 + \varepsilon) \cdot \pp$ with $\varepsilon \simeq 0$,
\begin{equation*}
 \pq^* = \dot{\f{f}}(\pq) = \dot{\f{f}}((1 + \varepsilon) \cdot \pp) \simeq (1 + \varepsilon) \cdot \dot{\f{f}}(\pp) = (1 + \varepsilon) \cdot \pp^*.
\end{equation*}
where $\dot{\f{f}}(\cdot)$ is the error correction function defined in Eq.~\ref{eq:f_comp_u_v}.
This means that, if we know how to ``correct'' a point $\pp$ (i.e. we know the correction function parameters for this point), we can correct close points with a good approximation using the same parameters.

This assumption is the basis of our algorithm to estimate both the undistortion map $\U$
and the global error correction map $\G$.
Exploiting the fact that both distortion and quantization errors become more severe for 
increasing distances, we introduce the idea to estimate the distortion error iteratively, 
starting from short distances and estimating the error for greater distances using as initial
guess the current correction parameters.

The proposed calibration framework requires the depth sensor to be coupled with a \textit{calibrated} RGB camera that
frames approximately the same scene: the rigid body transformation that relates the two sensors will be estimated
while inferring the depth error correction function.
It also requires the two sensors to collect data framing a scene that includes a wall with a
checkerboard attached on it, at different distances and orientations.

The calibration is performed in two steps: in the first step the algorithm estimates the undistortion map $\U$; only a rough calibration between the camera and the depth sensor is necessary during this step, 
the checkerboard is used just to have an idea of the wall location.
In the second step the global correction map $\G$ is computed. 
Here the checkerboard poses estimated with the (calibrated) RGB camera are used as a ground truth.
That is, the undistorted planes estimated in the first step are forced to match the ones defined by
the checkerboard.
To this end, the real rigid displacement between the RGB camera $\obj{C}$ and the depth sensor $\obj{D}$
needs to be known.
Unfortunately, to estimate the pose of one sensor with respect to the other, a good estimate of their intrinsic parameters is mandatory.
One way to satisfy this circular dependency is to estimate the global correction map $\G$ and the
rigid body displacement $\TF{\matx{T}}{D}{C}$ simultaneously.

At this point a question arises: why the depth error is corrected in two steps?\\
Actually, the reason is simple.
To guarantee the best results, the camera-depth sensor transformation $\TF{\matx{T}}{D}{C}$ and the
global correction map $\G$ need to be refined together within an optimization framework.
Refine inside the same framework a map such as $\U$, with a different function every pixel, is not a feasible solution.
On the other hand, the map $\G$, whose scope is to transform planes into planes, is defined by a dozen parameters (see \sectref{sec:global_map_estimation}), thus better suited to be efficiently estimated inside an optimization scheme.

\subsection{Pipeline}

\begin{figure}

\tikzstyle{main} = [text width=1.75cm, text centered, minimum height=0.75cm, minimum width=1.75cm]
\tikzstyle{block} = [main, rectangle, draw, fill=cyan!30, rounded corners]
\tikzstyle{block2} = [main, rectangle, draw, fill=yellow!30, rounded corners]
\tikzstyle{data} = [main, rectangle, draw, fill=green!30]
\tikzstyle{empty} = [main]
\tikzstyle{arrow} = [draw, ->, thick]
\tikzstyle{double_arrow} = [draw, -implies, double equal sign distance, thick]

\footnotesize
\centering
\begin{tikzpicture}

\def\d{0.85cm};

\node [data] (rgb) {RGB Images};
\node [empty] (initial_T) [right=\d of rgb] {};
\node [data] (depth-image) [right=\d of initial_T] {Depth Images};

\node [block] (und_est) [below=\d of initial_T] {Undistortion Map Estimation};
\node [block2] (corners) [below=\d of rgb] {Checkerboard Corners Extraction};
\node [block2] (point-cloud) [below=0.975cm of depth-image] {Point Cloud Generation};
\node [block2] (und) [below right=0.4cm and \d of und_est] {Point Cloud Undistortion};

\node [block, text width=2.5cm, minimum height=1cm, minimum width=2.5cm] (final) [below left=0.25cm and \d-0.375cm of und] {Extrinsic Parameters \& Global Error Correction Function Estimation};
\node [] (UG) [below=\d of final] {};

\path [double_arrow] (rgb) -- node[anchor=east] {$\I_{\obj{C},k}$} (corners);
\path [double_arrow] (corners) -- node[anchor=south] {$\TF{\map{B}_k}{}{\I_C}$} (und_est);
\path [double_arrow] (point-cloud) -- node[anchor=south] {$\TF{\map{C}_k}{}{D}$} (und_est);
\path [double_arrow] (depth-image) -- node[anchor=east] {$\I_{\obj{D},k}$} (point-cloud);
\path [arrow] (initial_T) -- node[anchor=west] {$\TF{\matx{T}}{D}{C}_0$} (und_est);

\path [double_arrow] (und_est) -- node[anchor=east] {$\mathbb{I}_k$} (final);

\node [empty] (tmp) [below right=-0.775cm and 1cm of corners] {};
\path [arrow] (tmp) |- node[anchor=south west] {$\U$} (und);

\path [double_arrow] (und) |- node[anchor=north west] {$\TF{\widehat{\map{C}}_k}{}{D}$} (final);
\path [double_arrow] (corners) |- node[anchor=north east] {$\TF{\map{B}_k}{}{\I_C}$} (final);
\path [double_arrow] (point-cloud) -- node[anchor=east] {$\TF{\map{C}_k}{}{D}$} (und);
\path [arrow] (final) -- node[anchor=west] {$\G$} node[anchor=east] {$\TF{\matx{T}}{D}{C}$} (UG);

\end{tikzpicture}
\caption[Calibration algorithm pipeline.]
        {Calibration algorithm pipeline. Double-lined arrows mean that a set of data is passed from 
         one block to the other.}
\label{fig:pipeline}

\end{figure}
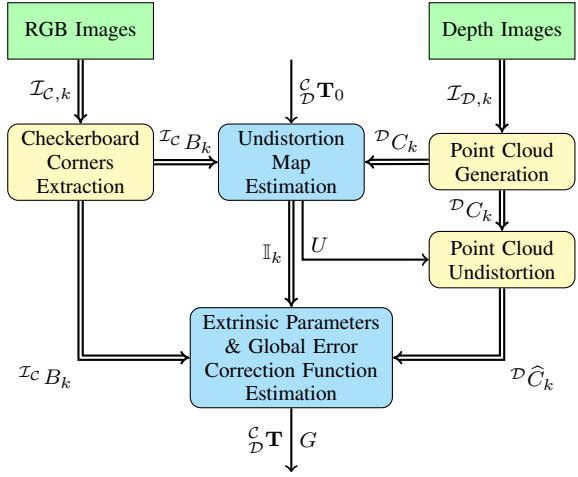

The algorithm is organized as in \figref{fig:pipeline}.
First of all, the checkerboard corners are extracted from all the collected RGB images $\I_{\obj{C},k}$ and the organized point clouds\footnote{An organized point cloud $\TF{\map{C}}{}{D}$  is a point cloud that reflects the depth image structure, i.e. the points are organized into rows and columns as the depth image, $z = \I_\obj{D}(u, v) \Rightarrow \pp = \TF{\map{C}}{}{D}(u,v)$.} are generated from the depth images $\I_{\obj{D},k}$, $k = 1, 2, \dots, M$.
The corners $\TF{\map{B}_k}{}{\I_C}$ (in pixel coordinates), the point clouds $\TF{\map{C}_k}{}{D}$ (in depth sensor reference frame) and the initial 
camera-depth transformation $\TF{\matx{T}}{D}{C}_0$ are the inputs for the undistortion map estimation
module.
Once the undistortion map $\U$ has been estimated, the point clouds are undistorted
($\TF{\widehat{\map{C}}_k}{}{D}$) and passed to the module that estimates both the
global correction map $\G$ and the final camera-depth transformation $\TF{\matx{T}}{D}{C}$.

%% file: undistortion_map.tex
\section{Undistortion Map Estimation}
\label{sec:undistortion_map_estimation}
The proposed algorithm (\algref{alg:undistortion_map_creation}) estimates the undistortion map $\U$ 
taking as input a list of point clouds $\{\TF{\map{C}_1}{}{D}, \TF{\map{C}_2}{}{D}, \dots, \TF{\map{C}_M}{}{D}\}$ acquired when the depth sensor $\obj{D}$ is pointing a planar surface (e.g. a wall) at different distances and orientations.
It also requires the positions of the checkerboard corners $\{\TF{\map{B}_1}{}{\I_C}, \TF{\map{B}_2}{}{\I_C}, \dots, \TF{\map{B}_M}{}{\I_C}\}$, extracted from the images, and a rough estimate of the rigid-body transformation that relates the two sensors $\TF{\matx{T}}{D}{C}_0$. The data structure $\E_{\U}$ used in \algref{alg:undistortion_map_creation} is a matrix of \textit{sample sets} (one for each sensor pixel) that keeps in memory the samples used to fit the undistortion functions $\f{u}_{u, v}(\cdot)$.


%
\begin{figure}
\small
\begin{algorithmic}[1]
\Require $\{\TF{\map{C}_1}{}{D}, \TF{\map{C}_2}{}{D}, \dots, \TF{\map{C}_M}{}{D}\}$ \Comment Point clouds
\Require $\{\TF{\map{B}_1}{}{\I_C}, \TF{\map{B}_2}{}{\I_C}, \dots, \TF{\map{B}_M}{}{\I_C}\}$ \Comment Checkerboard corners
\Require $\TF{\matx{T}}{D}{C}_0$ \Comment Camera-depth sensors initial transformation
\Ensure $\U$ \Comment Undistortion map
\State $\U \gets \mathfrak{1}(\cdot)$ for all $(u, v)^\T \in \I_{\obj{D}}$
\State $\E_{\U} \gets \{ \emptyset \}$ for all $(u, v)^\T \in \I_{\obj{D}}$ 
\State $\{\TF{\map{C}_{s_1}}{}{D}, \TF{\map{C}_{s_2}}{}{D}, \dots, \TF{\map{C}_{s_M}}{}{D}\} \gets \Call{sort}{\{\TF{\map{C}_1}{}{D}, \TF{\map{C}_2}{}{D}, \dots, \TF{\map{C}_M}{}{D}\}}$
\For{$k \gets 1, 2, \dots, M$}
  \ForAll{$(u, v)^\T \in \I_{\obj{D}}$}
    \State $\f{u}_{u, v}(\cdot) \gets \U(u, v)$ \Comment Current undistortion function
    \State $\TF{\pp}{}{D} \gets \TF{\map{C}_{s_k}(u, v)}{}{D}$ 
    \State $z \gets \I_{\obj{D},s_k}(u, v)$
    \State $\TF{\widehat{\map{C}}_{s_k}(u, v)}{}{D} \gets \dot{\f{u}}(\TF{\pp}{}{D}) = \TF{\pp}{}{D} \cdot \frac{\f{u}_{u, v}(z)}{z}$
  \EndFor
  \State $\mathbb{I}_{s_k} \gets \Call{selectWallPoints}{\TF{\widehat{\map{C}}_{s_k}}{}{D}, \TF{\map{B}_{s_k}}{}{\I_C}, \TF{\matx{T}}{D}{C}_0}$
  \State $\TF{\pi_{s_k}}{}{D} \gets \Call{fitPlane}{\TF{\map{C}_{s_k}}{}{D}, \mathbb{I}_{s_k}}$ 
  \State $(\U, \E_{\U}) \gets \Call{updateMap}{\U, \E_{\U}, \TF{\map{C}_{s_k}}{}{D}, \mathbb{I}_{s_k}, \TF{\pi_{s_k}}{}{D}}$
\EndFor
\end{algorithmic}
\caption{Pseudocode of the algorithm developed to estimate the undistortion map $\U$.}
\label{alg:undistortion_map_creation}
\end{figure}

Firstly (\alglines{1-2}) the undistortion map $\U$ is initialized as an $H_\obj{D} \times 
W_\obj{D}$ matrix of identity functions $\mathfrak{1}(\cdot) : \R \to \R$, while the sample matrix
$\E_{\U}$ is initialized as an $H_\obj{D} \times W_\obj{D}$ matrix of empty sets.
Then, the point cloud list $\{\TF{\map{C}_1}{}{D}, \TF{\map{C}_2}{}{D}, \dots, \TF{\map{C}_M}{}{D}\}$
is sorted in ascending order, $\{\TF{\map{C}_{s_1}}{}{D}, \TF{\map{C}_{s_2}}{}{D}, \dots, \TF{\map{C}_{s_M}}{}{D}\}$, according to the distance of the main plane (i.e., the plane with the checkerboard) from the sensor (\algline{3}), to exploit the smoothness described in \sectref{sec:calibration_algorithm_approach}.

Actually, no plane extraction is performed, the assumption that the RGB camera and the depth sensor are facing the same wall is exploited to sort the point clouds. That is, the checkerboard corners provided in input are used.

The undistortion map is created iteratively: at each step only one point cloud is analyzed (\alglines{4}).
At step $k$, the $k^\mathrm{th}$ cloud is undistorted using the \emph{current} estimation of $\U$ (\alglines{5-9}).
The coordinates $\mathbb{I}_{s_k} = \{(u,v)^\T_{s_k,1}, (u,v)^\T_{s_k,2},\dots,(u,v)^\T_{s_k,n_{s_k}}\}$ of the $n_{s_k}$ points that lie to the main plane are extracted from the undistorted cloud (\alglines{11}) as described in \sectref{sec:undistortion_map_wall_points_selection}, A plane $\TF{\pi_{s_k}}{}{D}$ is then fitted to the initial cloud $\TF{\map{C}_{s_k}}{}{D}$ (\alglines{12}) by using \textit{only} the points selected in \alglines{11}.
Actually, to increase stability, instead of fitting a plane to the whole original point cloud, only the pixels within a defined ray from the plane center are used, as reported in \cite{DiCicco2015}. Finally, the estimated plane $\TF{\pi_{s_k}}{}{D}$ is used to update the undistortion map $\U$ (\algline{13}) as described in \sectref{sec:undistortion_map_update}.
The procedure ends as soon as the last cloud in the list has been processed.

\subsection{Wall Points Selection}
\label{sec:undistortion_map_wall_points_selection}

\begin{figure}
\centering
\subfloat[Wall points extracted (in orange) from the original point cloud $\TF{\map{C}_{s_k}}{}{D}$.]{\includegraphics[width=0.95\linewidth]{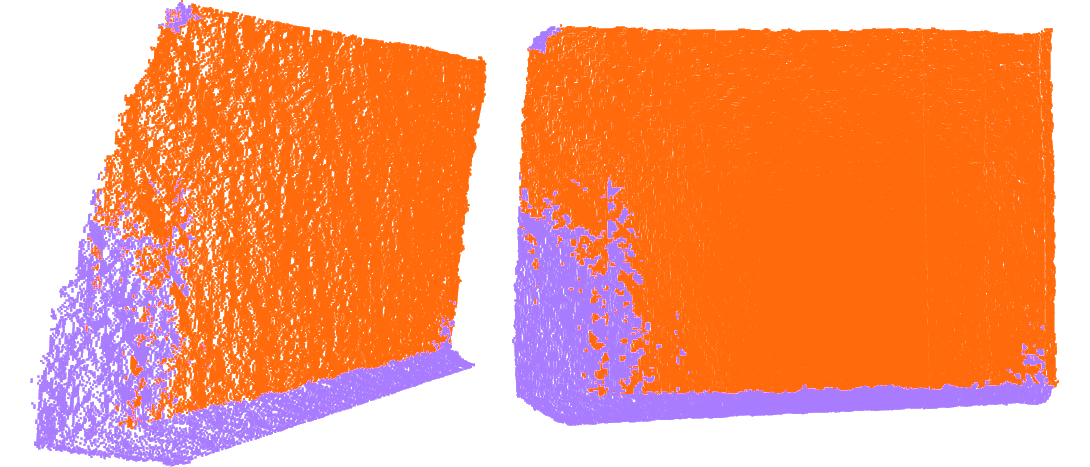}}\\
\subfloat[Wall points extracted (in pink) from the undistorted point cloud $\TF{\widehat{\map{C}}_{s_k}}{}{D}$.]{\includegraphics[width=0.95\linewidth]{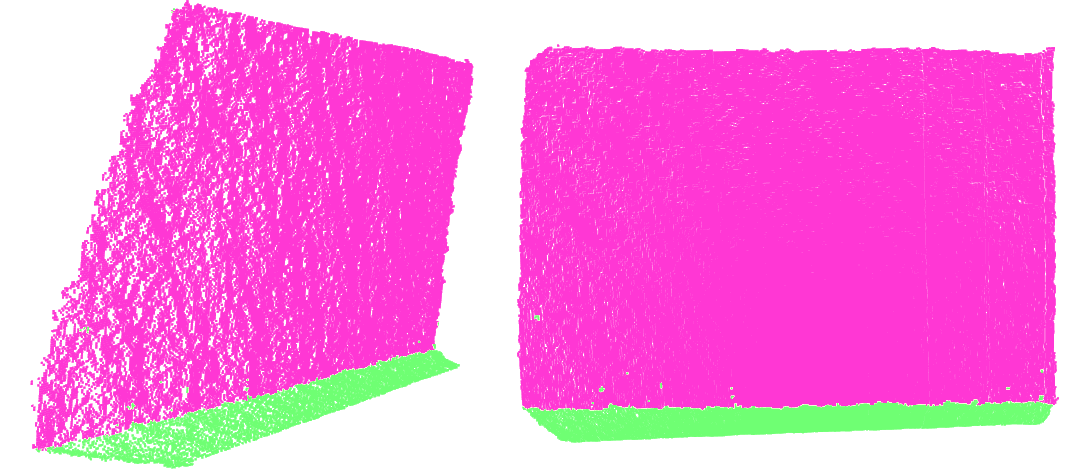}}
\caption[Comparison between the wall points extracted from the original point cloud e the ones extracted from the undistorted cloud]
        {Comparison between (a) the wall points extracted from the original point cloud
         $\TF{\map{C}_{s_k}}{}{D}$ and (b) the ones extracted from the undistorted cloud
         $\TF{\widehat{\map{C}}_{s_k}}{}{D}$.
         As clearly visible, in both cases the floor points are correctly discarded. In the former
         case, however, the wall segmentation is wrong.}
\label{fig:wall_points_extraction}
\end{figure}

The selection of the wall point-coordinates $\mathbb{I}_{s_k}$ is performed automatically, as opposed to the manual
selection of \cite{herrera2012joint}.
We take advantage of the RGB camera and the checkerboard to select the right plane and extract
the coordinates from the undistorted cloud with a RANSAC-based approach \cite{fischler1981random, pcl}.
As the example shown in \figref{fig:wall_points_extraction}, the undistorted cloud lets us
extract the correct points, where the original one does not. In particular, points near the image corners are likely to be excluded from the inliers when using the original cloud.

\subsection{Map Update}
\label{sec:undistortion_map_update}
In the map update function (\algref{alg:undistortion_map_update}), all the points
$\TF{\pp}{}{D} = (x, y, z)^\T$ of the cloud that lie to the main plane are projected on the previously extracted plane $\TF{\pi_{s_k}}{}{D}$ along their line-of-sight (\alglines{3-4}, where \textsc{losProject()} is a short name for the function \textsc{lineOfSightProject()}).
That is, let $\vect{n}^\T \vect{x} - d_\pi = 0$ be the plane equation, with $\vect{n}$ the plane unit normal vector and $d_\pi$ the distance of the plane from the origin, and let $l \TF{\pp}{}{D}$, $l \in \R$, be the points along $\TF{\pp}{}{D}$-line-of-sight, then the line-of-sight projection of $\TF{\pp}{}{D}$ onto $\TF{\pi_{s_k}}{}{D}$, say $\TF{\pp_\pi}{}{D} = (x_\pi, y_\pi, z_\pi)^\T$, is:
\begin{equation*}
  \pp_\pi = l \pp =  \frac{d_\pi \pp}{\vect{n}^\T \pp} \enspace.
\end{equation*}
where the superscript $\obj{D}$ has been omitted for the sake of simplicity. The pair $(z, z_\pi)$ is used as a sample for the curve-fitting procedure (\algline{5}), and the undistortion function $\U(u, v)$ is re-estimated by fitting a new curve to the sample 
set $\E_{\U}(u, v)$ (\algline{6}).
\begin{figure}
\small
\begin{algorithmic}[1]
\Function{updateMap}{$\U, \E_{\U}, \TF{\map{C}}{}{D}, \mathbb{I}, \TF{\pi}{}{D}$}
  \ForAll{$(u, v)^\T \in \mathbb{I}$}
    \State $\TF{\pp}{}{D} \gets \TF{\map{C}(u, v)}{}{D}$
    \State $\TF{\pp_\pi}{}{D} \gets \Call{losProject}{\TF{\pp}{}{D}, \TF{\pi}{}{D}}$
    \State $\E_{\U}(u, v) \gets \E_{\U}(u, v) \cup \{(z, z_\pi)\}$
    \State $\U(u, v) \gets \Call{fitCurve}{\E_{\U}(u, v)}$
  \EndFor
  \State \Return $(\U, \E_{\U})$
\EndFunction
\end{algorithmic}
\caption{Pseudocode of the algorithm for updating the undistortion map $\U$.}
\label{alg:undistortion_map_update}
\end{figure}

\subsection{Implementation Details}

\subsubsection{Undistortion Map}
\label{sec:undistortion_map_implementation}

To decrease the incidence of noise on the map estimation we reduce the number of
functions fitted to the data.
That is, instead of estimating an undistortion function for each pixel, similarly to
\cite{teichman2013unsupervised}, we discretize the map into bins.
So, let $\chi_\U, \psi_\U \in \N$ be the bin size in pixels, along the image $x$- and $y$-direction, respectively.

Given a pixel $(u, v)^\T \in \I_\obj{D}$, we define $\map{S}_\U(u, v)$ as the set of 4 pixels
\emph{surrounding} $(u, v)^\T$ according to the sampling factors $\chi_\U$ and $\psi_\U$ (see
\figref{fig:undistortion_map}).
We also define $\SU \triangleq \{ \map{S}_\U(u, v) : (u, v)^\T \in \I_\obj{D} \}$ as the set of all
the \emph{surrounding pixels}.

We estimate the undistortion function $\f{u}_{u, v}(\cdot)$ only for the pixels in $\SU$. 
For all the others, instead, this function is computed as a linear combination of the functions 
computed for the pixels set $\SU$.
That is, given a pixel $(u, v)^\T$, following a bilinear interpolation approach, its undistortion function $\U(u, v)$ can be computed as:
\begin{equation*}
  \U(u, v) = \sum_{(s, t) \in \map{S}_\U(u, v)} \f{w}_{\chi_\U}(u, s) \cdot \f{w}_{\psi_\U}(v, t) \cdot \U(s, t)
\end{equation*}
where
\begin{equation}
  \f{w}_{\chi_\U}(u, s) \triangleq 1 - \frac{|u - s|}{\chi_\U}, \quad \f{w}_{\psi_\U}(v, t) \triangleq 1 - \frac{|v - t|}{\psi_\U}
  \label{eq:weight}
\end{equation}
and
\begin{equation*}
  \sum_{(s, t) \in \map{S}_\U(u, v)} \f{w}_{\chi_\U}(u, s) \cdot \f{w}_{\psi_\U}(v, t) = 1 \enspace.
\end{equation*}
%

%
\begin{figure}
\tikzstyle{undistortion} = [cyan!90!black]
\footnotesize
\centering
\begin{tikzpicture}[scale=0.30]

\begin{scope}[shift={(-0.8, 0)}]
\draw[thin] (-0.3, 0) -- (0.3, 0);
\draw[thin] (-0.3, -4) -- (0.3, -4);
\draw[thin] (0, 0) -- node[fill=white]{$\psi_\U$} (0, -4);
\end{scope}

\begin{scope}[shift={(-1.6, 0)}]
\draw[thin] (-0.3, 0) -- (0.3, 0);
\draw[thin] (-0.3, -12) -- (0.3, -12);
\draw[thin] (0, 0) -- node[fill=white]{$H_\obj{D}$} (0, -12);
\end{scope}

\begin{scope}[shift={(0, 0.8)}]
\draw[thin] (0, 0.3) -- (0, -0.3);
\draw[thin] (4, 0.3) -- (4, -0.3);
\draw[thin] (0, 0) -- node[fill=white]{$\chi_\U$} (4, 0);
\end{scope}

\begin{scope}[shift={(0, 1.6)}]
\draw[thin] (0, 0.3) -- (0, -0.3);
\draw[thin] (16, 0.3) -- (16, -0.3);
\draw[thin] (0, 0) -- node[fill=white]{$W_\obj{D}$} (16, 0);
\end{scope}

\draw[step=1,gray,very thin] (0, -12) grid (16, 0);
\draw[black,very thick] (0, 0) rectangle (16, -12);

\begin{scope}[shift={(0.5, -0.5)}]
\foreach \i in {0,4,...,16}
{
  \foreach \j in {0,4,...,12}
  {
    \filldraw[undistortion,thick] (\i-0.25, -\j-0.25) rectangle (\i+0.25, -\j+0.25);
    \node[minimum size=0.2cm] (point_\i_\j) at (\i, -\j) {};
  }
}
\draw[undistortion,step=4,thin,dashed] (0, -12) grid (16, 0);

%

\filldraw[undistortion,thick,fill=white] (6.75, -6.75) rectangle (7.25, -7.25);
\node[minimum size=0.2cm] (pixel1) at (7, -7) {};
\draw[->,thick] (pixel1) -- (point_4_4);
\draw[->,thick] (pixel1) -- (point_4_8);
\draw[->,thick] (pixel1) -- (point_8_4);
\draw[->,thick] (pixel1) -- (point_8_8);

\filldraw[undistortion,thick,fill=white] (13.75, -3.75) rectangle (14.25, -4.25);
\node[minimum size=0.2cm] (pixel2) at (14, -4) {};
\draw[->,thick] (pixel2) -- (point_12_4);
\draw[->,thick] (pixel2) -- (point_16_4);
\draw[->,thick] (pixel2) -- (point_12_8);
\draw[->,thick] (pixel2) -- (point_16_8);

\end{scope}
\end{tikzpicture}
\vspace{0.5cm}\\
\begin{tikzpicture}[scale=0.35]

\filldraw[undistortion,thick] (-0.25, -0.25) rectangle (0.25, 0.25);
\node[anchor=west,text width=0.9\linewidth] at (0.5, -0.1) {Pixel $(u, v)^\T \in \SU$. For this pixel an undistortion function $\f{u}_{u, v}(\cdot)$ has been estimated.};

\filldraw[undistortion,thick,fill=white] (-0.25, -2.65) rectangle (0.25, -2.15);
\node[anchor=west,text width=0.9\linewidth] at (0.5, -2.5) {Pixel $(u, v)^\T \in \I_\obj{D}$. For this pixel the undistortion function $\f{u}_{u, v}(\cdot)$ is a linear combination of the functions of the pixels in $\map{S}_\U(u, v)$.};

\draw[->,thick] (-0.3, -4.4) -- (0.4, -4.4);
\node[anchor=west,text width=0.9\linewidth] at (0.5, -4.5) {Connection from a pixel $(u, v)^\T \in \I_\obj{D}$ to one in $\map{S}_\U(u, v)$.}; 

\end{tikzpicture}
\caption[Scheme of an undistortion map $\U$.]
        {Visualization of a discretized undistorted map $\U$. Given two parameters, $\chi_\U$ and $\psi_\U$,
        an undistortion function is estimated for all and only the pixels in $\SU$. For all the
        others, instead, the function is computed as a linear combination of the neighboring points in $\SU$.}
\label{fig:undistortion_map}
\end{figure}

\subsubsection{Curve Fitting}
\label{sec:curve_fitting_U}

As shown in \sectref{sec:depth_error_model} (\figref{fig:distortion_error}), the distortion is
super-linear, therefore an appropriate correction function must be chosen.
Moreover, as previously described, since we are not estimating a function for
every pixel, the fitting procedure is not straightforward.

For what concerns the former point, suppose the error is corrected by a second degree polynomial,
that is, $\U(u, v) = \f{u}_{u, v}(z) = a + bz + cz^2$, for some $a, b, c \in \R$.
To estimate the polynomial coefficients we solve a non-linear least squares problem of the form:
\begin{equation*}
  \arg \min_{a, b, c} \sum_{(z, z_\pi) \in \E_{\U}(u, v)} \frac{1}{\sigma^2(z)} \left\Vert a + bz+ cz^2 - z_\pi \right\Vert^2
\end{equation*}
where $\sigma(z)$ is a normalization term, that is, the error on the depth measurements.
For the Kinect 1 we choose $\sigma(z)$ to be the quantization error (as reported in \cite{smisek20113d}), i.e.
\begin{equation*}
  \sigma(z) = -0.00029 + 0.00037 \cdot z + 0.001365 \cdot z^2.
\end{equation*}
For Kinects 2, since we didn't find any suitable equation in the literature, we fitted a second degree polynomial to the samples reported in \figref{fig:distortion_error}, i.e.
\begin{equation*}
  \sigma(z) = 0.00313 + 0.00116 \cdot z + 0.00052 \cdot z^2.
\end{equation*}
For what concerns the latter point, i.e. how to deal with the discretized undistortion map, we 
slightly modify the sample set generation and the function fitting procedure described in
\algref{alg:undistortion_map_update}.
In the new algorithm (\algref{alg:undistortion_map_creation_2}), every pixel $(u, v)^\T$
contributes to the sample set of its four surrounding pixels $\map{S}_\U(u, v)$ with a weight
calculated as in \eqref{eq:weight}.
That is, let:
\begin{equation*}
  \map{S}^{-1}_\U(s, t) \triangleq  \{ (u, v)^\T \in \I_\obj{D} : (s, t) \in \map{S}_\U(u, v) \}
\end{equation*}
be the set of pixels which have $(s, t)$ as one of their surrounding pixels.
For each cloud, the temporary sample set $\E_w(s, t)$ for a pixel $(s, t) \in \SU$, is
(\alglines{2-10}):
\begin{equation*}
  \E_w(s, t) \triangleq  \bigcup_{(u, v)^\T \in \map{S}^{-1}(s, t)} (w, z, z_\pi)
\end{equation*}
where
\begin{equation*}
  w \triangleq  \f{w}_{\chi_\U}(u, s) \cdot \f{w}_{\psi_\U}(v, t).
\end{equation*}
$\E_w(s, t)$ is used to generate the sample set for the aforementioned curve fitting
procedure (\alglines{11-15}).
Basically, the pair $(\bar{z}, \bar{z}_\pi)$ is calculated as the weighted arithmetic mean of the values in $\E_w(s, t)$, that is:
\begin{align*}
    W           &\triangleq              \sum_{(w, z, z_\pi) \in \E_w(u, v)} w, \\
    \bar{z}     &\triangleq  \frac{1}{W} \sum_{(w, z, z_\pi) \in \E_w(u, v)} w \cdot z, \\
    \bar{z}_\pi &\triangleq  \frac{1}{W} \sum_{(w, z, z_\pi) \in \E_w(u, v)} w \cdot z_\pi,
\end{align*}
and added to the sample set $\E_{\U}(s, t)$.

\begin{figure}
\small
\begin{algorithmic}[1]
\Function{updateMap}{$\U, \E_{\U}, \TF{\map{C}}{}{D}, \mathbb{I}, \TF{\pi}{}{D}$}
  \State $\E_w \gets \{ \emptyset \}$ for all $(u, v)^\T \in \I_{\obj{D}}$
  \ForAll{$(u, v)^\T \in \mathbb{I}$}
    \State $\TF{\pp}{}{D} \gets \TF{\map{C}(u, v)}{}{D}$
    \State $\TF{\pp_\pi}{}{D} \gets \Call{losProject}{\TF{\pp}{}{D}, \TF{\pi}{}{D}}$
    \ForAll{$(s, t) \in \map{S}_\U(u, v)$}
      \State $w \gets \f{w}_{\chi_\U}(u, s) \cdot \f{w}_{\psi_\U}(v, t)$
      \State $\E_w(s, t) \gets \E_w(s, t) \cup \{(w, z, z_\pi)\}$
    \EndFor
  \EndFor
  \ForAll{$(s, t) \in \SU$}
    \State $(\bar{z}, \bar{z}_\pi) \gets \Call{weightedMean}{\E_w(s, t)}$
    \State $\E_{\U}(s, t) \gets \E_{\U}(s, t) \cup \{(\bar{z}, \bar{z}_\pi) \}$
    \State $\U(s, t) \gets \Call{fitCurve}{\E_{\U}(s, t)}$
  \EndFor
  \State \Return $(\U, \E_{\U})$
\EndFunction
\end{algorithmic}
\caption{Pseudocode of the algorithm for updating the undistortion map $\U$ taking into account the pixel binning.}
\label{alg:undistortion_map_creation_2}
\end{figure}

%% file: global_map.tex
\section{Global Correction Map Estimation} \label{sec:global_map_estimation}

Our original solution to deal with the global, systematic error was to have a unique function, say
$\f{g}(\cdot)$, to correct the wrong depth measurements after the undistortion phase, i.e.
$\G(u, v) = \f{g}(\cdot)$, for all $(u, v)^\T \in \I_\obj{D}$.
Unfortunately, this solution had one important limitation: in some cases the undistorted clouds
were both translated and rotated around a non-predictable axis. 
For this reason we moved to a correction map $\G$ someway similar to the previously described
undistortion map $\U$.
The actual implementation of such map is described in \sectref{sec:global_map_implementation}.
Our algorithm takes as input a set of already undistorted point clouds
$\{\TF{\widehat{\map{C}}_1}{}{D}, \TF{\widehat{\map{C}}_2}{}{D}, \dots,
\TF{\widehat{\map{C}}_M}{}{D}\}$, the correspondent wall point coordinates $\{\mathbb{I}_1, \mathbb{I}_2, \dots, \mathbb{I}_M\}$
and the checkerboard corners extracted from the RGB images, $\{\TF{\map{B}_1}{}{\I_C}, \TF{\map{B}_2}{}{\I_C},
\dots, \TF{\map{B}_M}{}{\I_C}\}$.
After an initialization step where a rough estimate of the map $\G$ is computed
(\sectref{sec:global_map_initialization}), the map is refined, along with the camera-depth sensor
transformation $\TF{\matx{T}}{C}{D}$, within a non-linear optimization framework
(\sectref{sec:global_map_optimization}).


\subsection{Initial Estimation}
\label{sec:global_map_initialization}

The algorithm used for the initial estimation of the map functions as well as the computation of the rigid transform
between the RGB and the depth sensor is reported in \algref{alg:global_map_estimation}.
Firstly, the pose of one sensor with respect to the other is estimated, that is, for each
color-depth image pair both the plane defined by the checkerboard in the image
$\TF{\pi_{\map{B}_k}}{}{C}$ (in the RGB camera $\obj{C}$ reference frame), and the one extracted from the point cloud
$\TF{\pi_{\widehat{\map{C}}_k}}{}{D}$ (in the depth sensor $\obj{D}$ reference frame) are computed from the given
input data (\alglines{1-6}).
The transformation between the checkerboard and the RGB camera ($\tf{\matx{T}}{\fr{B}_k}{\fr{C}}$) is estimated given the checkerboard 3D points $\TF{\map{B}}{}{B}$, their corresponding image projections $\TF{\map{B}_k}{}{I_C}$, the camera matrix $\matx{K}_\obj{C}$ and the distortion coefficients $\vect{dist}_\obj{C}$ using an iterative optimization based on the Levenberg-Marquardt method (OpenCV \cite{KaehlerBradski2014} function \emph{solvePnP}, \algline{2}, that solves a Perspective-n-Point problem)
The checkerboard 3D points are then transformed in the RGB camera frame (\algline{3}). The equation of the plane framed by 
the RGB camera is hence computed taking 3 non-collinear corners (\algline{4}).
The equation of the plane in the depth image, instead, is computed using a SVD approach (\algline{5}).
Once all the planes have been computed, the rigid displacement between the two sensors is estimated
(\alglines{7-9}) using the plane-to-plane calibration method described in \cite{unnikrishnan2005fast}.

The plane equations extracted from the images are then represented w.r.t. the depth sensor
reference frames using the transformation matrices just computed.
These planes are used as reference locations for the curve fitting procedure
(\alglines{10-13}), as we did with the undistortion map $\U$ in \algref{alg:undistortion_map_creation}.

\begin{figure}
\small
\begin{algorithmic}[1]
\Require $\{\TF{\widehat{\map{C}}_1}{}{D}, \TF{\widehat{\map{C}}_2}{}{D}, \dots, \TF{\widehat{\map{C}}_M}{}{D}\}$ \Comment Undistorted point clouds
\Require $\{\mathbb{I}_1, \mathbb{I}_2, \dots, \mathbb{I}_M\}$ \Comment Wall point coordinates
\Require $\{\TF{\map{B}_1}{}{C}, \TF{\map{B}_2}{}{C}, \dots, \TF{\map{B}_M}{}{C}\}$ \Comment Checkerboard corners
\Ensure $\G$ \Comment Global error correction map
\Ensure $\TF{\matx{T}}{C}{D}$ \Comment Camera-depth sensor transformation matrix
\For{$k \gets 1, 2, \dots, M$}
  \State $\tf{\matx{T}}{\fr{B}_k}{\fr{C}} \gets \Call{solvePnP}{\matx{K}_\obj{C}, \vect{dist}_\obj{C}, \TF{\map{B}}{}{B}, \TF{\map{B}_k}{}{\I_C}}$
  \State $\TF{\map{B}_k}{}{C} \gets \tf{\matx{T}}{\fr{B}_k}{\fr{C}} \cdot \TF{\map{B}}{}{B}$
  \State $\TF{\pi_{\map{B}_k}}{}{C} \gets \Call{fitPlane}{\TF{\map{B}_k}{}{C}}$
  \State $\TF{\pi_{\widehat{\map{C}}_k}}{}{D} \gets \Call{fitPlane}{\TF{\widehat{\map{C}}_k}{}{D}, \mathbb{I}_k}$
\EndFor
\State $\set{\Pi}_{\map{B}} \gets (\TF{\pi_{\map{B}_1}}{}{C}, \TF{\pi_{\map{B}_2}}{}{C}, \dots, \TF{\pi_{\map{B}_M}}{}{C})$
\State $\set{\Pi}_{\widehat{\map{C}}} \gets (\TF{\pi_{\widehat{\map{C}}_1}}{}{D}, \TF{\pi_{\widehat{\map{C}}_2}}{}{D}, \dots, \TF{\pi_{\widehat{\map{C}}_M}}{}{D})$
\State $\TF{\matx{T}}{C}{D} \gets \Call{estimateTransform}{\set{\Pi}_{\map{B}}, \set{\Pi}_{\widehat{\map{C}}}}$
\For{$k \gets 1, 2, \dots, M$}
  \State $\TF{\pi_{\map{B}_k}}{}{D} \gets \TF{\matx{T}}{C}{D} \cdot \TF{\pi_{\map{B}_k}}{}{C}$
  \State $(\G, \E) \gets \Call{updateMap}{\G, \E, \TF{\widehat{\map{C}}_k}{}{D}, \mathbb{I}_k, \TF{\pi_{\map{B}_k}}{}{D}}$
\EndFor
\end{algorithmic}
\caption{Pseudocode of the algorithm for the initial estimation of the global correction map $\G$.}
\label{alg:global_map_estimation}
\end{figure}

\subsection{Non-linear Refinement}
\label{sec:global_map_optimization}

Once the global correction map $\G$ and the camera-depth sensor transformation matrix
$\TF{\matx{T}}{C}{D}$ have been estimated, we refine them within a non-linear optimization
framework.
To take into account the error on the checkerboard poses estimation, we follow the
bundle-adjustment approach as described in \cite{bundle-adjustment}: we also refine \emph{all} 
the checkerboard poses $\tf{\matx{T}}{\fr{B}_k}{\fr{C}}$, with $k = 1, \dots, M$.
Moreover, to take into account the error on the intrinsic parameters of the depth camera
$\matx{K}_\obj{D}$, the focal lengths and the principal point are refined too.
So, let define $\tf{\myspace{T}_\obj{B}}{}{C} \triangleq \left\{ \tf{\matx{T}}{\fr{B}_1}{\fr{C}},
\tf{\matx{T}}{\fr{B}_2}{\fr{C}}, \dots, \tf{\matx{T}}{\fr{B}_M}{\fr{C}} \right\}$ as the set of
checkerboard poses in the camera coordinates, estimated with the \emph{solvePnP} function.\\
Formally, the results of the non-linear refinement is:
\begin{equation*}
  \left( \G, \TF{\matx{T}}{C}{D}, \tf{\myspace{T}_\obj{B}}{}{C}, \matx{K}_\obj{D} \right) = 
  \underset{\G, \TF{\matx{T}}{C}{D}, \tf{\myspace{T}_\obj{B}}{}{C}, \matx{K}_\obj{D}}{\arg \min} \sum_{k = 1}^M 
  e_\f{repr}(k) + e_\f{pos}(k) \enspace,
\end{equation*}
Here $e_\f{repr}$ takes into account the reprojection error of the checkerboard corners onto the
images and depends on the checkerboard poses only. This error component is defined as:
\begin{equation*}
  e_\f{repr}(k) \triangleq \sum_{(r, c) \in \obj{B}} \frac{1}{\sigma^2_\obj{C}} \cdot 
  \left\Vert \f{proj}_\obj{C} \left( \tf{\matx{T}}{\fr{B}_k}{\fr{C}} \cdot
  \TF{\map{B}_k}{}{B}(r, c) \right) - \TF{\map{B}_k}{}{I_C}(r, c) \right\Vert^2
\end{equation*}
The summation is performed over all the checkerboard corners, $\f{proj}_\obj{C}$ is a general projection function that depends on both 
the camera matrix $\matx{K}_\obj{C}$ and the distortion coefficients $\vect{dist}_\obj{C}$. The residuals are weighted by the inverse of the variance of the corner estimation error $\sigma_\obj{C}^2$, where $\sigma_\obj{C} = 0.2$.\\

$e_\f{pos}$ represents the error between the planes defined by the checkerboards and the ones defined by the undistorted point cloud
\begin{multline*}
  e_\f{pos}(k) \triangleq \sum_{(u, v) \in \mathbb{I}_k} \frac{1}{|\mathbb{I}_k| \cdot
  \sigma_\U^2(z)} \cdot \\ \cdot
  \left\Vert \f{p}_{\TF{\pi_k}{}{D}} \left( \f{g}_{u, v} \left( \TF{\widehat{\map{C}}_k}{}{D}(u, v) \right) \right)
  - \f{g}_{u, v} \left( \TF{\widehat{\map{C}}_k}{}{D}(u, v) \right) \right\Vert^2 \enspace.
\end{multline*}
%
where $\f{p}_{\pi}(\pp)$ is the function that orthogonally projects a point $\pp$ onto plane $\pi$.
Formally speaking, this error is the distance between the cloud point
$\TF{\widehat{\map{C}}_k}{}{D}(u, v)$ corrected with the current estimation of $\G$ and its line-of-sight projection onto the plane $\TF{\pi_k}{}{D}$ defined by the checkerboard corner set $\TF{\map{B}_k}{}{D}$.
Such set is computed as
\begin{equation*}
  \TF{\map{B}_k}{}{D} = \TF{\matx{T}}{C}{D} \cdot \tf{\matx{T}}{\fr{B}_k}{\fr{C}} \cdot \TF{\map{B}}{}{B} \enspace.
\end{equation*}
Finally, each residual is weighted by the inverse of the variance on the depth measurements \emph{after} the
undistortion phase $\sigma_\U^2(z)$, multiplied by the number of wall points, i.e. $|\mathbb{I}_k|$.

\subsection{Implementation Details}

\subsubsection{Global Correction Map}
\label{sec:global_map_implementation}

As mentioned before, the global correction map $\G$ is someway similar to the undistortion map 
$\U$, but more simple.
In fact, $\G$ needs to transform planes into planes and, recalling that a plane transformation has
3 degrees of freedom, we just need 3 functions to satisfy this requirement.
Thus we define $\G$ as a discretized map constructed as $\U$ with $\chi_\G = W_\obj{D}$ and
$\psi_\G = H_\obj{D}$, that is, only 4 pixels contain a correction function $\f{g}_{u, v}(\cdot)$. For what concerns the other pixels, the global correction function is computed as a linear combination of the functions for the 4 boundary pixels.
Allowing 4 pixels to control the whole map usually leads to wrong results.
Let suppose, for example that 3 of such pixels contain an identity function and the fourth does 
not. Clearly the resulting surface will not be a plane anymore.
For this reason, only 3 of these pixels are actually computed, the fourth is instead estimated
exploiting the following invariant:
\begin{equation}
  \label{eq:f_g_invariant}
  \frac{\f{g}_{0, 0}(z) + \f{g}_{W_\obj{D}, H_\obj{D}}(z)}{2} = \frac{\f{g}_{W_\obj{D}, 0}(z) + \f{g}_{0, H_\obj{D}}(z)}{2} \enspace.
\end{equation}
Suppose now that $\f{g}_{W_\obj{D}, H_\obj{D}}(\cdot)$ is the dependent function:
$\f{g}_{W_\obj{D}, H_\obj{D}}(\cdot)$ can be estimated by fitting a function on an adequate set of pairs $\left( d, \f{g}_{W_\obj{D}, H_\obj{D}}(z) \right)$, where
\begin{equation*}
  \f{g}_{W_\obj{D}, H_\obj{D}}(z) = \f{g}_{W_\obj{D}, 0}(z) + \f{g}_{0, H_\obj{D}}(z) - \f{g}_{0, 0}(z)
\end{equation*}
is computed from \eqref{eq:f_g_invariant}.
An example of the presented global correction map $\G$ is reported in \figref{fig:global_map}.

\begin{figure}
\tikzstyle{undistortion} = [cyan!90!black]
\footnotesize
\centering
\begin{tikzpicture}[scale=0.30]

\draw[step=1,gray,very thin] (0, -6) grid (8, 0);
\draw[black,very thick] (0, 0) rectangle (8, -6);

\begin{scope}[shift={(-1.6, 0)}]
\draw[thin] (-0.3, 0) -- (0.3, 0);
\draw[thin] (-0.3, -6) -- (0.3, -6);
\draw[thin] (0, 0) -- node[fill=white]{$H_\obj{D}$} (0, -6);
\end{scope}

\begin{scope}[shift={(0, 1.6)}]
\draw[thin] (0, 0.3) -- (0, -0.3);
\draw[thin] (8, 0.3) -- (8, -0.3);
\draw[thin] (0, 0) -- node[fill=white]{$W_\obj{D}$} (8, 0);
\end{scope}

\begin{scope}[shift={(0.5, -0.5)}]
\foreach \i in {0,8}
{
  \foreach \j in {0,6}
  {
    \filldraw[undistortion,thick] (\i-0.25, -\j-0.25) rectangle (\i+0.25, -\j+0.25);
    \node[minimum size=0.2cm] (point_\i_\j) at (\i, -\j) {};
  }
}
\draw[undistortion,thin,dashed] (0, -6) rectangle (8, 0);

\filldraw[undistortion,thick,fill=white] (8-0.25, -6-0.25) rectangle (8+0.25, -6+0.25);
\filldraw[undistortion] (8-0.25, -6-0.25) -- (8+0.25, -6+0.25) -- (8+0.25, -6-0.25) -- cycle;

\filldraw[undistortion,thick,fill=white] (2.75, -3.75) rectangle (3.25, -4.25);
\node[minimum size=0.2cm] (pixel1) at (3, -4) {};
\draw[->,thick] (pixel1) -- (point_0_0);
\draw[->,thick] (pixel1) -- (point_8_6);
\draw[->,thick] (pixel1) -- (point_0_6);
\draw[->,thick] (pixel1) -- (point_8_0);

\end{scope}
\end{tikzpicture}
\vspace{0.5cm}\\
\begin{tikzpicture}[scale=0.35]

\filldraw[undistortion,thick] (-0.25, -0.25) rectangle (0.25, 0.25);
\node[anchor=west,text width=0.9\linewidth] at (0.5, -0.1) {Pixel $(u, v)^\T \in \SG$. For this pixel a correction function $\f{g}_{u, v}(\cdot)$ has been estimated.};

\filldraw[undistortion,thick,fill=white] (-0.25, -2.45) rectangle (0.25, -1.95);
\filldraw[undistortion] (-0.25, -2.45) -- (0.25, -1.95) -- (0.25, -2.45) -- cycle;
\node[anchor=west,text width=0.9\linewidth] at (0.5, -2.3) {Pixel $(u, v)^\T \in \SG$.
To guarantee planarity, for this pixel the correction function $\f{g}_{u, v}(\cdot)$ has been computed starting from the ones of the other 3 pixels in $\SG$.};

\filldraw[undistortion,thick,fill=white] (-0.25, -4.45) rectangle (0.25, -3.95);
\node[anchor=west,text width=0.9\linewidth] at (0.5, -4.5) {Pixel $(u, v)^\T \in \I_\obj{D}$. For this pixel the correction function $\f{g}_{u, v}(\cdot)$ is a linear combination of the functions of the pixels in $\map{S}_\G(u, v)$.};

\draw[->,thick] (-0.3, -6.4) -- (0.4, -6.4);
\node[anchor=west,text width=0.9\linewidth] at (0.5, -6.5) {Connection from a pixel $(u, v)^\T \in \I_\obj{D}$ to one in $\map{S}_\G(u, v)$.}; 

\end{tikzpicture}
\caption[Scheme of a global correction map $\G$.]
        {Visualization of a global correction map $\G$. A global correction function is estimated
        for all and only the pixels in $\SG$. For all the others pixels, instead, the function is computed
        as a linear combination of the ones estimated for the pixels in $\SG$.}
\label{fig:global_map}
\end{figure}

\subsubsection{Curve Fitting}

Since the correction map $\G$ is constructed in the same way as the undistortion map $\U$, the
considerations on the curve fitting procedure made in \sectref{sec:curve_fitting_U} are still
valid in the global error case.

%% file: experiments.tex
\section{Experimental Evaluation} \label{sec:depth_experiments}

\begin{figure}
\centering
\vspace{0.1cm}
\subfloat[]{\includegraphics[width=0.48\linewidth]{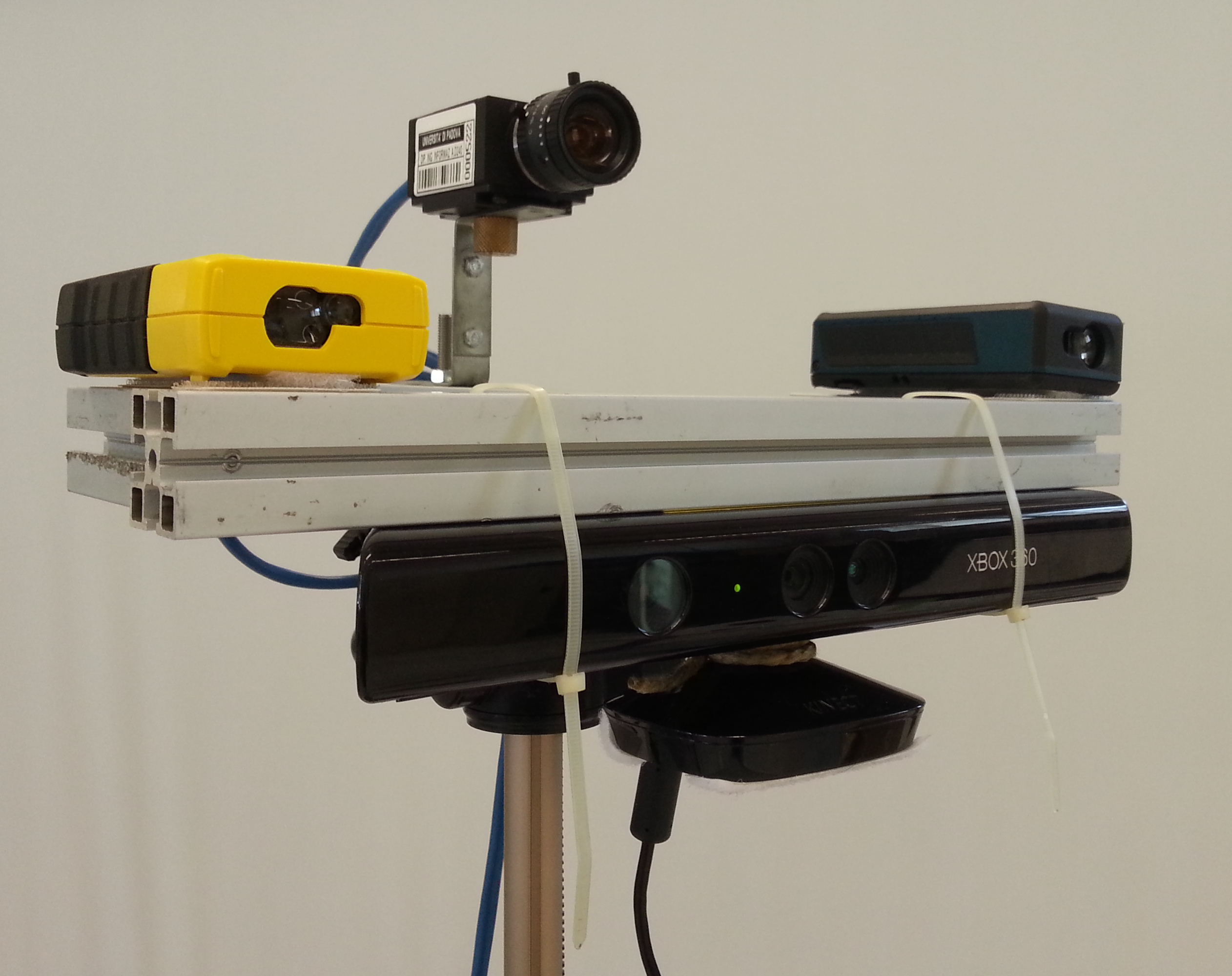}\label{fig:setup_a}}
\vspace{0.1cm}
\subfloat[]{\includegraphics[width=0.356\linewidth]{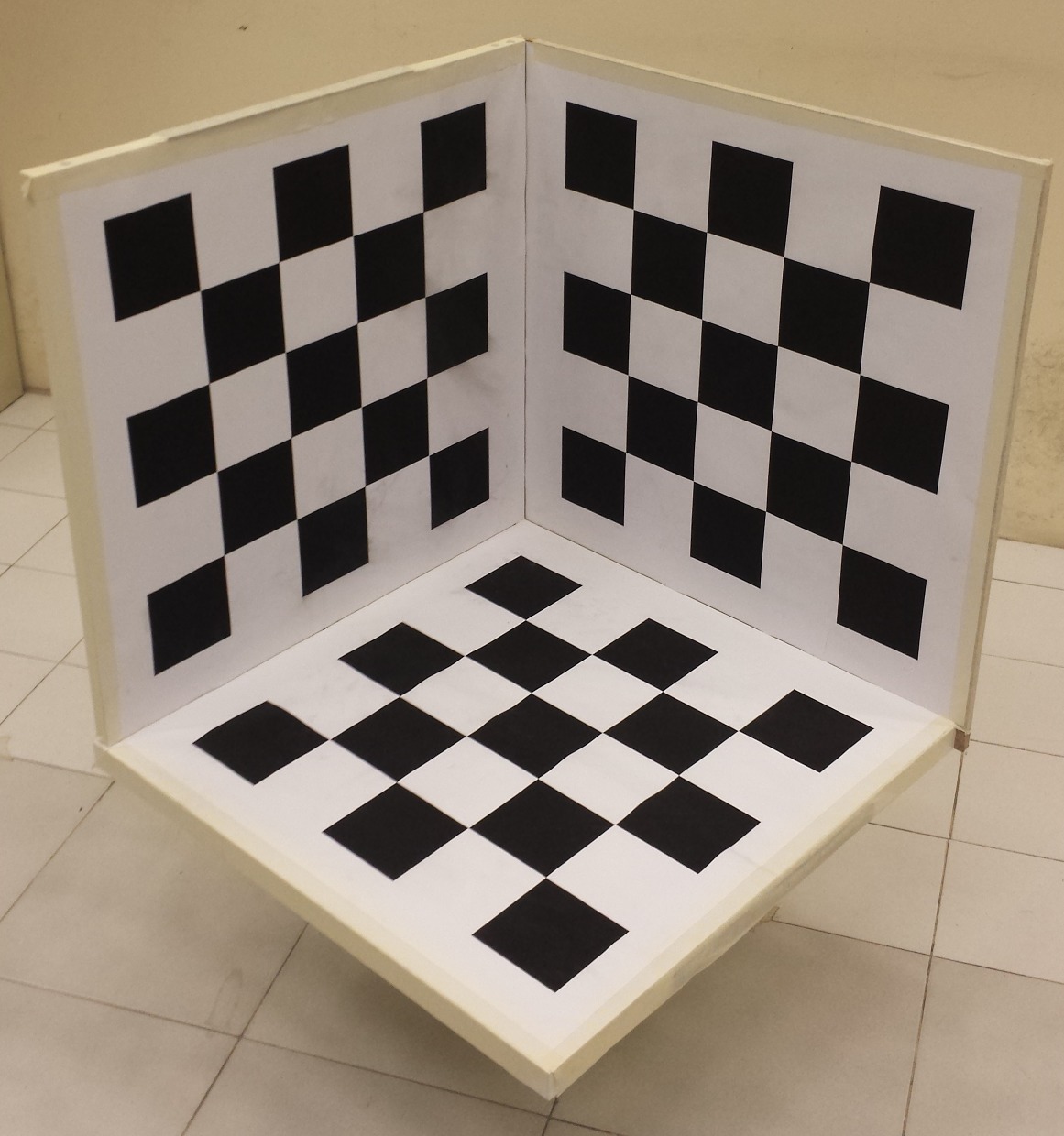}\label{fig:setup_b}}\\
\caption[Sensor setup]
        {(a) The support used to acquire the data for the experimental evaluations.
         The two laser meters are located on the left and right of the support to guarantee that the sensor is correctly aligned; (b) The reference hollow cube used as a ground truth in the performance evaluations.}
\end{figure}

The goal of the presented experimental evaluations is to show that our method is able to provide robust and state-of-the-art calibration results for different types of RGB-D sensors. We used four RGB-D sensors: two Microsoft Kinect 1 (called \textsc{kinect1a} and \textsc{kinect1b} in the plots), an Asus Xtion Pro Live (called \textsc{asus} in the plots), and a Microsoft Kinect 2. The RGB camera of each device has been previously calibrated exploiting a standard calibration tool. Actually, a good RGB camera calibration is an essential requirement of all the tested systems. Each sensor has been mounted, one at a time, on a support that includes two rigidly mounted high precision laser meters, and a high resolution RGB camera (see \figref{fig:setup_a}).

We attached a checkerboard on a wall, collecting for each device two datasets: a \emph{training set} (i.e., the dataset used to perform the calibration) and a \emph{test set} (i.e., the dataset used to evaluate the calibration accuracy).
The training set contains views of the checkerboard from the device camera, the depth sensor and the high resolution camera from different locations and orientations. 
The test set, instead, has been acquired by positioning the support of \figref{fig:setup_a} orthogonal to the wall at different distances and measuring such distances with the two laser distance meters. 


We independently analyze the results of our undistortion approach (\sectref{sec:undistortion_map_exp}), the results of our global correction approach (\sectref{sec:global_experiments_exp}), and the provided camera-depth sensor transformation accuracy (\sectref{sec:total_experiments_exp}). We also compare our method with other state-of-the-art calibration systems (\sectref{sec:comparison}), using the original implementations provided by the authors and a reference pattern as a ground truth (\figref{fig:setup_b}). In almost all tests, our system outperforms the other evaluated systems. We finally report some experiments of an RGB-D visual odometry system applied to a mobile robot, where we show that the accuracy in the ego-motion estimation highly benefits from using RGB-D data calibrated with our system (\sectref{sec:vo_usecase}).

%
%
%

\subsection{Undistortion Map}\label{sec:undistortion_map_exp}

To evaluate the performance of our undistortion approach, we introduce a metric called \emph{planarity error}.
For each cloud of the test set, we extract the wall point indices $\mathbb{I}$ from its undistorted 
version as described in \sectref{sec:undistortion_map_wall_points_selection}.
We define the planarity error as:
\begin{equation*}
e_{\f{plan}} = \sqrt{ \frac{1}{|\mathbb{I}|} \sum_{(u, v) \in \mathbb{I}} \left\Vert \vect{n}^\T \map{C}(u, v) - d \right\Vert^2 } \enspace,
\end{equation*}
where $\map{C}$ is a generic point cloud of the test set (we consider both the original and the
undistorted versions) and $\pi$ is the plane with equation $\vect{n}^\T \vect{x} - d = 0$ fitted to
the wall points with indices in $\mathbb{I}$.

\subsubsection{Undistortion Map Functions}

\begin{figure}
\centering
\footnotesize
\begin{tikzpicture}
\begin{axis}[
    height=0.50\linewidth,
    width=0.80\linewidth,
	xlabel={Average plane distance [m]},
	ylabel={Planarity error [cm]},
	ymin=0, ymax=5,
	xmin=0.95, xmax=4.65,
	cycle list name=color list,
	legend pos=outer north east,
	legend cell align=left,
	]
	\addplot+[no marks,dashed] table[x index=0,y index=1] {data/U_changing_poly_degree.txt};
	\addlegendentry{Original};
	\addplot+[no marks] table[x index=0,y index=2] {data/U_changing_poly_degree.txt};
	\addlegendentry{$1^\text{st}$ degree};
	\addplot+[no marks] table[x index=0,y index=3] {data/U_changing_poly_degree.txt};
	\addlegendentry{$2^\text{nd}$ degree};
	\addplot+[no marks,green!60!black] table[x index=0,y index=4] {data/U_changing_poly_degree.txt};
	\addlegendentry{$3^\text{rd}$ degree};
	\addplot+[no marks] table[x index=0,y index=5] {data/U_changing_poly_degree.txt};
	\addlegendentry{$4^\text{th}$ degree};
\end{axis}
\end{tikzpicture}
\caption{Planarity error when varying the degree of the undistortion polynomials.
         From the plot we can see that linear functions are not able to correctly model the distortion introduced by the sensor.
         On the other hand, quartic functions tend instead to overfit the training data (e.g., in the right part of the plot the error of the quartic functions increases w.r.t. the errors of the quadratic and cubic functions).}
\label{fig:U_varying_degree_results}
\end{figure}
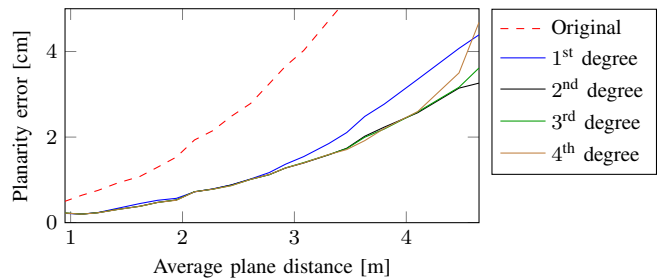

In the previous sections, we have always talked about ``undistortion functions'' without providing many details about the nature of these functions. Actually, analyzing the error on the plane estimation described in \sectref{sec:depth_error_model}, we evinced that for both the SL and ToF sensors, this error is well described by a \emph{quadratic polynomial}: this hypothesis is further confirmed also by our experiments. We calibrated each sensor using different types of undistortion functions (linear, quadratic, cubic, \dots): the higher degree functions has been tested to provide an additional proof of our hypothesis. For each cloud in the test sets, we computed the planarity error introduced above. The plot in \figref{fig:U_varying_degree_results} clearly shows that all the super-linear functions provide better undistortion results when compared to the linear functions. Moreover, higher degree polynomials get similar of even worst results compared with quadratic functions, since they tend to overfit the training data. Therefore, all the tests presented in following sections have been performed using quadratic undistortion functions.

\subsubsection{Map Discretization}

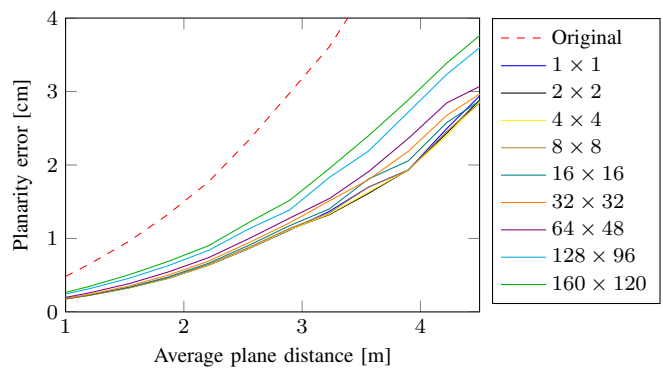
\begin{figure}
\centering
\footnotesize
\begin{tikzpicture}
\begin{axis}[
    height=0.62\linewidth,
    width=0.80\linewidth,
	xlabel={Average plane distance [m]},
	ylabel={Planarity error [cm]},
	ymin=0, ymax=4,
	xmin=1, xmax=4.5,
	cycle list name=color list,
	legend pos=outer north east,
	legend cell align=left,
	]
	\addplot+[no marks, dashed] table[x index=0,y index=1] {data/U_changing_bin_size_error.txt};
	\addlegendentry{Original};
	\addplot+[no marks] table[x index=0,y index=2] {data/U_changing_bin_size_error.txt};
	\addlegendentry{$1 \times 1$};
	\addplot+[no marks] table[x index=0,y index=3] {data/U_changing_bin_size_error.txt};
	\addlegendentry{$2 \times 2$};
	\addplot+[no marks] table[x index=0,y index=4] {data/U_changing_bin_size_error.txt};
	\addlegendentry{$4 \times 4$};
	\addplot+[no marks] table[x index=0,y index=5] {data/U_changing_bin_size_error.txt};
	\addlegendentry{$8 \times 8$};
	\addplot+[no marks] table[x index=0,y index=6] {data/U_changing_bin_size_error.txt};
	\addlegendentry{$16 \times 16$};
	\addplot+[no marks] table[x index=0,y index=7] {data/U_changing_bin_size_error.txt};
	\addlegendentry{$32 \times 32$};
	\addplot+[no marks] table[x index=0,y index=8] {data/U_changing_bin_size_error.txt};
	\addlegendentry{$64 \times 48$};
	\addplot+[no marks] table[x index=0,y index=9] {data/U_changing_bin_size_error.txt};
	\addlegendentry{$128 \times 96$};
	\addplot+[no marks] table[x index=0,y index=10] {data/U_changing_bin_size_error.txt};
	\addlegendentry{$160 \times 120$};
	
\end{axis}
\end{tikzpicture}
\caption{Planarity error of the wall points when changing the bin size.
         In the plot it is visible that increasing the bin size also the error increases, but not
         as much as expected.
         Actually, only with bin sizes staring from $16 \times 16$ the error increase starts being
         non-negligible.}
\label{fig:U_changing_bin_size_error}
\end{figure}

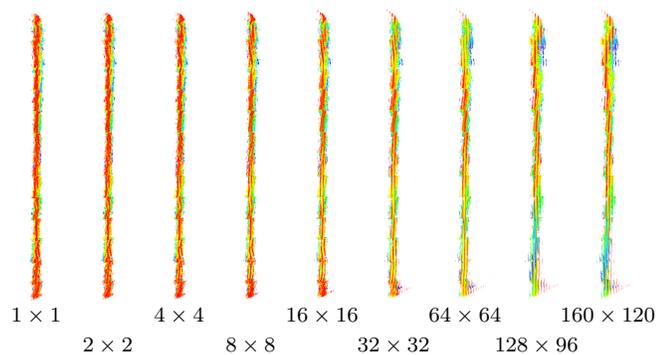
\begin{figure}
\centering
\footnotesize
\begin{tikzpicture}[]
	\node[text width=0.9\linewidth] at (-0.2cm,0) [anchor=south west] {\includegraphics[width=\linewidth]{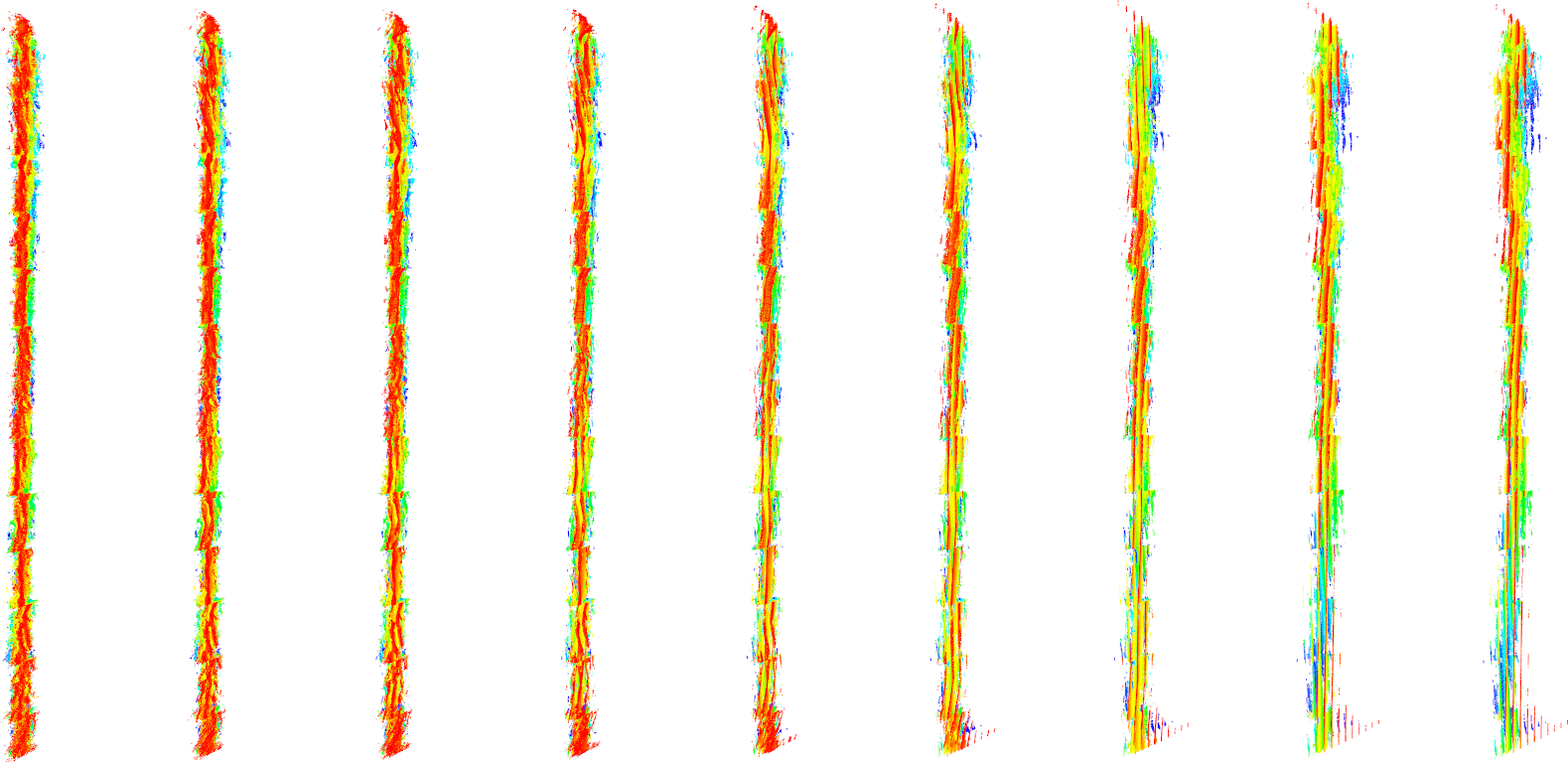}};
	\node[align=center] at (0cm,0.1cm) [anchor=north] {$1 \times 1$};
	\node[align=center] at (0.95cm,-0.3cm) [anchor=north] {$2 \times 2$};
	\node[align=center] at (1.9cm,0.1cm) [anchor=north] {$4 \times 4$};
	\node[align=center] at (2.85cm,-0.3cm) [anchor=north] {$8 \times 8$};
	\node[align=center] at (3.8cm,0.1cm) [anchor=north] {$16 \times 16$};
	\node[align=center] at (4.75cm,-0.3cm) [anchor=north] {$32 \times 32$};
	\node[align=center] at (5.7cm,0.1cm) [anchor=north] {$64 \times 64$};
	\node[align=center] at (6.65cm,-0.3cm) [anchor=north] {$128 \times 96$};
	\node[align=center] at (7.6cm,0.1cm) [anchor=north] {$160 \times 120$};
\end{tikzpicture}
\caption[Top-view of the cloud of a planar surface undistorted using maps with different bin
         sizes.]
        {Top-view of the cloud of a planar surface undistorted using maps with different bin sizes. Note that the resulting clouds are similar, especially the four on the left (points with different y-coordinates are drawn with different colors).}
\label{fig:U_changing_bin_size_results}
\end{figure}

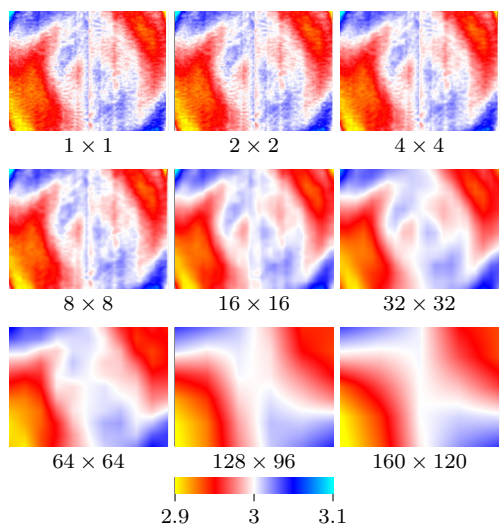
\begin{figure}
\centering
\footnotesize
\tikzstyle{map4} = [draw,black,inner sep=0,text width=2.1cm]

\begin{tikzpicture}[xscale=2.2,yscale=-2.1]
	\node[map4] at (0,0) [anchor=south] {\includegraphics[width=\linewidth]{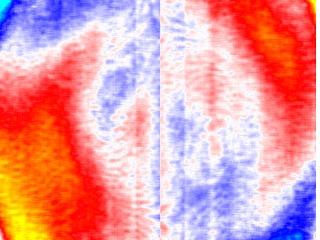}};
	\node[align=center] at (0,0) [anchor=north] {$1 \times 1$};
	\node[map4] at (1,0) [anchor=south] {\includegraphics[width=\linewidth]{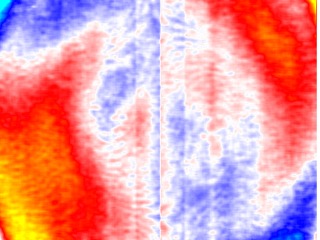}};
	\node[align=center] at (1,0) [anchor=north] {$2 \times 2$};
	\node[map4] at (2,0) [anchor=south] {\includegraphics[width=\linewidth]{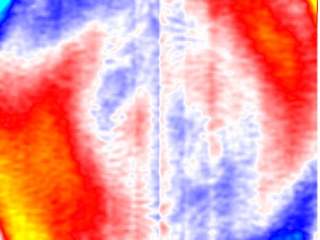}};
	\node[align=center] at (2,0) [anchor=north] {$4 \times 4$};

	\node[map4] at (0,1) [anchor=south] {\includegraphics[width=\linewidth]{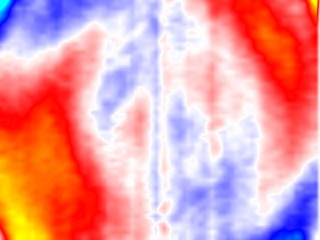}};
	\node[align=center] at (0,1) [anchor=north] {$8 \times 8$};
	\node[map4] at (1,1) [anchor=south] {\includegraphics[width=\linewidth]{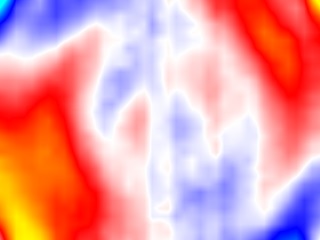}};
	\node[align=center] at (1,1) [anchor=north] {$16 \times 16$};
	\node[map4] at (2,1) [anchor=south] {\includegraphics[width=\linewidth]{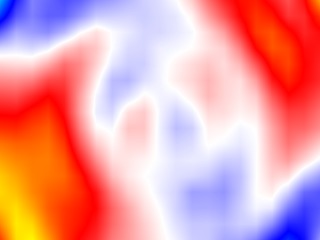}};
	\node[align=center] at (2,1) [anchor=north] {$32 \times 32$};

	\node[map4] at (0,2) [anchor=south] {\includegraphics[width=\linewidth]{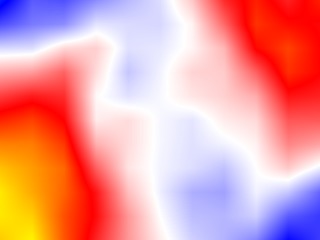}};
	\node[align=center] at (0,2) [anchor=north] {$64 \times 64$};
	\node[map4] at (1,2) [anchor=south] {\includegraphics[width=\linewidth]{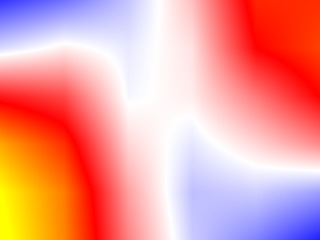}};
	\node[align=center] at (1,2) [anchor=north] {$128 \times 96$};
	\node[map4] at (2,2) [anchor=south] {\includegraphics[width=\linewidth]{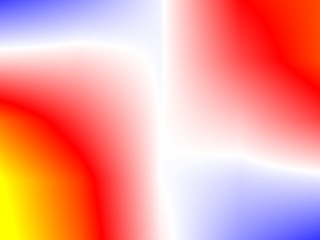}};
	\node[align=center] at (2,2) [anchor=north] {$160 \times 120$};
\end{tikzpicture}

\begin{tikzpicture}[]
	\node[map4] at (0, 0) [anchor=south] {\includegraphics[width=\linewidth]{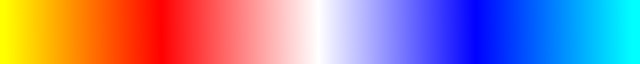}};
	\node[] at (0, -0.1) [anchor=north] {$3$};
	\node[] at (-1.05cm, -0.1) [anchor=north] {$2.9$};
	\node[] at (1.05cm,- 0.1) [anchor=north] {$3.1$};
	\draw[] (0, -0.1) -- (0, 0);
	\draw[] (-1.05cm, -0.1) -- (-1.05cm, 0);
	\draw[] (1.05cm, -0.1) -- (1.05cm, 0);
\end{tikzpicture}

\caption{Undistortion maps computed using different bin size, evaluated at a distance of 3 meters.}
\label{fig:U_changing_bin_size}
\end{figure}

To select the most appropriate bin size values (i.e., $\chi_\U$ and $\psi_\U$, described in \sectref{sec:undistortion_map_implementation}), we evaluated the planarity error varying the two parameters.
The results are reported in \figref{fig:U_changing_bin_size_error}.
Differently from what we expected, such parameters do not affect so much the results. Actually, up to a $8 \times 8$ pixels size, the planarity error is almost identical. Only with greater sizes, starting from $16 \times 16$, the error increases, especially close to the image corners (a qualitative comparison is reported in \figref{fig:U_changing_bin_size_results}).
Some examples of the generated undistortion maps, computed associating at each pixel $(u, v)^\T$ the value $\f{u}_{u, v}(z) - z$ for a given $z$, are reported in \figref{fig:U_changing_bin_size}: obviously, the maps become smoother as the bin size increases.

In our experience, we found that a bin size of $4 \times 4$ pixels represents a good trade-off between computational efficiency and robustness. Actually,
as mentioned before, larger bins tend to fail close to the image corners, while in other experiments we noticed that smaller bins tend to perform badly with small calibration datasets because of the lack of data for some pixels.

%

\subsubsection{Test Set Results}

We finally tested our algorithm against the test sets. Results of the planarity error evaluation are reported in \figref{fig:U_final_results}.
As expected, for the three SL sensors the proposed method permits to drastically improve the planarity of the depth data generated by the calibrated sensor.
Looking at the plots one could argue: why isn't the error after the undistortion closer to zero?
Actually, to the best of our knowledge, the error curve calculated after the undistortion is mainly due to the sensor noise and the quantization error.
Therefore it is not possible to further reduce this error.
The error reduction is instead less substantial in the case of kinect 2, especially for increasing distances. Actually, for increasing distances, large areas of the image near the borders just provide random depth values: most of the remaining error in \figref{fig:U_final_results} is due to this unpredictable white noise.

\begin{figure}
\centering
\footnotesize
\subfloat[\textsc{kinect1a} -- top]{\includegraphics[scale=0.18]{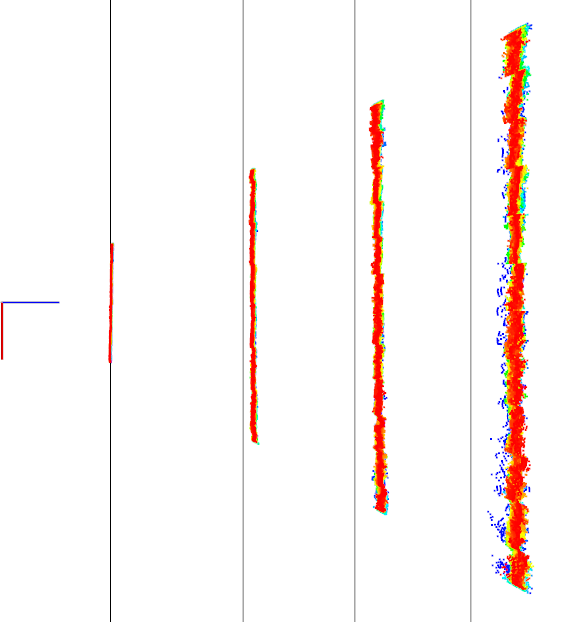}}
\subfloat[\textsc{kinect1b} -- top]{\includegraphics[scale=0.18]{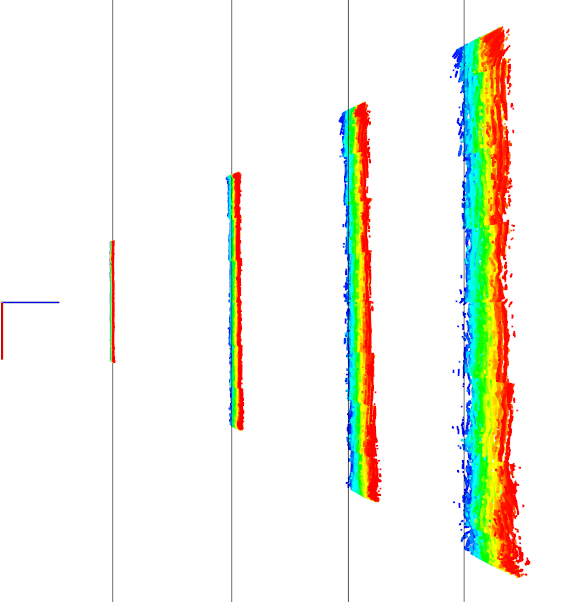}}\\
\subfloat[\textsc{asus} -- top]{\includegraphics[scale=0.18]{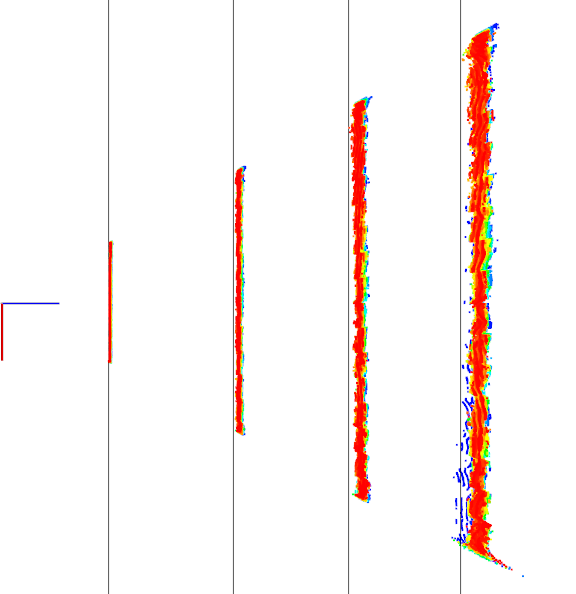}}
\subfloat[\textsc{kinect1b} -- side]{\includegraphics[scale=0.18]{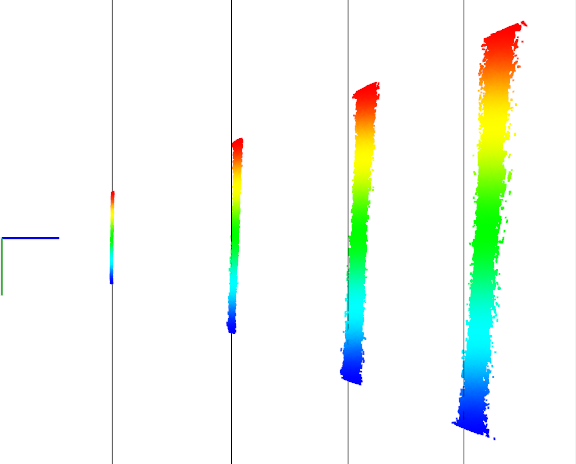}}
\caption{Top and side views of the point clouds of \figref{fig:depth_error1} after the undistortion
         phase. Again, the gray lines show the depth measured by means of the laser meters, while points with different y-coordinates are drawn with different colors.
         As we can see, the clouds are now more planar than the original ones, however, they are not in the right position, not even correctly oriented.}
\label{fig:undistorted_depth_error}
\end{figure}

In \figref{fig:U_final_maps} the undistortion maps estimated using a bin size of $4 \times 4$ pixels are shown.
Looking at the scales, we can see that the magnitude of the correction is consistent with the planarity error of the original data.
In the Kinect 2 case, for increasing distances the polynomials close to the borders tends to diverge: this is due to the fact that they are trying to correct just white noise.
In \figref{fig:undistorted_depth_error} the results of our undistortion algorithm applied to the clouds of \figref{fig:depth_error1} are reported.
As expected, the clouds are now planar but they are not in the right position, not even correctly oriented. These errors will be corrected in the next stage of our algorithm.
\begin{figure}
\centering
\footnotesize
\pgfplotslegendfromname{leg:U_final_results_legend}\\\vspace{0.25cm}

\subfloat[Planarity error for the three SL sensors.]
{
\begin{tikzpicture}
\begin{axis}[
    height=0.40\linewidth,
    width=0.42\linewidth,
	xlabel={Average depth [m]},
	ylabel={Planarity error [cm]},
	ymin=0, ymax=10,
	xmin=1, xmax=4.6,
	cycle list name=color list,
	legend cell align=left,
	legend columns=-1,
	legend to name=leg:U_final_results_legend,
	legend style={/tikz/every even column/.append style={column sep=0.25cm}},
	title={\textsc{kinect1a}},
	]
	\addplot+[no marks] table[x index=0,y index=1] {data/U_kinect_47A.txt};
	\addplot+[no marks] table[x index=0,y index=2] {data/U_kinect_47A.txt};
    \addplot+[no marks, dashed] {100*(-0.00029 + 0.00037*x + 0.001365*x^2)};
	\addlegendentry{Original};
	\addlegendentry{Undistorted};
	\addlegendentry{Quantization error};
\end{axis}
\end{tikzpicture}
\hfill
\begin{tikzpicture}
\begin{axis}[
    height=0.40\linewidth,
    width=0.42\linewidth,
	xlabel={Average depth [m]},
	ymin=0, ymax=10,
	xmin=1, xmax=4.5,
	cycle list name=color list,
	legend pos=north west,
	legend cell align=left,
	title={\textsc{kinect1b}},
	]
	\addplot+[no marks] table[x index=0,y index=1] {data/U_kinect_51A.txt};
	\addplot+[no marks] table[x index=0,y index=2] {data/U_kinect_51A.txt};
    \addplot+[no marks, dashed] {100*(-0.00029 + 0.00037*x + 0.001365*x^2)};
\end{axis}
\end{tikzpicture}
\hfill
\begin{tikzpicture}
\begin{axis}[
    height=0.40\linewidth,
    width=0.42\linewidth,
	xlabel={Average depth [m]},
	ymin=0, ymax=10,
	xmin=1, xmax=4.4,
	cycle list name=color list,
	legend pos=north west,
	legend cell align=left,
	title={\textsc{asus}},
	]
	\addplot+[no marks] table[x index=0,y index=1] {data/U_asus.txt};
	\addplot+[no marks] table[x index=0,y index=2] {data/U_asus.txt};
    \addplot+[no marks, dashed] {100*(-0.00029 + 0.00037*x + 0.001365*x^2)};
\end{axis}
\end{tikzpicture}
}\\%
\subfloat[Planarity error for the Kinect 2 sensor.]
{
\hspace{1cm}
\begin{tikzpicture}
\begin{axis}[
    height=0.40\linewidth,
    width=0.48\linewidth,
	xlabel={Average depth [m]},
	ylabel={Planarity error [cm]},
	ymin=0, ymax=2,
	xmin=1, xmax=5,
	cycle list name=color list,
	legend pos=north west,
	legend cell align=left,
	title={\textsc{kinect2}},
	]
	\addplot+[no marks] table[x index=0,y index=1] {data/U_kinect2.txt};
	\addplot+[no marks] table[x index=0,y index=2] {data/U_kinect2.txt};
\end{axis}
\end{tikzpicture}
\hspace{1cm}
}
\caption{Planarity error for the four sensors. For the SL sensors, the proposed approach is able to drastically reduce the distance of the measured points from the plane that best fits the data. Here, the error after the undistortion phase is mainly due to quantization \cite{smisek20113d}. For the Kinect 2 sensor, instead, the difference is bounded by the random noise that appears with increasing distances.}
\label{fig:U_final_results}
\end{figure}
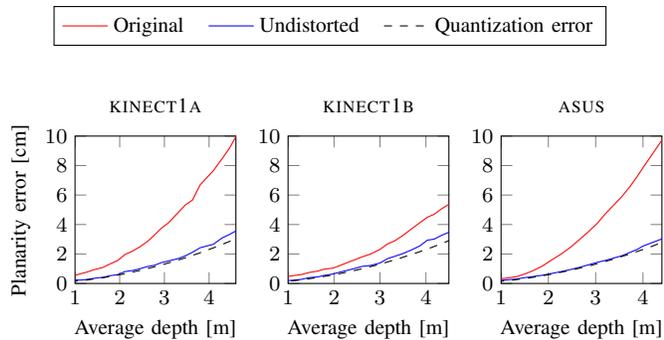
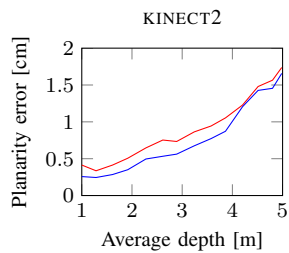
\begin{figure}
\centering
\footnotesize
\tikzstyle{map4} = [draw,black,inner sep=0,text width=2.1cm]

\subfloat
{
\begin{tikzpicture}
	\node[map4] (image) at (0, 0.4) [anchor=south] {\includegraphics[width=\linewidth]{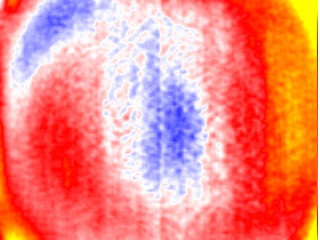}};
	\node[map4] at (0, 0) [anchor=south] {\includegraphics[width=\linewidth]{scale.jpg}};
	\node[above=0.2cm of image,align=center] {\textsc{kinect1a}}; 
	\node[inner sep=0] at (0, -0.15) [anchor=north] {$1$};
	\node[inner sep=0] at (-1.05cm, -0.15) [anchor=north west] {$0.98$};
	\node[inner sep=0] at (1.05cm, -0.15) [anchor=north east] {$1.02$};
	\draw[] (0, -0.1) -- (0, 0);
	\draw[] (-1.05cm, -0.1) -- (-1.05cm, 0);
	\draw[] (1.05cm, -0.1) -- (1.05cm, 0);
\end{tikzpicture}
\begin{tikzpicture}[]
	\node[map4] (image) at (0, 0.4) [anchor=south] {\includegraphics[width=\linewidth]{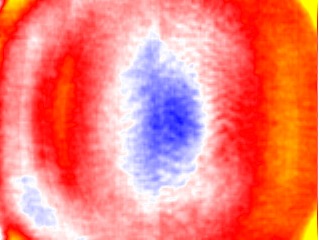}};
	\node[map4] at (0, 0) [anchor=south] {\includegraphics[width=\linewidth]{scale.jpg}};
	\node[above=0.2cm of image,align=center]  {\textsc{kinect1b}}; 
	\node[inner sep=0] at (0, -0.15) [anchor=north] {$1$};
	\node[inner sep=0] at (-1.05cm, -0.15) [anchor=north west] {$0.98$};
	\node[inner sep=0] at (1.05cm,- 0.15) [anchor=north east] {$1.02$};
	\draw[] (0, -0.1) -- (0, 0);
	\draw[] (-1.05cm, -0.1) -- (-1.05cm, 0);
	\draw[] (1.05cm, -0.1) -- (1.05cm, 0);
\end{tikzpicture}
\begin{tikzpicture}[]
	\node[map4] (image) at (0, 0.4) [anchor=south] {\includegraphics[width=\linewidth]{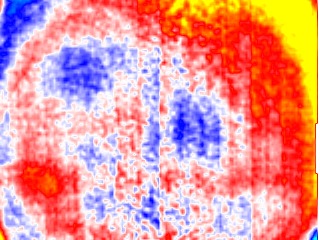}};
	\node[map4] at (0, 0) [anchor=south] {\includegraphics[width=\linewidth]{scale.jpg}};
	\node[above=0.2cm of image,align=center] {\textsc{asus}}; 
	\node[inner sep=0] at (0, -0.15) [anchor=north] {$1$};
	\node[inner sep=0] at (-1.05cm, -0.15) [anchor=north west] {$0.99$};
	\node[inner sep=0] at (1.05cm,- 0.15) [anchor=north east] {$1.01$};
	\draw[] (0, -0.1) -- (0, 0);
	\draw[] (-1.05cm, -0.1) -- (-1.05cm, 0);
	\draw[] (1.05cm, -0.1) -- (1.05cm, 0);
\end{tikzpicture}
\begin{tikzpicture}[]
	\node[map4] at (0, 0.4) [anchor=south] {\includegraphics[width=\linewidth, height=1.575cm]{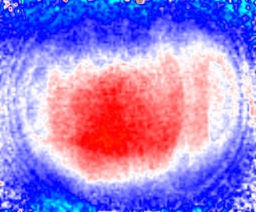}};
	\node[map4] at (0, 0) [anchor=south] {\includegraphics[width=\linewidth]{scale.jpg}};
	\node[above=0.2cm of image,align=center] {\textsc{kinect2}}; 
	\node[inner sep=0] at (0, -0.15) [anchor=north] {$1$};
	\node[inner sep=0] at (-1.05cm, -0.15) [anchor=north west] {$0.99$};
	\node[inner sep=0] at (1.05cm,- 0.15) [anchor=north east] {$1.01$};
	\draw[] (0, -0.1) -- (0, 0);
	\draw[] (-1.05cm, -0.1) -- (-1.05cm, 0);
	\draw[] (1.05cm, -0.1) -- (1.05cm, 0);
\end{tikzpicture}
}\\
\subfloat
{
\begin{tikzpicture}[]
	\node[map4] at (0, 0.4) [anchor=south] {\includegraphics[width=\linewidth]{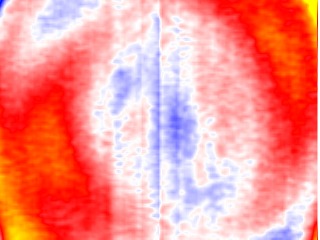}};
	\node[map4] at (0, 0) [anchor=south] {\includegraphics[width=\linewidth]{scale.jpg}};
	\node[inner sep=0] at (0, -0.15) [anchor=north] {$2$};
	\node[inner sep=0] at (-1.05cm, -0.15) [anchor=north west] {$1.95$};
	\node[inner sep=0] at (1.05cm,- 0.15) [anchor=north east] {$2.05$};
	\draw[] (0, -0.1) -- (0, 0);
	\draw[] (-1.05cm, -0.1) -- (-1.05cm, 0);
	\draw[] (1.05cm, -0.1) -- (1.05cm, 0);
\end{tikzpicture}
\begin{tikzpicture}[]
	\node[map4] at (0, 0.4) [anchor=south] {\includegraphics[width=\linewidth]{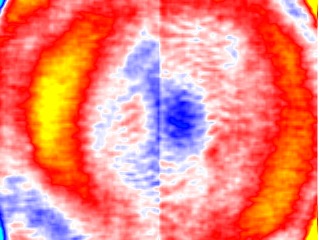}};
	\node[map4] at (0, 0) [anchor=south] {\includegraphics[width=\linewidth]{scale.jpg}};
	\node[inner sep=0] at (0, -0.15) [anchor=north] {$2$};
	\node[inner sep=0] at (-1.05cm, -0.15) [anchor=north west] {$1.97$};
	\node[inner sep=0] at (1.05cm,- 0.15) [anchor=north east] {$2.03$};
	\draw[] (0, -0.1) -- (0, 0);
	\draw[] (-1.05cm, -0.1) -- (-1.05cm, 0);
	\draw[] (1.05cm, -0.1) -- (1.05cm, 0);
\end{tikzpicture}
\begin{tikzpicture}[]
	\node[map4] at (0, 0.4) [anchor=south] {\includegraphics[width=\linewidth]{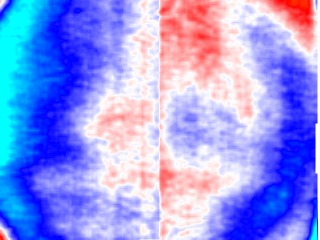}};
	\node[map4] at (0, 0) [anchor=south] {\includegraphics[width=\linewidth]{scale.jpg}};
	\node[inner sep=0] at (0, -0.15) [anchor=north] {$2$};
	\node[inner sep=0] at (-1.05cm, -0.15) [anchor=north west] {$1.95$};
	\node[inner sep=0] at (1.05cm,- 0.15) [anchor=north east] {$2.05$};
	\draw[] (0, -0.1) -- (0, 0);
	\draw[] (-1.05cm, -0.1) -- (-1.05cm, 0);
	\draw[] (1.05cm, -0.1) -- (1.05cm, 0);
\end{tikzpicture}
\begin{tikzpicture}[]
	\node[map4] at (0, 0.4) [anchor=south] {\includegraphics[width=\linewidth, height=1.575cm]{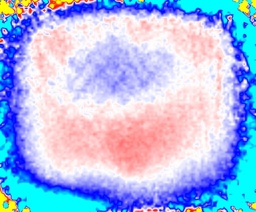}};
	\node[map4] at (0, 0) [anchor=south] {\includegraphics[width=\linewidth]{scale.jpg}};
	\node[inner sep=0] at (0, -0.15) [anchor=north] {$2$};
	\node[inner sep=0] at (-1.05cm, -0.15) [anchor=north west] {$1.985$};
	\node[inner sep=0] at (1.05cm,- 0.15) [anchor=north east] {$2.015$};
	\draw[] (0, -0.1) -- (0, 0);
	\draw[] (-1.05cm, -0.1) -- (-1.05cm, 0);
	\draw[] (1.05cm, -0.1) -- (1.05cm, 0);
\end{tikzpicture}
}\\
\subfloat
{
\begin{tikzpicture}[]
	\node[map4] at (0, 0.4) [anchor=south] {\includegraphics[width=\linewidth]{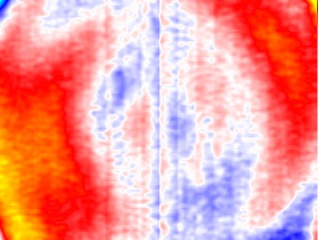}};
	\node[map4] at (0, 0) [anchor=south] {\includegraphics[width=\linewidth]{scale.jpg}};
	\node[inner sep=0] at (0, -0.15) [anchor=north] {$3$};
	\node[inner sep=0] at (-1.05cm, -0.15) [anchor=north west] {$2.89$};
	\node[inner sep=0] at (1.05cm,- 0.15) [anchor=north east] {$3.11$};
	\draw[] (0, -0.1) -- (0, 0);
	\draw[] (-1.05cm, -0.1) -- (-1.05cm, 0);
	\draw[] (1.05cm, -0.1) -- (1.05cm, 0);
\end{tikzpicture}
\begin{tikzpicture}[]
	\node[map4] at (0, 0.4) [anchor=south] {\includegraphics[width=\linewidth]{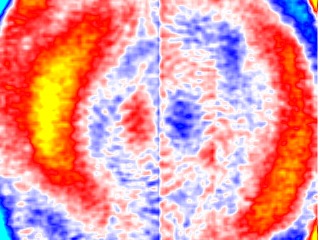}};
	\node[map4] at (0, 0) [anchor=south] {\includegraphics[width=\linewidth]{scale.jpg}};
	\node[inner sep=0] at (0, -0.15) [anchor=north] {$3$};
	\node[inner sep=0] at (-1.05cm, -0.15) [anchor=north west] {$2.95$};
	\node[inner sep=0] at (1.05cm,- 0.15) [anchor=north east] {$3.05$};
	\draw[] (0, -0.1) -- (0, 0);
	\draw[] (-1.05cm, -0.1) -- (-1.05cm, 0);
	\draw[] (1.05cm, -0.1) -- (1.05cm, 0);
\end{tikzpicture}
\begin{tikzpicture}[]
	\node[map4] at (0, 0.4) [anchor=south] {\includegraphics[width=\linewidth]{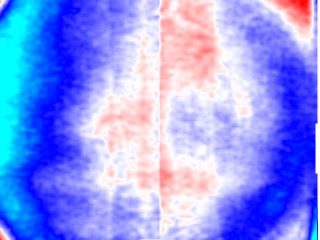}};
	\node[map4] at (0, 0) [anchor=south] {\includegraphics[width=\linewidth]{scale.jpg}};
	\node[inner sep=0] at (0, -0.15) [anchor=north] {$3$};
	\node[inner sep=0] at (-1.05cm, -0.15) [anchor=north west] {$2.86$};
	\node[inner sep=0] at (1.05cm,- 0.15) [anchor=north east] {$3.14$};
	\draw[] (0, -0.1) -- (0, 0);
	\draw[] (-1.05cm, -0.1) -- (-1.05cm, 0);
	\draw[] (1.05cm, -0.1) -- (1.05cm, 0);
\end{tikzpicture}
\begin{tikzpicture}[]
	\node[map4] at (0, 0.4) [anchor=south] {\includegraphics[width=\linewidth, height=1.575cm]{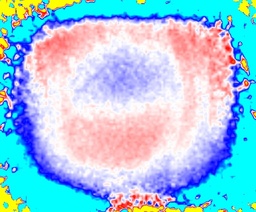}};
	\node[map4] at (0, 0) [anchor=south] {\includegraphics[width=\linewidth]{scale.jpg}};
	\node[inner sep=0] at (0, -0.15) [anchor=north] {$3$};
	\node[inner sep=0] at (-1.05cm, -0.15) [anchor=north west] {$2.98$};
	\node[inner sep=0] at (1.05cm,- 0.15) [anchor=north east] {$3.02$};
	\draw[] (0, -0.1) -- (0, 0);
	\draw[] (-1.05cm, -0.1) -- (-1.05cm, 0);
	\draw[] (1.05cm, -0.1) -- (1.05cm, 0);
\end{tikzpicture}
}\\
\subfloat
{
\begin{tikzpicture}[]
	\node[map4] at (0, 0.4) [anchor=south] {\includegraphics[width=\linewidth]{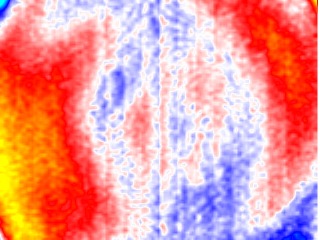}};
	\node[map4] at (0, 0) [anchor=south] {\includegraphics[width=\linewidth]{scale.jpg}};
	\node[inner sep=0] at (0, -0.15) [anchor=north] {$4$};
	\node[inner sep=0] at (-1.05cm, -0.15) [anchor=north west] {$3.82$};
	\node[inner sep=0] at (1.05cm,- 0.15) [anchor=north east] {$4.18$};
	\draw[] (0, -0.1) -- (0, 0);
	\draw[] (-1.05cm, -0.1) -- (-1.05cm, 0);
	\draw[] (1.05cm, -0.1) -- (1.05cm, 0);
\end{tikzpicture}
\begin{tikzpicture}[]
	\node[map4] at (0, 0.4) [anchor=south] {\includegraphics[width=\linewidth]{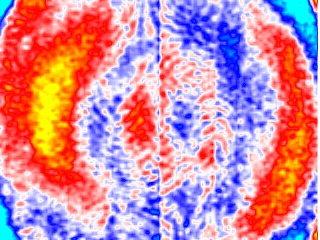}};
	\node[map4] at (0, 0) [anchor=south] {\includegraphics[width=\linewidth]{scale.jpg}};
	\node[inner sep=0] at (0, -0.15) [anchor=north] {$4$};
	\node[inner sep=0] at (-1.05cm, -0.15) [anchor=north west] {$3.92$};
	\node[inner sep=0] at (1.05cm,- 0.15) [anchor=north east] {$4.08$};
	\draw[] (0, -0.1) -- (0, 0);
	\draw[] (-1.05cm, -0.1) -- (-1.05cm, 0);
	\draw[] (1.05cm, -0.1) -- (1.05cm, 0);
\end{tikzpicture}
\begin{tikzpicture}[]
	\node[map4] at (0, 0.4) [anchor=south] {\includegraphics[width=\linewidth]{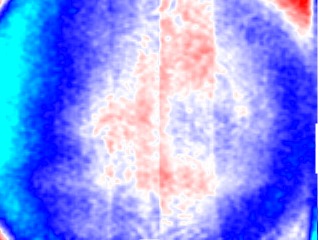}};
	\node[map4] at (0, 0) [anchor=south] {\includegraphics[width=\linewidth]{scale.jpg}};
	\node[inner sep=0] at (0, -0.15) [anchor=north] {$4$};
	\node[inner sep=0] at (-1.05cm, -0.15) [anchor=north west] {$3.72$};
	\node[inner sep=0] at (1.05cm,- 0.15) [anchor=north east] {$4.28$};
	\draw[] (0, -0.1) -- (0, 0);
	\draw[] (-1.05cm, -0.1) -- (-1.05cm, 0);
	\draw[] (1.05cm, -0.1) -- (1.05cm, 0);
\end{tikzpicture}
\begin{tikzpicture}[]
	\node[map4] at (0, 0.4) [anchor=south] {\includegraphics[width=\linewidth, height=1.575cm]{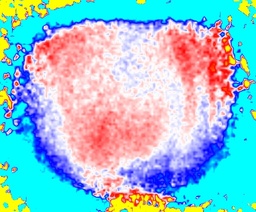}};
	\node[map4] at (0, 0) [anchor=south] {\includegraphics[width=\linewidth]{scale.jpg}};
	\node[inner sep=0] at (0, -0.15) [anchor=north] {$4$};
	\node[inner sep=0] at (-1.05cm, -0.15) [anchor=north west] {$3.975$};
	\node[inner sep=0] at (1.05cm,- 0.15) [anchor=north east] {$4.025$};
	\draw[] (0, -0.1) -- (0, 0);
	\draw[] (-1.05cm, -0.1) -- (-1.05cm, 0);
	\draw[] (1.05cm, -0.1) -- (1.05cm, 0);
\end{tikzpicture}
}
\caption{Maps generated by the algorithm for the four sensors with a bin size of $4 \times 4$ pixels. For each sensor the evaluation of the respective map at 1, 2, 3, and 4 meters is reported. Note that each map has its own scale and all values are in meters.}
\label{fig:U_final_maps}
\end{figure}

\subsection{Global Correction Map} \label{sec:global_experiments_exp}

In this section we report some results of the evaluation of the estimated global correction map
$\G$.
Recalling that $\G$ is computed using the checkerboard as the reference plane and that $\G$ and the RGB camera-depth sensor transformation are refined together, also the estimated transformation is taken into account to evaluate the results.
We estimate the error of the plane that results after the global correction with respect
to the plane defined by the checkerboard.

Firstly, the pose of the checkerboard with respect to the camera is estimated using the corners extracted from the image.
Then, the checkerboard plane is transformed into depth sensor coordinates. Finally, the average distance of the wall points (extracted from the cloud) to the checkerboard-defined plane is computed.


\subsubsection{Global Correction Map Functions}

Before evaluating the global correction map $\G$, as we did for the undistortion map $\U$, we
analyze the sample sets generated to compute the map to evince the most appropriate function type to fit to the data.
A first analysis of the error has been reported in \sectref{sec:depth_error_model}, in
\figref{fig:average_distance_error}.
Such error was computed by evaluating the difference between the average depth of a distorted
point cloud and the measurements provided by a laser distance meter.
Our sample sets, instead, contain pairs $(\widehat{z}, z_{\pi_\map{B}})$, where $\widehat{z}$ is 
the depth value after the undistortion phase, and $z_{\pi_\map{B}}$ is the expected depth, i.e. the
depth of the pixel as if it were laying on checkerboard-defined plane $\pi_\map{B}$.
Even in this case we tried to fit polynomial functions with different degrees: in \figref{fig:g_sample_sets_analysis} we report a comparison between quadratic and linear functions for two of the four sample sets used to generate the global correction map (we used in this case a SL sensor).
From the figure, it is clear that linear functions are not suitable since they do not fit properly to the data. A further confirmation of this fact is visible in \figref{fig:G_changing_poly_degree}, where the results of the calibration process varying the maximum degree of the polynomials from 1 to 4 is
reported. All the calibrations of the SL sensors reported in the following were performed treating each $\f{g}_{u, v}(\cdot)$
as a quadratic function with the constant factor set to zero.
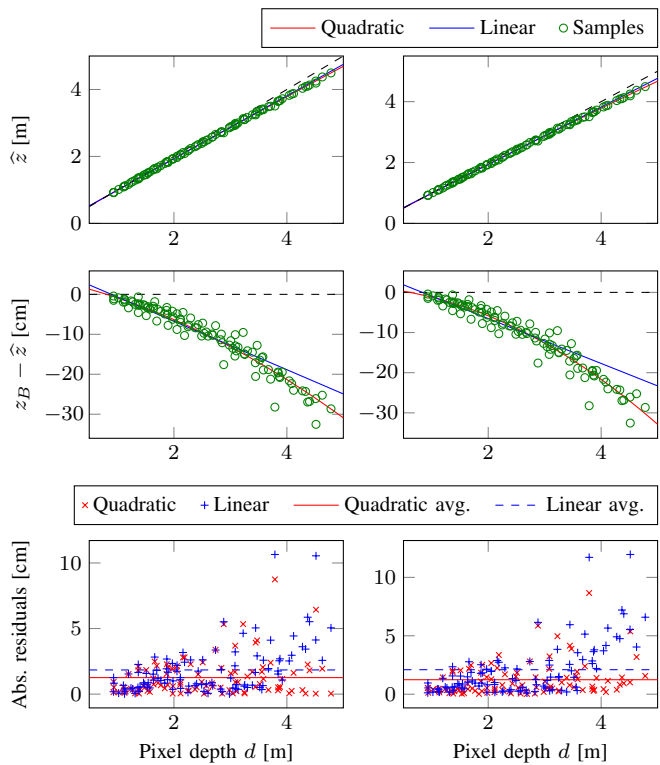
\begin{figure}
\begin{flushright}
\footnotesize
\pgfplotsset{yticklabel style={text width=0.5cm,align=right}}
\pgfplotslegendfromname{leg:G_sample_sets_analysis}\\\vspace{0.1cm}
\begin{tikzpicture}
\begin{axis}[
    height=0.43\linewidth,
    width=0.56\linewidth,
	ylabel={$\widehat{z}$ [m]},
	xmin=0.5, xmax=5,
	ymin=0, ymax=5,
	domain=0.5:5,
	cycle list name=color list,
	legend pos=north west,
	legend cell align=left,
	legend columns=-1,
	legend to name=leg:G_sample_sets_analysis,
	legend style={/tikz/every even column/.append style={column sep=0.25cm}},
	]
	\pgfplotsset{samples=50};
	\addplot+[no marks] {(0.0315562 + 0.966411*x - 0.00690876*x^2)};
	\addplot+[no marks] {(0.0540855 + 0.939321*x)};
	\addplot+[only marks,mark=o,mark size=1.5pt,black!50!green] table[x index=0,y index=1] {data/g_poly_0-0.txt};
	\addplot[dashed, black] {x};
	
	\addlegendentry{Quadratic};
	\addlegendentry{Linear};
	\addlegendentry{Samples};
\end{axis}
\end{tikzpicture}
\begin{tikzpicture}
\begin{axis}[
    height=0.43\linewidth,
    width=0.56\linewidth,
	xmin=0.5, xmax=5,
	ymin=0, ymax=5.5,
	domain=0.5:5,
	cycle list name=color list,
	legend pos=north west,
	legend cell align=left,
	]
	\pgfplotsset{samples=50};
	\addplot+[no marks] {(0.0122388 + 0.986659*x - 0.0109653*x^2)};
	\addplot+[no marks] {(0.0466517 + 0.944093*x)};
	\addplot+[only marks,mark=o,mark size=1.5pt,black!50!green] table[x index=0,y index=1] {data/g_poly_1-1.txt};
	\addplot[dashed, black] {x};
\end{axis}
\end{tikzpicture}
\\
\vspace{0.25cm}
\begin{tikzpicture}
\begin{axis}[
    height=0.43\linewidth,
    width=0.56\linewidth,
	ylabel={$z_\map{B} - \widehat{z}$ [cm]},
	xmin=0.5, xmax=5,
	domain=0.5:5,
	cycle list name=color list,
	legend pos=north west,
	legend cell align=left,
	]
	\pgfplotsset{samples=50};
	\draw[dashed] (axis cs:0.5,0) -- (axis cs:5.5,0);
	\addplot+[no marks] {100*((0.0315562 + 0.966411*x - 0.00690876*x^2) - x)};
	\addplot+[no marks] {100*((0.0540855 + 0.939321*x) - x)};
	\addplot+[only marks,mark=o,mark size=1.5pt,black!50!green] table[x index=0,y index=2] {data/g_poly_0-0.txt};
\end{axis}
\end{tikzpicture}
\begin{tikzpicture}
\begin{axis}[
    height=0.43\linewidth,
    width=0.56\linewidth,
	xmin=0.5, xmax=5,
	domain=0.5:5,
	cycle list name=color list,
	legend pos=south west,
	legend cell align=left,
	]
	\pgfplotsset{samples=50};
	\draw[dashed] (axis cs:0.5,0) -- (axis cs:5.5,0);
	\addplot+[no marks] {100*((0.0122388 + 0.986659*x - 0.0109653*x^2) - x)};
	\addplot+[no marks] {100*((0.0466517 + 0.944093*x) - x)};
	\addplot+[only marks,mark=o,mark size=1.5pt,black!50!green] table[x index=0,y index=2] {data/g_poly_1-1.txt};
\end{axis}
\end{tikzpicture}
\\\vspace{0.25cm}
\pgfplotslegendfromname{leg:G_sample_sets_analysis_residuals}\\\vspace{0.1cm}
\begin{tikzpicture}
\begin{axis}[
    height=0.43\linewidth,
    width=0.56\linewidth,
	xlabel={Pixel depth $d$ [m]},
	ylabel={Abs. residuals [cm]},
	xmin=0.5, xmax=5,
	domain=0.5:5,
	cycle list name=color list,
	legend cell align=left,
	legend columns=-1,
	legend to name=leg:G_sample_sets_analysis_residuals,
	legend style={/tikz/every even column/.append style={column sep=0.25cm}},
	]
	\pgfplotsset{samples=50};
	\addplot+[only marks,mark=x,mark size=1.5pt] table[x index=0,y index=4] {data/g_poly_0-0.txt};
	\addplot+[only marks,mark=+,mark size=1.5pt] table[x index=0,y index=3] {data/g_poly_0-0.txt};
	\addplot[red] {1.25881763};
	\addplot[dashed,blue] {1.84368988};
	
	\addlegendentry{Quadratic};
	\addlegendentry{Linear};
	\addlegendentry{Quadratic avg.};
	\addlegendentry{Linear avg.};
\end{axis}
\end{tikzpicture}
\begin{tikzpicture}
\begin{axis}[
    height=0.43\linewidth,
    width=0.56\linewidth,
	xlabel={Pixel depth $d$ [m]},
	xmin=0.5, xmax=5,
	domain=0.5:5,
	cycle list name=color list,
	legend pos=north west,
	legend cell align=left,
	]
	\pgfplotsset{samples=50};
	\addplot+[only marks,mark=x,mark size=1.5pt] table[x index=0,y index=4] {data/g_poly_1-1.txt};
	\addplot+[only marks,mark=+,mark size=1.5pt] table[x index=0,y index=3] {data/g_poly_1-1.txt};
	\addplot[red] {1.2392550111};
	\addplot[dashed,blue] {2.0958941326};
\end{axis}
\end{tikzpicture}
\end{flushright}

\caption{Analysis of two of the sample sets used to estimate the global correction map $\G$.
         The sample set is composed by pairs $(x, y) = (\widehat{z}, z_\map{B})$, where
         $\widehat{z}$ is the sensor-provided depth while $z_\map{B}$ is the depth defined by the
         checkerboard $\obj{B}$ attached on the wall.
         The plots at the top show the samples as well as a linear and a quadratic function that
         best approximate them.
         The plots in the middle, instead, show the difference between the samples' $y$ and $x$
         values, i.e. $z_\map{B} - \widehat{z}$.
         Finally, in the bottom plots, the residuals for the two functions are reported.}
\label{fig:g_sample_sets_analysis}
\end{figure}
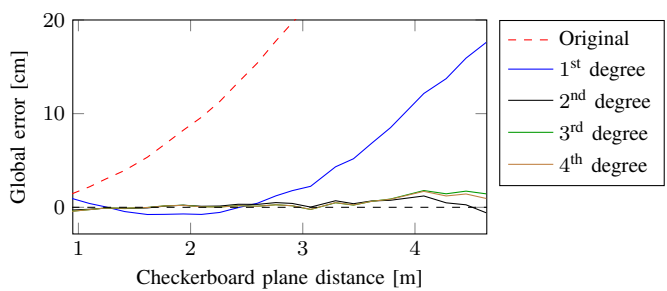
\begin{figure}
\centering
\footnotesize
\begin{tikzpicture}
\begin{axis}[
    height=0.50\linewidth,
    width=0.8\linewidth,
	xlabel={Checkerboard plane distance [m]}, ylabel={Global error [cm]},
	ymax=20,
	xmin=0.95, xmax=4.64,
	cycle list name=color list,
	legend pos=outer north east,
	legend cell align=left,
	]
	\addplot+[no marks,dashed] table[x index=0,y index=2] {data/G_changing_poly_degree.txt};
	\addlegendentry{Original};
	\addplot+[no marks] table[x index=0,y index=3] {data/G_changing_poly_degree.txt};
	\addlegendentry{$1^\text{st}$ degree};
	\addplot+[no marks] table[x index=0,y index=4] {data/G_changing_poly_degree.txt};
	\addlegendentry{$2^\text{nd}$ degree};
	\addplot+[no marks,green!60!black] table[x index=0,y index=5] {data/G_changing_poly_degree.txt};
	\addlegendentry{$3^\text{rd}$ degree};
	\addplot+[no marks] table[x index=0,y index=6] {data/G_changing_poly_degree.txt};
	\addlegendentry{$4^\text{th}$ degree};
	\addplot[dashed,black] {0};
\end{axis}
\end{tikzpicture}
\caption[Global error when varying the maximum degree of the global correction polynomials.]
        {Global error when varying the maximum degree of the global correction polynomials.
         From the plot we can see that linear functions are not able to model (and correct) the
         error on the average depth estimation.}
\label{fig:G_changing_poly_degree}
\end{figure}
We repeated such tests also for the ToF camera, in this case testing a 3-degree and a 6-degree polynomial, that are functions suitable to model the wiggling error, as reported by \cite{FoixSensors2011}. We reported the better results using the 6-degree polynomial with non zero constant factor: all the calibrations of the ToF sensor reported in the following were performed using this function.

\subsubsection{Test Set Results}

\begin{figure}
\centering
\footnotesize
\pgfplotslegendfromname{leg:G_final_results_legend}
\subfloat[Results of the calibration for the three SL sensors using the device camera.]
{
\begin{tikzpicture}
\begin{axis}[
    height=0.40\linewidth,
    width=0.42\linewidth,
	xlabel={Estim. depth [m]},
	ylabel={Global error [cm]},
	xmin=1, xmax=4.6,
	cycle list name=color list,
	legend cell align=left,
	legend columns=-1,
	legend to name=leg:G_final_results_legend,
	legend style={/tikz/every even column/.append style={column sep=0.25cm}},
	title={\textsc{kinect1a}},
	]
	\addplot+[no marks] table[x index=0,y index=1] {data/G_kinect_47A.txt};
	\addplot+[no marks] table[x index=0,y index=3] {data/G_kinect_47A.txt};
	\addplot+[no marks,green!60!black] table[x index=0,y index=2] {data/G_kinect_47A.txt};
	\addplot+[dashed,black] {0};
	\addlegendentry{Original};
	\addlegendentry{Undistorted};
	\addlegendentry{Final};
\end{axis}
\end{tikzpicture}
\begin{tikzpicture}
\begin{axis}[
    height=0.40\linewidth,
    width=0.42\linewidth,
	xlabel={Estim. depth [m]},
	xmin=1, xmax=4.5,
	cycle list name=color list,
	title={\textsc{kinect1b}},
	]
	\addplot+[no marks] table[x index=0,y index=1] {data/G_kinect_51A.txt};
	\addplot+[no marks] table[x index=0,y index=3] {data/G_kinect_51A.txt};
	\addplot+[no marks,green!60!black] table[x index=0,y index=2] {data/G_kinect_51A.txt};
	\addplot+[dashed,black] {0};
\end{axis}
\end{tikzpicture}
\begin{tikzpicture}
\begin{axis}[
    height=0.40\linewidth,
    width=0.42\linewidth,
	xlabel={Estim. depth [m]},
	xmin=1, xmax=4.4,
	cycle list name=color list,
	title={\textsc{asus}},
	]
	\addplot+[no marks] table[x index=0,y index=1] {data/G_asus.txt};
	\addplot+[no marks] table[x index=0,y index=3] {data/G_asus.txt};
	\addplot+[no marks,green!60!black] table[x index=0,y index=2] {data/G_asus.txt};
	\addplot+[dashed,black] {0};
\end{axis}
\end{tikzpicture}
}\\
%
\subfloat[Results of the calibration of two Kinect 1 sensors using the external, high resolution camera.]
{
\begin{tikzpicture}
\begin{axis}[
    height=0.40\linewidth,
    width=0.42\linewidth,
	xlabel={Estim. depth [m]},
	ylabel={Global error [cm]},
	xmin=1, xmax=4.6,
	cycle list name=color list,
	title={\textsc{kinect1a}},
	]
	\addplot+[no marks] table[x index=0,y index=1] {data/G_kinect_47A_p.txt};
	\addplot+[no marks] table[x index=0,y index=3] {data/G_kinect_47A_p.txt};
	\addplot+[no marks,green!60!black] table[x index=0,y index=2] {data/G_kinect_47A_p.txt};
	\addplot+[dashed,black] {0};
\end{axis}
\end{tikzpicture}
\begin{tikzpicture}
\begin{axis}[
    height=0.40\linewidth,
    width=0.42\linewidth,
	xlabel={Estim. depth [m]},
	xmin=1, xmax=4.5,
	cycle list name=color list,
	title={\textsc{kinect1b}},
	]
	\addplot+[no marks] table[x index=0,y index=1] {data/G_kinect_51A_p.txt};
	\addplot+[no marks] table[x index=0,y index=3] {data/G_kinect_51A_p.txt};
	\addplot+[no marks,green!60!black] table[x index=0,y index=2] {data/G_kinect_51A_p.txt};
	\addplot+[dashed,black] {0};
\end{axis}
\end{tikzpicture}
}\\
\subfloat[Results of the calibration of the Kinect 2 sensor.]
{
\hspace{1cm}
\begin{tikzpicture}
\begin{axis}[
    height=0.40\linewidth,
    width=0.48\linewidth,
	xlabel={Estim. depth [m]},
	ylabel={Global error [cm]},
	xmin=1, xmax=5,
	cycle list name=color list,
	title={\textsc{kinect2}},
	]
	\addplot+[no marks] table[x index=0,y index=1] {data/G_kinect2_p.txt};
	\addplot+[no marks] table[x index=0,y index=3] {data/G_kinect2_p.txt};
	\addplot+[no marks,green!60!black] table[x index=0,y index=2] {data/G_kinect2_p.txt};
	\addplot+[dashed,black] {0};
\end{axis}
\end{tikzpicture}
\hspace{1cm}
}
\caption{Global error for the three tested SL depth sensors and the Kinect 2 ToF camera. We report the error for the original point cloud (Original), the error after the undistortion step (Undistorted), and the error after both the undistortion and the global error correction steps (Final). For the three SL sensors, to further assess the validity of the proposed approach, in (a) we calibrated the depth sensor using the device camera, while in (b) we used the external high resolution camera (see \figref{fig:setup_a}).}
\label{fig:G_final_results}
\end{figure}
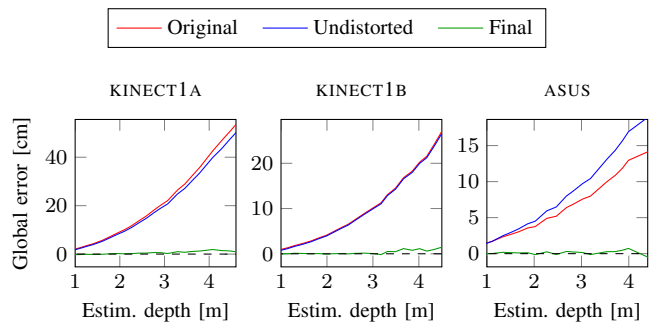
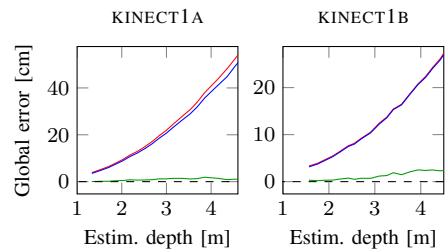
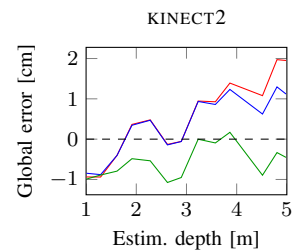

We finally estimated the global correction results on the acquired test sets.
For each test cloud we evaluated the distances of the points of the main plane to the plane
defined by the checkerboard and computed their mean.
The plots in \figref{fig:G_final_results} show the results of such evaluation. 
The proposed error correction approach is working as expected: all the points are correctly
translated to the right location, with respect to the checkerboard pose.

\subsection{Testing the Whole Procedure}\label{sec:total_experiments_exp}

The following tests are meant to evaluate the results of the proposed calibration approach when dealing with real world data.
To this aim, we first compare the wall average depths obtained after the calibration with the measurements given by the laser meters, then the transformations between the depth sensors and the cameras are evaluated in terms of visual results and expected values.

\subsubsection{Depth Calibration}

\begin{figure}
\centering
\footnotesize
\pgfplotslegendfromname{leg:final_results_legend}
\subfloat[Results of the calibration of the three SL sensors using the device camera.]
{
\begin{tikzpicture}
\begin{axis}[
    height=0.40\linewidth,
    width=0.42\linewidth,
	xlabel={G.t. depth [m]},
	ylabel={Distance [cm]},
	ymin=-5, ymax=20,
	xmin=1, xmax=4.6,
	cycle list name=color list,
	title={\textsc{kinect1a}},
	]
	\addplot+[no marks] table[x index=0,y index=2] {data/final_kinect_47A.txt};
	\addplot+[no marks] table[x index=0,y index=1] {data/final_kinect_47A.txt};
	\addplot+[no marks,green!60!black] table[x index=0,y index=3] {data/final_kinect_47A.txt};
	\addplot+[dashed,black] {0};
\end{axis}
\end{tikzpicture}
\begin{tikzpicture}
\begin{axis}[
    height=0.40\linewidth,
    width=0.42\linewidth,
	xlabel={G.t. depth [m]},
	ymin=-5, ymax=20,
	xmin=1, xmax=4.5,
	cycle list name=color list,
	title={\textsc{kinect1b}},
	]
	\addplot+[no marks] table[x index=0,y index=2] {data/final_kinect_51A.txt};
	\addplot+[no marks] table[x index=0,y index=1] {data/final_kinect_51A.txt};
	\addplot+[no marks,green!60!black] table[x index=0,y index=3] {data/final_kinect_51A.txt};
	\addplot+[dashed,black] {0};
\end{axis}
\end{tikzpicture}
\begin{tikzpicture}
\begin{axis}[
    height=0.40\linewidth,
    width=0.42\linewidth,
	xlabel={G.t. depth [m]},
	ymin=-5, ymax=20,
	xmin=1, xmax=4.4,
	cycle list name=color list,
	title={\textsc{asus}},
	]
	\addplot+[no marks] table[x index=0,y index=2] {data/final_asus.txt};
	\addplot+[no marks] table[x index=0,y index=1] {data/final_asus.txt};
	\addplot+[no marks,green!60!black] table[x index=0,y index=3] {data/final_asus.txt};
	\addplot+[dashed,black] {0};
\end{axis}
\end{tikzpicture}
}\\
\subfloat[Results of the calibration of two Kinect 1 sensors using the external, high resolution camera.]
{
\hspace{1.2cm}
\begin{tikzpicture}
\begin{axis}[
    height=0.40\linewidth,
    width=0.42\linewidth,
	xlabel={G.t. depth [m]},
	ylabel={Distance [cm]},
	ymin=-5, ymax=20,
	xmin=1.5, xmax=4.6,
	cycle list name=color list,
	title={\textsc{kinect1a}},
	]
	\addplot+[no marks] table[x index=0,y index=2] {data/final_kinect_47A_p.txt};
	\addplot+[no marks] table[x index=0,y index=1] {data/final_kinect_47A_p.txt};
	\addplot+[no marks,green!60!black] table[x index=0,y index=3] {data/final_kinect_47A_p.txt};
	\addplot+[dashed,black] {0};
\end{axis}
\end{tikzpicture}
\begin{tikzpicture}
\begin{axis}[
    height=0.40\linewidth,
    width=0.42\linewidth,
	xlabel={G.t. depth [m]},
	ymin=-5, ymax=20,
	xmin=1.5, xmax=4.5,
	cycle list name=color list,
	legend cell align=left,
	legend columns=-1,
	legend to name=leg:final_results_legend,
	legend style={/tikz/every even column/.append style={column sep=0.25cm}},
	title={\textsc{kinect1b}},
	]
	\addplot+[no marks] table[x index=0,y index=2] {data/final_kinect_51A_p.txt};
	\addplot+[no marks] table[x index=0,y index=1] {data/final_kinect_51A_p.txt};
	\addplot+[no marks,green!60!black] table[x index=0,y index=3] {data/final_kinect_51A_p.txt};
	\addplot+[dashed,black] {0};
	
	\addlegendentry{Original};
	\addlegendentry{Checkerboard};
	\addlegendentry{Corrected};
\end{axis}
\end{tikzpicture}
\hspace{1.2cm}
}\\
\subfloat[Results of the calibration of the Kinect 2.]
{
\hspace{1cm}
\begin{tikzpicture}
\begin{axis}[
    height=0.40\linewidth,
    width=0.48\linewidth,
	xlabel={G.t. depth [m]},
	ylabel={Distance [cm]},
	ymin=-2, ymax=2,
	xmin=1, xmax=5,
	cycle list name=color list,
	legend cell align=left,
	legend columns=-1,
	legend to name=leg:final_results_legend,
	legend style={/tikz/every even column/.append style={column sep=0.25cm}},
	title={\textsc{kinect2}},
	]
	\addplot+[no marks] table[x index=0,y index=2] {data/final_kinect2_p.txt};
	\addplot+[no marks] table[x index=0,y index=1] {data/final_kinect2_p.txt};
	\addplot+[no marks,green!60!black] table[x index=0,y index=3] {data/final_kinect2_p.txt};
	\addplot+[dashed,black] {0};
	
	\addlegendentry{Original};
	\addlegendentry{Checkerboard};
	\addlegendentry{Corrected};
\end{axis}
\hspace{1cm}
\end{tikzpicture}
}
\caption
        {Distance between the real wall depth and the one estimated with the calibration procedure.
         The error is computed for the original point clouds (Original) and for the clouds after
         both the undistortion and the global error correction (Corrected).
         Moreover, the distance of the wall from the color sensor estimated using the checkerboard
         (Checkerboard) is reported.
         For the Kinect 1 sensors, the error is computed using (a) the device camera as the reference camera as well as (b) the external high resolution one.
         Note that there is a fixed offset of about than 1 cm between the laser meters and the
         two Kinect 1  and the Kinect 2 sensors (the sensors are closer to the wall) and about 9
         cm between the laser meters and the high resolution camera (the camera is farther).}
\label{fig:final_results}
\end{figure}
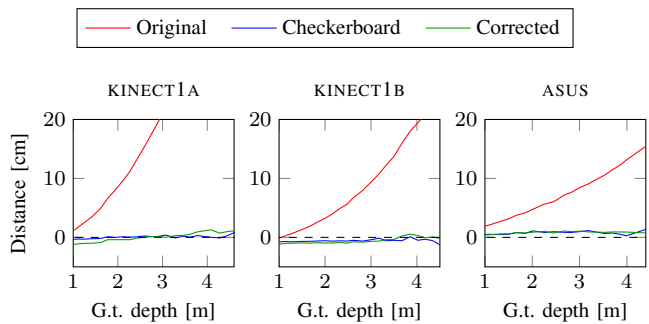
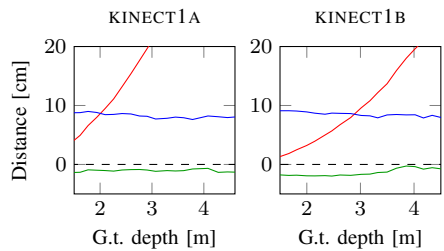
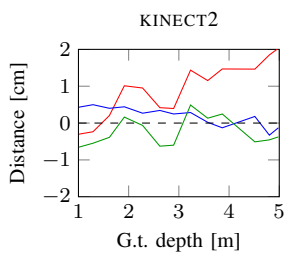

The plots in \figref{fig:final_results} show a quantitative evaluation of the depth error, i.e. the distance between the wall measured by means of the laser meters and the average depth of the points after both the undistortion and global correction phases.
The measured planes are within a couple of centimeters from the real ones: these good results confirm the soundness of our choices.

\begin{figure}
\centering
\footnotesize
\pgfplotslegendfromname{leg:final_results_angle_legend}\\\vspace{0.2cm}
\begin{tikzpicture}
\begin{axis}[
    height=0.40\linewidth,
    width=0.42\linewidth,
	xlabel={G.t. depth [m]}, ylabel={$x$-axis rotation error [\textdegree]},
	ymin=-2, ymax=6,
	xmin=1.5, xmax=4.5,
	cycle list name=color list,
	title={\textsc{kinect1b}},
	]
	\addplot+[no marks] table[x index=0,y index=1] {data/final_kinect_51A_angle.txt};
	\addplot+[no marks] table[x index=0,y index=2] {data/final_kinect_51A_angle.txt};
	\addplot+[dashed,black] {0};
\end{axis}
\end{tikzpicture}
\hspace{0.5cm}
\begin{tikzpicture}
\begin{axis}[
    height=0.40\linewidth,
    width=0.42\linewidth,
	xlabel={G.t. depth [m]},
	ylabel={$y$-axis rotation error [\textdegree]},
	ymin=-2, ymax=6,
	xmin=1.5, xmax=4.5,
	cycle list name=color list,
	legend cell align=left,
	legend columns=-1,
	legend to name=leg:final_results_angle_legend,
	legend style={/tikz/every even column/.append style={column sep=0.25cm}},
	title={\textsc{kinect1b}},
	]
	\addplot+[no marks] table[x index=0,y index=3] {data/final_kinect_51A_angle.txt};
    \addplot+[no marks] table[x index=0,y index=4] {data/final_kinect_51A_angle.txt};
	\addplot+[dashed,black] {0};
	
	\addlegendentry{Original};
	\addlegendentry{Corrected};
\end{axis}
\end{tikzpicture}
\caption[Rotation error for \textsc{kinect1b} sensor.
         Both the rotation about the $x$- and $y$-axis are compared.]
        {Rotation error for \textsc{kinect1b} sensor.
         Both the rotation about the $x$- and $y$-axis are compared.
         The error is the angle estimated between the normal of the plane fitted to the corrected
         data and the theoretical wall plane $x$-axis and $y$-axis.}
\label{fig:final_results_angle}
\end{figure}
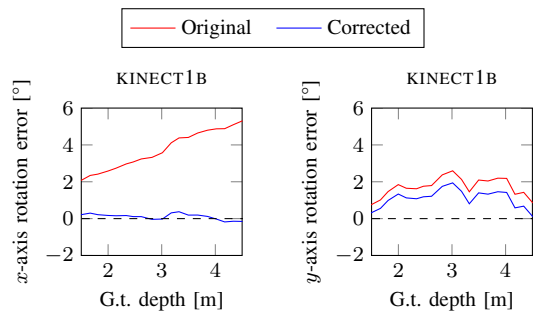

We also evaluated how much the resulting plane is rotated with respect to the real one.
To this aim we computed the angle between the normal of the plane fitted to the corrected data, and
the $x$- and $y$-axis of the wall plane, i.e. $(1, 0, 0)^\T$ and $(0, 1, 0)^\T$ respectively.
Let $\vect{n}$ be the fitted-plane normal and let $\vect{a}$ be the axis with respect to which the
error is computed, the rotation error, $e_\f{rot}$, is
\begin{equation*}
  e_\f{rot} = \arccos(\vect{n}^\T \vect{a}) - \frac{\pi}{2} \enspace.
\end{equation*}
Results of the error computation for sensor \textsc{kinect1b} are reported in
\figref{fig:final_results_angle}.
The figure shows that the rotation about the $x$-axis is completely corrected.
For what concerns the rotation about the $y$-axis, instead, the results are worse.
The reason for this fact is likely to be the error on the real depth estimation.
In fact, a difference of about 2 mm in the depth measures (note that this is the nominal error of
the two laser meters) leads to a rotation of about 0.5\textdegree.

\begin{figure}
\centering
\footnotesize
\subfloat[\textsc{kinect1a} -- top]{\includegraphics[scale=0.18]{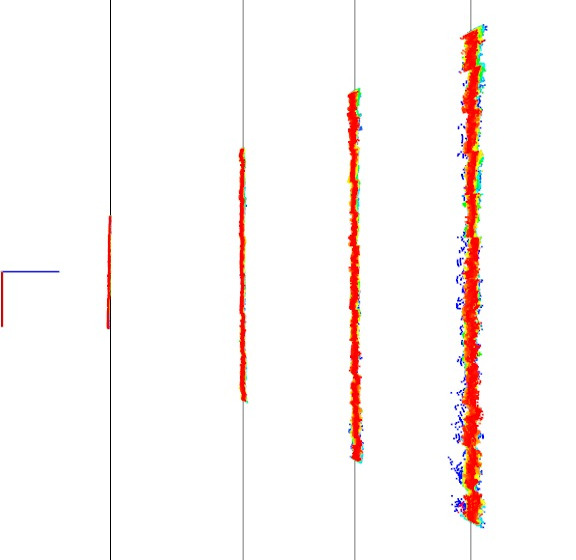}}
\subfloat[\textsc{kinect1b} -- top]{\includegraphics[scale=0.18]{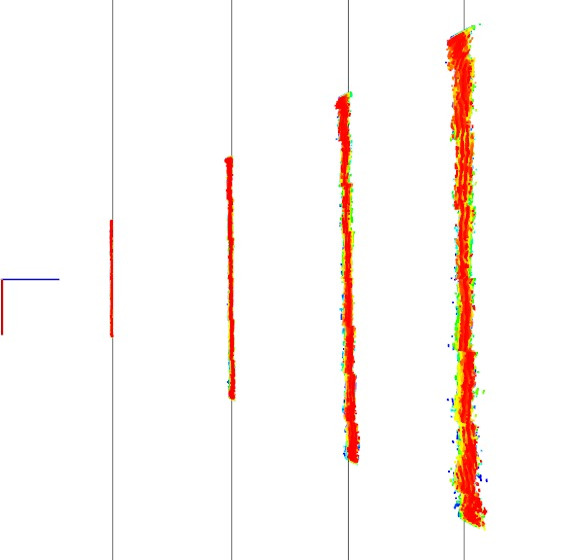}}\\
\subfloat[\textsc{asus} -- top]{\includegraphics[scale=0.18]{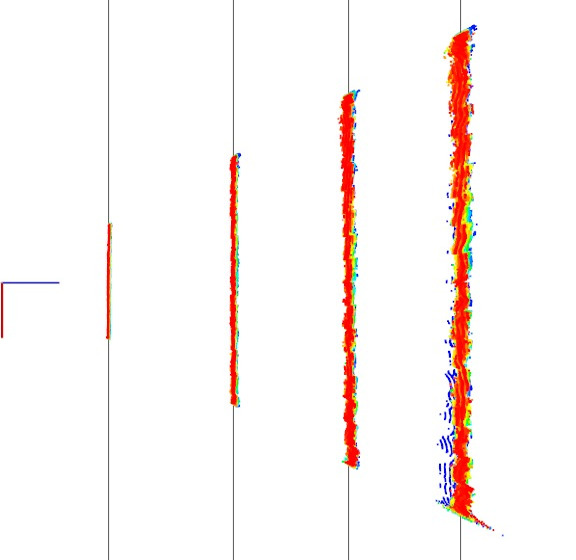}}
\subfloat[\textsc{kinect1b} -- side]{\includegraphics[scale=0.18]{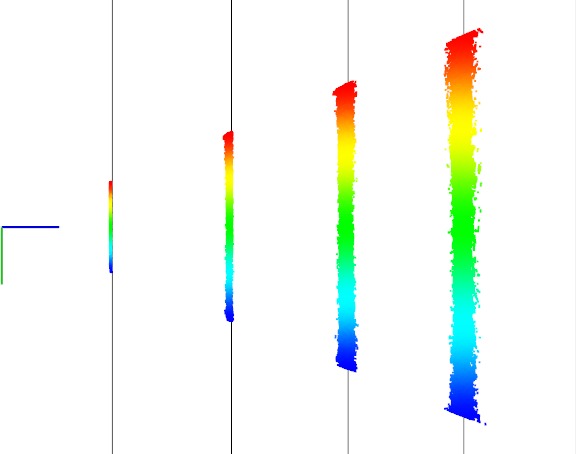}}
\caption[Top and side views of the point clouds of \figref{fig:depth_error1} after the undistortion
         phase (\figref{fig:undistorted_depth_error}) and the global correction phase.]
        {Top and side views of the point clouds of \figref{fig:depth_error1} after the undistortion
         phase (\figref{fig:undistorted_depth_error}) and the global correction phase.
         Again, the gray lines show the depth measured by means of the laser meters.
         As we can see, every cloud is now planar as well as in the right position and correctly
         oriented.}
\label{fig:final_depth_error}
\end{figure}

A further confirmation that the proposed approach works properly, is shown in
\figref{fig:final_depth_error}.
The pictures report the clouds of \figref{fig:depth_error1} after both the undistortion phase (see
\figref{fig:undistorted_depth_error}) and the global error correction.
As expected, all the clouds are now both planar and located correctly.

\subsubsection{Camera-Depth Sensor Transform}

Even if the camera-depth sensor transform $\TF{\matx{T}}{C}{D}$ estimated during the optimization
phase is a sort of ``side effect'' of the depth calibration, a good transformation can be seen as a
proof of the validity of the proposed approach.
In \tabref{tab:transforms} and \tabref{tab:transforms_k2} the transformations that resulted from the calibration of the sensors
using their device cameras, are reported.
Moreover, to give a comparison metric, also the factory-provided transformation is reported.
\begin{table*}
\centering
\small
\caption{Camera-depth sensor transform estimated for the three SL sensors when calibrated using the device camera.
         Both the translation $\vect{t} = (t_x, t_y, t_z)^\T$ and the rotation, represented as a quaternion $\vect{q} = (q_w, q_x, q_y, q_z)^\T$, are reported.
         The \textsc{factory} line contains the factory-provided calibration parameters.}
\label{tab:transforms}
\begin{tabular}{l|ccc|cccc}
\hline 
  & $t_x$ [m] & $t_y$ [m] & $t_z$ [m] & $q_x$ & $q_y$ & $q_z$ & $q_w$ \\ 
\hline 
\hline 
\textsc{factory} & 0.025 & 0 & 0 & 0 & 0 & 0 & 1 \\ 
\hline 
\textsc{kinect1a} & 0.0237 & 0.0044 & -0.0063 & 0.0034 & 0.0060 & -0.0017 & 0.9999 \\ 
\hline 
\textsc{kinect1b} & 0.0276 & 0.0024 & -0.0036 & 0.0025 & 0.0007 & -0.0010 & 0.9999 \\ 
\hline 
\textsc{asus} & 0.0294 & -0.0040 & -0.0011 & 0.0048 & 0.0059 & -0.0004 & 0.9999 \\ 
\hline
\end{tabular} 
\end{table*}
\begin{table*}
\centering
\small
\caption{Camera-depth sensor transform estimated for the Kinect 2 sensor.
         Both the translation $\vect{t} = (t_x, t_y, t_z)^\T$ and the rotation, represented as a quaternion $\vect{q} = (q_w, q_x, q_y, q_z)^\T$, are reported.
         The \textsc{factory} line contains the factory-provided calibration parameters.}
\label{tab:transforms_k2}
\begin{tabular}{l|ccc|cccc}
\hline 
  & $t_x$ [m] & $t_y$ [m] & $t_z$ [m] & $q_x$ & $q_y$ & $q_z$ & $q_w$ \\ 
\hline 
\hline 
\textsc{factory} & 0.052 & 0 & 0 & 0 & 0 & 0 & 1 \\ 
\hline 
\textsc{kinect2} & 0.0565 & 0.0014 & 0.0021 & 0.0200 & -0.0020 & 0.0033 & 0.9997 \\ 
\hline 
\end{tabular}
\end{table*}
The values obtained are similar to those obtainable with other state-of-the-art calibration tools
for RGB-D devices \cite{herrera2012joint, staranowicz2014easy, staranowicz2015pratical,
zhang2011calibration}.

\subsection{Performance Comparison}\label{sec:comparison}

\begin{table*}[t]
\centering
\small
\caption{Comparison of the calibration accuracy for a Kinect 1 sensor.}
\label{tab:performance_eval_kinect1}
\begin{tabular}{l|cccc|ccc}
\hline 
 & $\mu(\varepsilon_3)$ [m] & $\sigma(\varepsilon_3)$ [m] & $\mu(\varepsilon_2)$ [pixels] &  $\sigma(\varepsilon_2)$ [pixels] &  $\mu(\measuredangle_x)$ [\textdegree] & $\mu(\measuredangle_y)$ [\textdegree] & $\mu(\measuredangle_z)$ [\textdegree]\\ 
\hline 
\hline 
\textsc{Original} & 0.129 & 0.059 & 5.836 & 0.955 & 1.596 & 1.541 & 1.796 \\ 
\hline 
\textsc{Herrera \emph{et al.} \cite{herrera2012joint}}  & 0.028 & 0.017 & 2.388 & 0.800& 0.814 & 1.258 & 1.404 \\ 
\hline 
\textsc{Staranowicz \emph{et al.} \cite{staranowicz2015pratical}}  & 0.172 & 0.056 & 4.246 & 1.254 &  1.364 & 1.186 & 1.842\\ 
\hline 
\textsc{Our Method}  & \textbf{0.011} & \textbf{0.004} & \textbf{1.901} & \textbf{0.717} & \textbf{0.691} & \textbf{0.617} & \textbf{0.930} \\ 
\hline 
\end{tabular} 
\end{table*}

\begin{figure}
\centering
\footnotesize
\begin{tikzpicture}
\begin{axis}[
    height=0.58\linewidth,
    width=0.95\linewidth,
	xlabel={Reference cube distance [m]}, ylabel={$\varepsilon_3$ [m]},
	ymax=0.45,
	xmin=1.8, xmax=3,
	cycle list name=color list,
	legend pos=north west,
	legend cell align=left,
	]
	\addplot+[red] table[x index=0,y index=1] {data/performance_eval.txt};
	\addlegendentry{Original};
	\addplot+[yellow] table[x index=0,y index=2] {data/performance_eval.txt};
	\addlegendentry{Herrera \emph{et al.} \cite{herrera2012joint}};
	\addplot+[blue] table[x index=0,y index=3] {data/performance_eval.txt};
	\addlegendentry{Staranowicz \emph{et al.} \cite{staranowicz2015pratical}};
	\addplot+[no marks,green!60!black] table[x index=0,y index=4] {data/performance_eval.txt};
	\addlegendentry{Our Method};	
	\addplot[dashed,black] {0};
\end{axis}
\end{tikzpicture}
\caption{Accuracy in the reference cube localization for increasing depths.}
\label{fig:performance_eval_plot}
\end{figure}
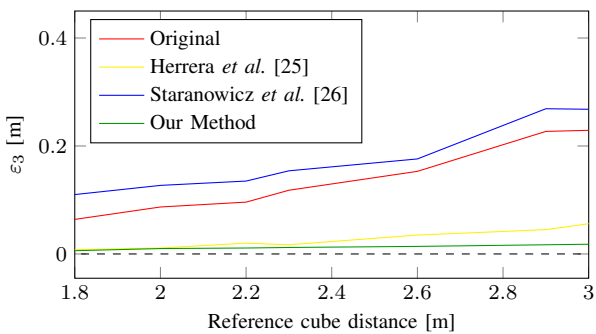

\begin{table*}[t]
\centering
\small
\caption{Comparison of the calibration accuracy for a Kinect 2 sensor.}
\label{tab:performance_eval_kinect2}
\begin{tabular}{l|cccc|ccc}
\hline 
 & $\mu(\varepsilon_3)$ [m] & $\sigma(\varepsilon_3)$ [m] & $\mu(\varepsilon_2)$ [pixels] &  $\sigma(\varepsilon_2)$ [pixels] &  $\mu(\measuredangle_x)$ [\textdegree] & $\mu(\measuredangle_y)$ [\textdegree] & $\mu(\measuredangle_z)$ [\textdegree]\\ 
\hline 
\hline 
\textsc{Original} &  0.081 & 0.014 & 7.733&  2.839  & 2.534 & 2.687 & 1.799 \\
\hline 
\textsc{Our Method}  & \textbf{0.057} & \textbf{0.009} & \textbf{4.680} & \textbf{2.773} & \textbf{1.470} & \textbf{1.932} & \textbf{1.029}\\ 
\hline 
\end{tabular} 
\end{table*}

We tested the calibration accuracy of our system against two state-of-the-art calibration methods, the one from Herrera \emph{et al.} \cite{herrera2012joint} and the one from Staranowicz \emph{et al.} \cite{staranowicz2015pratical}, using the original implementations provided by the authors. Thanks to its robustness and accuracy, \cite{herrera2012joint} is often used as a benchmarking method to calibrate RGB-D pairs; \cite{staranowicz2015pratical} is a more recent method that, compared to \cite{herrera2012joint} and our method, employs a novel, alternative calibration procedure based on a spherical pattern. We used two different types of RGB-D sensors: a Kinect 1 and a Kinect 2. For each method, we acquired large training sets, framing the calibration pattern (a wall for \cite{herrera2012joint} and our method, a basket ball for \cite{staranowicz2015pratical}) from several positions. In order to compare the calibration accuracy, we collected a test set framing a big reference hollow cube with large checkerboards attached to each visible side (\figref{fig:setup_b}). The three checkerboards allow us to compute the plane equations of the three cube sides even if they are not orthogonal to each other. We use these planes, their intersection point $\pp_c$ and its projection in the image plane $\f{proj}_\obj{C}(\pp_c)$ as ground truth data. For each tested method, we estimated the plane equations by fitting the three planes to the (corrected) point clouds, computing also their intersection point $\pp'_c$ as well as its projection in the image plane $\f{proj}_\obj{C}(\pp'_c)$. In our performance comparison, we report the average $\mu(\cdot)$ and standard deviation $\sigma(\cdot)$ of both the errors $\varepsilon_3 = \| \pp_c - \pp'_c \|$ and $\varepsilon_2 = \| \f{proj}_\obj{C}(\pp_c) - \f{proj}_\obj{C}(\pp'_c) \|$, computed for all the images included in the test set. We also report the average of the angular deviations $\measuredangle_x, \measuredangle_y, \measuredangle_z$ between the ground truth planes and the estimated planes. For a perfect calibration, these errors should be obviously $0$.\\
\tabref{tab:performance_eval_kinect1} shows the performance comparison results of the three tested method for a Kinect 1 sensor; as baseline, we also report the results obtained using the original, factory-calibrated data. \figref{fig:performance_eval_plot} shows error $\varepsilon_3$ for increasing depths from the reference cube. Our method clearly provide the better calibration accuracy. The good results obtained by the method from Herrera \emph{et al.} confirm that the choice of a planar calibration pattern enables to obtain superior calibration results. The poor results obtained by the method from \cite{staranowicz2015pratical} are mainly due to the inaccuracies of the calibration pattern detector: we collected several training sets, framing a basket ball (as suggested by the authors) from a large number of different positions, but the provided ball detector often failed to provide an accurate localization.\\
We also compared the calibration accuracy of our method against the factory calibration in the case of a Kinect 2 sensor (\tabref{tab:performance_eval_kinect2}). The method from Herrera \emph{et al.} is here not applicable since it operates directly on the disparity map provided by a SL sensor: clearly a ToF sensor does not provide such map. Also in this case, our method outperforms the factory calibration, enabling to obtain better data also in the case of a ToF sensor. 

\subsection{Visual Odometry Use Case}\label{sec:vo_usecase}

\begin{figure*}[t]
\centering
\footnotesize
\includegraphics[width=0.26\linewidth]{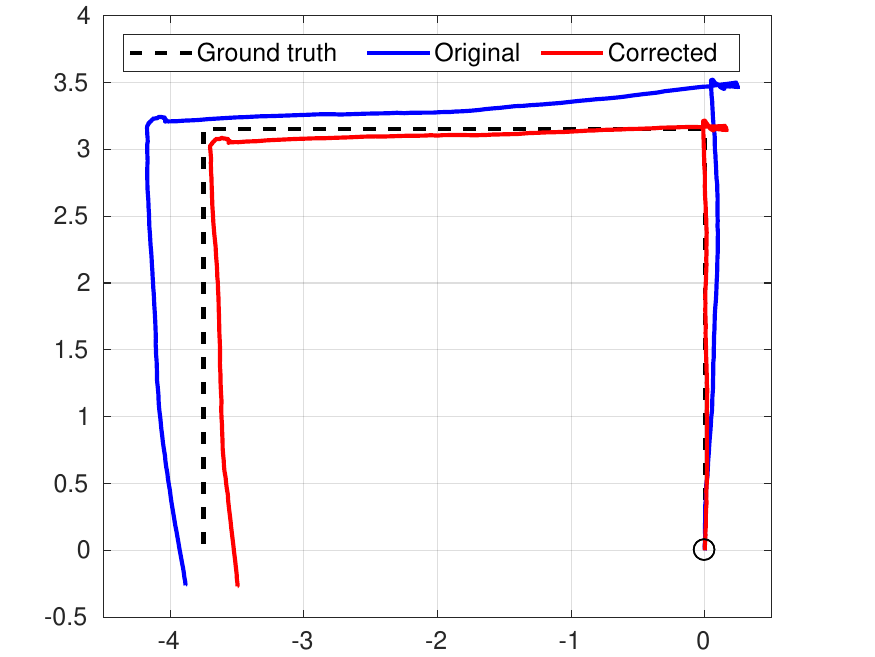}
\includegraphics[width=0.36\linewidth]{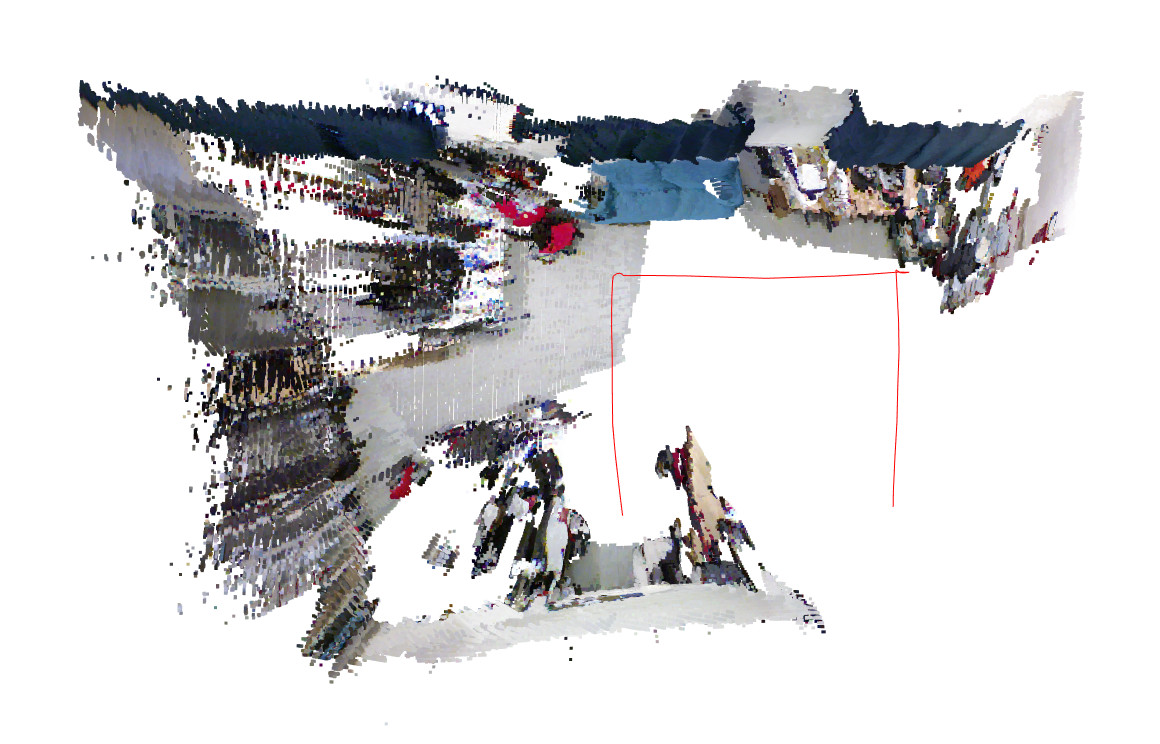}
\includegraphics[width=0.36\linewidth]{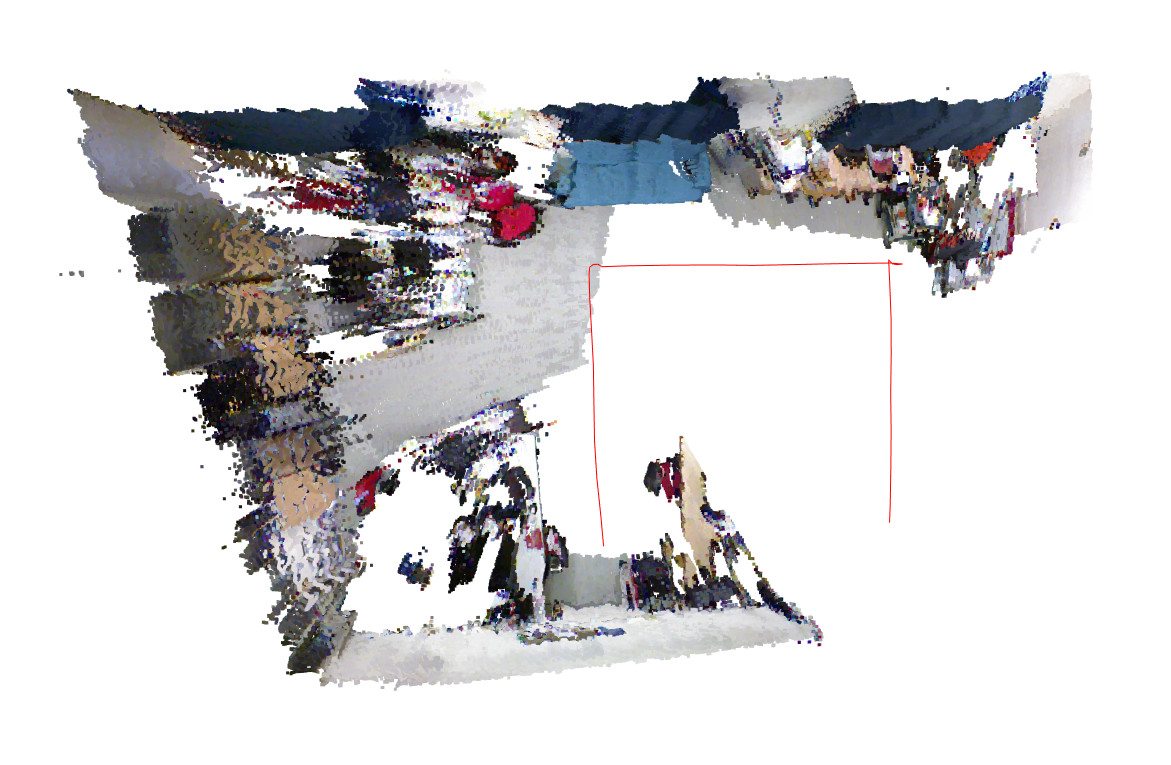}\\
\vspace{0.01cm}
\includegraphics[width=0.27\linewidth]{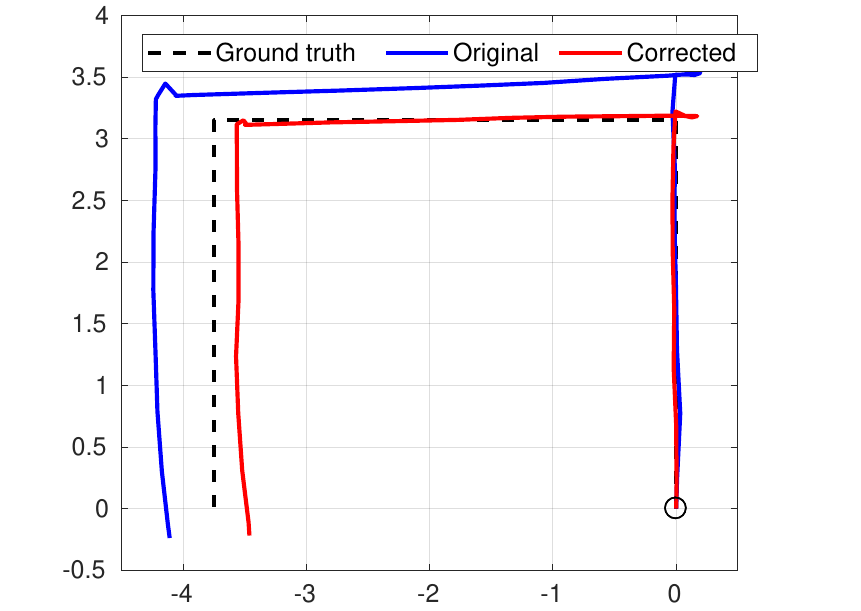}
\includegraphics[width=0.36\linewidth]{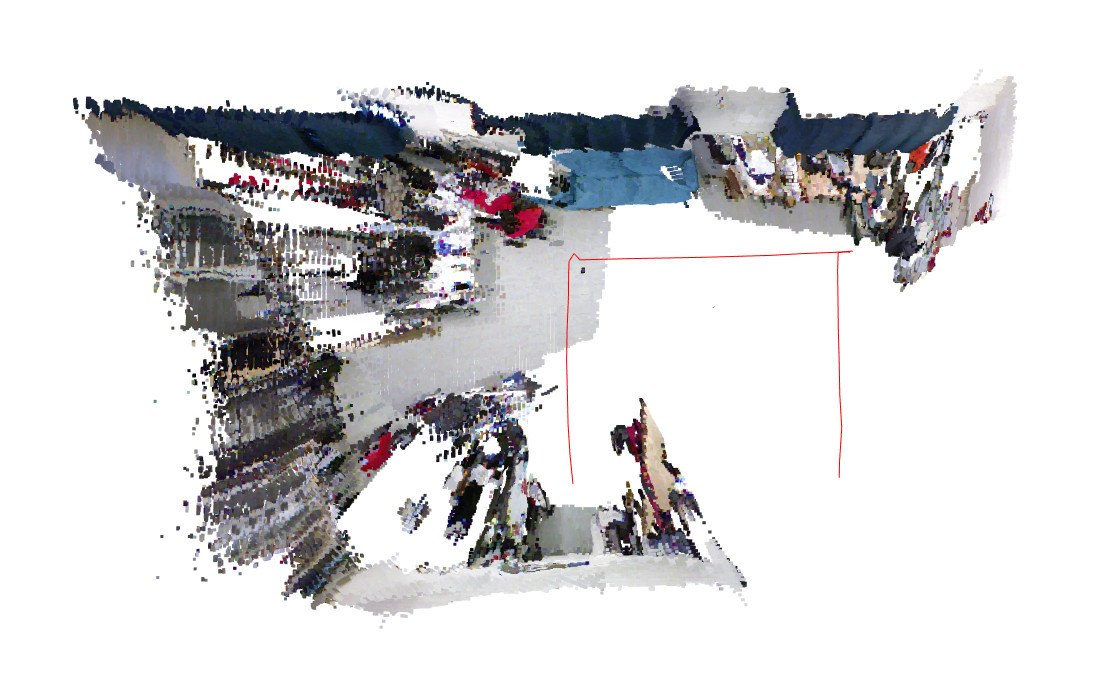}
\includegraphics[width=0.36\linewidth]{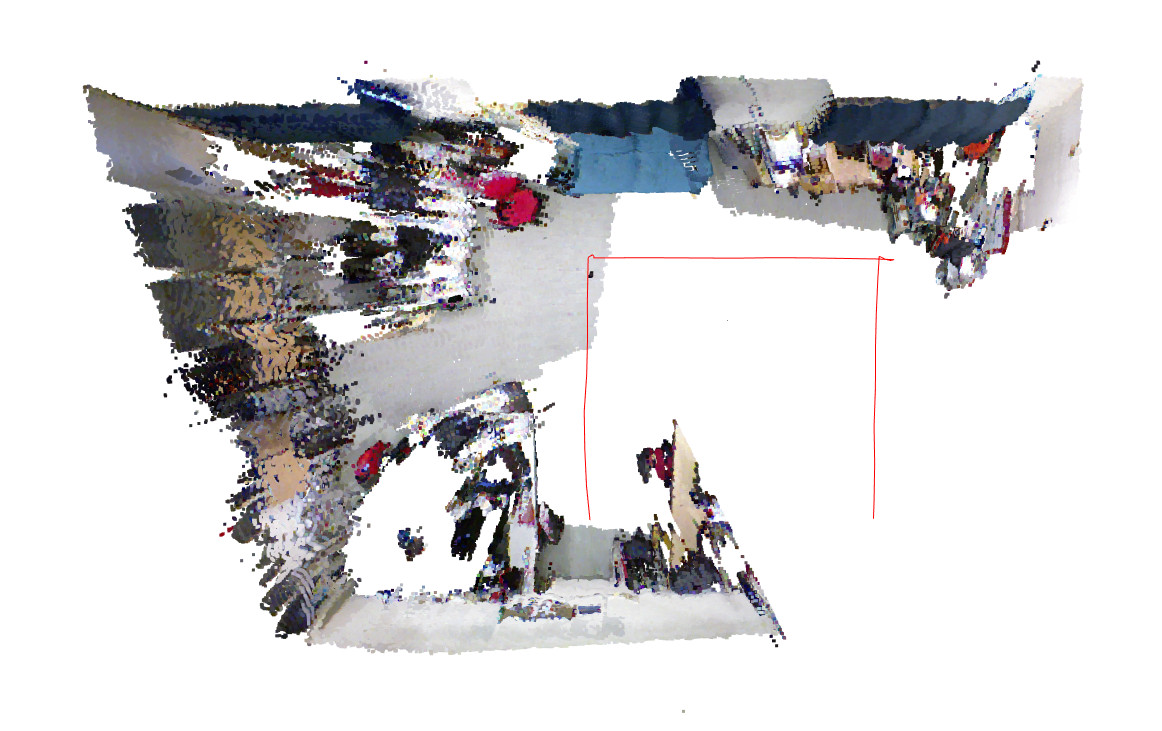}\\
\vspace{0.01cm}
\subfloat[Top view of the trajectories. The start location is surrounded with a circle.]{\hspace{0.27\linewidth}}
\subfloat[Point cloud obtained using the original data.]{\hspace{0.36\linewidth}}
\subfloat[Point cloud obtained using the corrected data.]{\hspace{0.36\linewidth}}
\caption{Qualitative results of the visual odometry experiments: (first row) DVO \cite{kerl13icra} visual odometry systems; (second row) OFVO visual odometry systems.}
\label{fig:visual_odometry}
\end{figure*}

As a further validation of our method, we present an experiment performed using a real robot running an RGB-D visual odometry system. Here we show how the accuracy in the ego-motion estimation and the 3D reconstruction can highly benefit from using RGB-D data calibrated with our method. We employed two different RGB-D visual odometry systems: the popular and robust DVO (Dense Visual Odometry) \cite{kerl13icra} and a very simple custom-built system based on dense optical flow, in the following OFVO (Optical Flow Visual Odometry). DVO registers two consecutive RGB-D frames by minimizing the photometric error between corresponding pixels while OFVO firstly computes the dense optical flow between a reference and the current RGB image and, after generating the point cloud of the reference frame using the depth image, it estimates the relative rigid motion motion by solving a Perspective-n-Point problem\footnote{To solve this problem, we used the OpenCV function \emph{solvePnP}.}. These methods strongly rely on both the intrinsic and extrinsic calibration of the RGB-D pair, so they represent a perfect benchmark for our method which provides the complete calibration for such sensors.\\ 
We moved a mobile robot (a MobileRobots Pioneer 3-AT) equipped with a Microsoft Kinect 1 along a known trajectory, then we estimated the motion\footnote{We estimated the \textit{full} 3D motion, without using any planar motion assumption or wheel odometry prior.} using both DVO and OFVO on the original RGB-D data and on the same data correct with our method. Table \tabref{tab:vo_evaluation} shows the root mean square error (RMSE) of the estimated motion for each experiment: in both cases the accuracy improvement is remarkable. \figref{fig:visual_odometry}(a) shows a top view of the estimated and ground truth trajectories, along with the generated point clouds (\figref{fig:visual_odometry}(b), (c)). For both methods, the trajectory estimated using the corrected data is clearly closer to the ground truth compared to the one estimated using the original data: in the first case, most of the misalignment is mainly due to a drift accumulated in the first turn, where the robot acquired a sequence of frames with very few visual features. The trajectory estimated using the original data tends to diverge also due to a sort of ``scale drift'' effect that supports our choice to introduce the global component in our error model. Similarly, the quality and the precision of the reconstructed point cloud highly benefits from using the corrected data.

\begin{table}
\centering
\small
\caption{RMSE of the drift in meters for the whole path.}
\label{tab:vo_evaluation}
\begin{tabular}{l|cc}
\hline 
  & Original data [m] & Corrected data [m] \\ 
\hline 
\hline 
\textsc{DVO \cite{kerl13icra}} & 0.3371 &  \textbf{0.1589} \\ 
\hline 
\textsc{OFVO} &  0.3530 & \textbf{0.2501}\\ 
\hline 
\end{tabular} 
\end{table}

\subsection{Runtime Performance}

The proposed algorithm has been implemented in C++ as a ROS \cite{ros} package\footnote{\url{http://iaslab-unipd.github.io/rgbd_calibration}}, using the OpenCV \cite{opencv} library for image processing, the Point Cloud Library (PCL) \cite{pcl} for 3D data processing, and the Ceres Solver library \cite{ceres-solver} to solve the optimization problems. To perform the calibration, the user is asked to capture a training set using a tool provided in the package, by moving the sensor in front of a wall with a checkerboard attached on it. Collecting a typical dataset of a hundred images takes no more than 10 minutes.\\

We tested both the calibration and the correction stages in terms of the execution time. The whole calibration process takes about 45 minutes on a consumer laptop\footnote{CPU: Intel Core i7-4700MQ, RAM: 16GB, SSD, GPU: NVidia GTX 750M.} for $640 \times 480$ depth images and 5 minutes by downsampling the depth images to a resolution of $320 \times 240$ pixels: it is worth to mention that in the last case, the calibration accuracy is not significantly reduced.\\
The calibration is an operation that is performed once and so the execution time is not critical. On the other side, the execution time of the correction stage node is critical, since the data generated by the RGB-D sensor, typically with a frequency of 30 Hz, should be processed in real-time. We tested the performance of 3 different implementations of the correction algorithm: 
\begin{itemize}
\item a standard CPU implementation;
\item a parallel CPU implementation exploiting the OpenMP directives;
\item a parallel GPU implementation using CUDA.
\end{itemize}
The results are reported in \tabref{tab:time_results}: note that these execution times include the time to generate the point cloud from the depth image too.
Clearly the GPU implementation outperforms the CPU ones, but dedicated hardware is needed.
Most of the time (about 95\%) spent by the GPU implementation is dedicated to the copy of the data to and from the GPU memory.
Anyway, all the implementations are able to correct the data in real-time.
Also, it is worth to say that the bin size does not affect the performance, since the implementation exploits a lookup-table to store the correction functions.
\begin{table}
\centering
\caption{Time comparison between the 3 different implementations of the correction node.}
\label{tab:time_results}
\begin{tabular}{ccc}
\hline 
CPU [ms] & OpenMP [ms] & GPU [ms] \\ 
\hline
\hline 
15.5 & 8.3 & 2.83 \\ 
\hline 
\end{tabular}
\end{table}
%
%

%% file: conclusions.tex
\section{Conclusions} \label{sec:depth_conclusions}

In this paper we presented a novel method to calibrate a general RGB-D sensor. The proposed calibration procedure only requires the user to collect data in a minimally structured environment, providing in output both the intrinsic and extrinsic parameters of the sensor. We proposed to generalize the depth sensor error by means of two different components, a distortion error and a global, systematic error. The distortion error is modeled using a per-pixel parametric undistortion map, estimated in the first stage of the algorithm. The depth systematic error along with the camera-depth sensor alignment are estimated in the second stage of the algorithm, inside a robust optimization framework. We reported a comprehensive set of tests that support the introduced model. We finally presented exhaustive experiments performed using several sensors, showing that our approach provides highly accurate results, outperforming other state-of-the-art methods.\\
Comparing with other methods, our approach is well suited for different types of depth sensors while requiring a relatively easy calibration protocol. 

\section{Acknowledgments}
The research work is partially supported by: the European Commission under 601116-ECHORD++ (FlexSight experiment) and the University of Padua under the project DVL-SLAM.